\pdfoutput=1
\documentclass{article}
\usepackage{iclr2027_conference,times}
\iclrfinalcopy 

\usepackage{amsmath,amsfonts,bm}

\def\eqref#1{equation~\ref{#1}}

\def\1{\bm{1}}

\DeclareMathAlphabet{\mathsfit}{\encodingdefault}{\sfdefault}{m}{sl}
\SetMathAlphabet{\mathsfit}{bold}{\encodingdefault}{\sfdefault}{bx}{n}

\usepackage{xcolor}

\usepackage{amsthm}
\usepackage{xurl}
\usepackage{hyperref}
\usepackage{cleveref}
\usepackage{amsmath,amssymb}
\usepackage{graphicx}
\usepackage{booktabs}
\usepackage{multirow}
\usepackage{enumitem}
\usepackage{paralist}

\newtheorem{proposition}{Proposition}
\title{Bayesian Evidence Fusion for Uncertainty \\ Quantification of AI Agents}

\author{\bf Artem Vazhentsev\textsuperscript{1} \quad Viktor Mazanov\textsuperscript{2} \quad Aleksandr Kutsakov\textsuperscript{3}\\
\bf Denis Son\textsuperscript{3} \quad Daniil Maslov\textsuperscript{4} \quad Vladislav Smirnov\textsuperscript{1}\\
\bf Timothy Baldwin\textsuperscript{1} \quad Theodore Papamarkou\textsuperscript{5,6} \quad Artem Shelmanov\textsuperscript{1}\\[6pt]
\textsuperscript{1}MBZUAI \quad \textsuperscript{2}Innopolis University \quad \textsuperscript{3}Independent Researcher \quad \textsuperscript{4}YSDA\\
\textsuperscript{5}National Technical University of Athens \quad \textsuperscript{6}PolyShape}

\newcommand{\ecoparagraph}[1]{\par\noindent\textbf{#1}\;}
\newcommand{\uv}{\mathbf{u}}

\begin{document}

\maketitle
\lhead{Preprint} 
\begin{abstract}
    AI agents complete tasks through sequences of diverse actions: large language model (LLM) generations, tool calls, and revisions. Existing uncertainty quantification (UQ) techniques typically assess the reliability of individual LLM generations, leaving open how to aggregate and propagate uncertainty across different types of actions within an agent trajectory. We formulate UQ for LLM agents as Bayesian belief tracking over final task correctness, integrating generation-level uncertainty with tool outcomes through Bayesian updates. Our framework requires no changes to the agentic harness or LLM, and can be applied post hoc without additional agent executions, thus introducing negligible latency. We conduct an extensive evaluation across five LLMs, eight agent harnesses, and 15 agentic environments spanning coding, retrieval, and interactive tasks. 
    The basic Bayesian fusion achieves the highest average outcome-ranking scores on coding and retrieval, while a variant that separately models trajectory length, history, and final-step evidence leads on interactive tasks. Bayesian fusion generalizes well across models, harnesses, and datasets within the same domain. For larger environment and domain shifts, adaptation with little or no labeled target data improves performance.
\end{abstract}

\section{Introduction}
\label{sec:introduction}

Large language models (LLMs) can produce plausible but incorrect responses. Uncertainty quantification (UQ) aims to assess their reliability, enabling systems to flag potential errors, abstain, or request verification. For individual LLM generations, existing methods derive uncertainty from predictive probabilities, semantic disagreement across sampled responses, and model self-assessment~\citep{malinin2021uncertainty,kuhn2023semantic,tian2023calibration}. The effectiveness of these signals depends on the task, output length, and computational budget~\citep{vashurin2025benchmarking}.

UQ is more complex in AI agents, where an LLM completes a task through a sequence of generations and tool calls~\citep{yao2023react,shinn2023reflexion}. Here, the reliability of individual generations does not determine the reliability of the final outcome. Generation-level uncertainty 
provides evidence about the reliability of LLM outputs, whereas tests, critic judgments, and environmental feedback provide partial observations of task progress. For example, a coding agent may confidently generate incorrect code, yet test failures can expose its errors. 
However, on the contrary, passing an incomplete test suite does not guarantee that the final solution is correct. These observations therefore provide imperfect evidence of final task success, and feedback from intermediate steps can play a different role from evidence at the final step.

Neither uncertainty in the final generation alone nor an unweighted average of intermediate uncertainty scores necessarily captures these distinctions. Reliable assessment of the final outcome therefore requires a principled aggregation of generation-level uncertainty and tool feedback across the trajectory. To be practical, this aggregation should also be computationally efficient, minimizing model calls and environment interactions. Recent uncertainty-propagation methods address aspects of this sequential setting~\citep{zhao2025saup,duan2025uprop}, but effectively combining generation-level uncertainty with tool feedback across various agent configurations remains an open challenge.

In this work, to address this challenge, we propose a meta-model that aggregates diverse evidence into a trajectory-level score using a Bayesian approach: observations update a belief about task correctness according to their likelihood under successful and unsuccessful outcomes~\citep{sarkka2013bayesian}. We fit these likelihoods using trajectory-level outcome labels. Our fusion estimators -- termed ``Bayes Fused'' -- combine tool feedback with generation-level uncertainty, including variants that distinguish history from the final generation. A further variant, Structured Bayesian Fusion (SBF), takes into account trajectory length and normalizes accumulated history evidence. The estimators operate post hoc on recorded rollouts without changing the agent's policy or requiring additional agent executions. Their scores can help determine whether to accept an agent’s output, send it for human review, or perform additional verification.

\begin{figure}[t]
    \centering
    \hspace*{1mm}
    \includegraphics[width=0.94\linewidth]{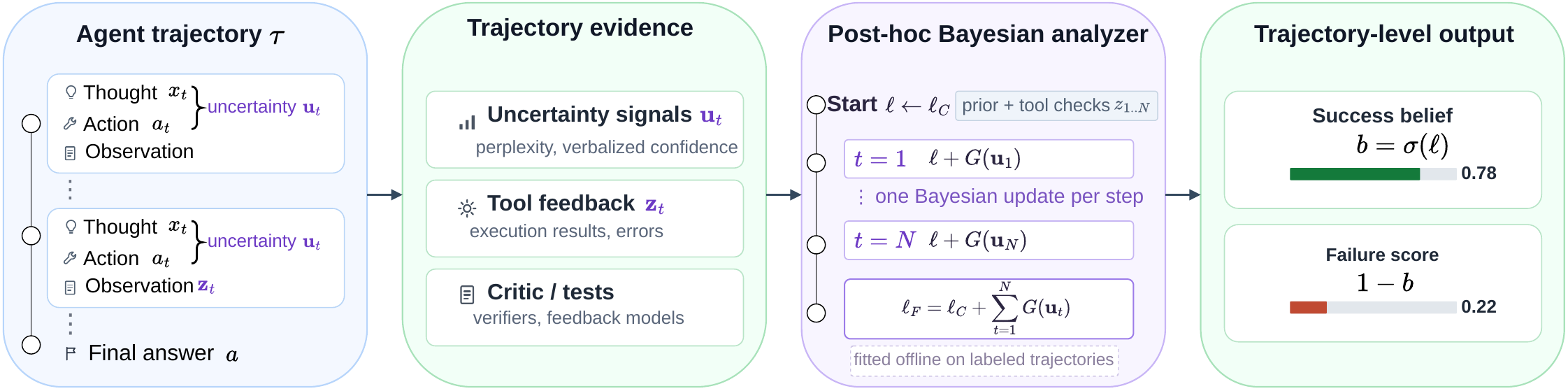}
    \caption{
    \textbf{Overview of our Bayesian framework} for uncertainty quantification
    of agent outcomes.
    }
    \label{fig:framework-overview}
\end{figure}

We evaluate outcome ranking across 15 agentic environments: five for coding, four for retrieval, and six for interactive tasks spanning customer-service interactions, web and desktop use, and embodied planning. The evaluation covers five LLMs and 80 benchmark-model-harness configurations, including both in-distribution performance and OOD generalization (\Cref{sec:id,sec:ood}).

Our main contributions are as follows:
\begin{compactitem}
    \item We formulate UQ for LLM agents as Bayesian belief tracking over final task correctness. Our estimators combine generation-level uncertainty with tool feedback without fine-tuning the agent model. A tool-only variant requires neither token probabilities nor additional LLM calls when tool outcomes are already recorded.

    \item We compare generation-level signals, trajectory summaries, and fitted predictors across 15 agentic environments spanning coding, retrieval, and interactive tasks. Bayesian fusion achieves the highest average outcome-ranking scores on coding and retrieval, while Structured Bayesian Fusion leads on interactive tasks.

    \item We evaluate OOD generalization across models, harnesses, datasets, environments, and domains, and introduce label-free and few-shot adaptation to address larger shifts. Averaged across the evaluated within-domain shifts, Bayes Fused largely preserves ranking performance, while logistic regression degrades substantially. Unlabeled target trajectories improve transfer between WebArena~\citep{2307.13854} and OSWorld~\citep{xie2024osworld}. For the more challenging transfer between ALFWorld~\citep{shridhar2021alfworld} and DeepSWE~\citep{huang2026deepswe}, unlabeled data alone are insufficient, but approximately 10-15 labeled target trajectories enable Bayes Fused to outperform raw uncertainty scores.
\end{compactitem}

\section{Related Work}
\label{sec:related_work}

\ecoparagraph{Uncertainty quantification for LLMs.}
Much work on UQ for LLMs focuses on the reliability of individual generations. Common approaches derive uncertainty from predictive probabilities~\citep{malinin2021uncertainty}, verbalized confidence~\citep{tian2023calibration}, or semantic disagreement across sampled responses~\citep{kuhn2023semantic,lin2024confidence}. Other methods examine the factual content of a response closely. CCP estimates uncertainty about individual claim values~\citep{fadeeva2024ccp}, while FRANQ conditions factuality estimation on faithfulness to retrieved evidence~\citep{fadeeva2026franq}.
For agents, generation-level estimates provide evidence about individual steps, but leave open how to combine that evidence with tool observations to assess the final outcome.

\ecoparagraph{Uncertainty in agent control.}
One way to use uncertainty in an agent is to let it guide decisions during execution. For example, KnowNo uses conformal prediction to determine when to seek help, providing task-completion guarantees under its assumptions~\citep{ren2023knowno}. UALA similarly uses uncertainty to regulate tool use and assistance~\citep{han2024uala}. Uncertainty can also inform how an agent revisits its earlier decisions. Agentic Uncertainty Quantification retains verbalized confidence in memory and uses it to trigger reflection, reporting improvements in task performance and trajectory-level calibration~\citep{zhang2026agentic}. In these approaches, uncertainty estimation is part of the agent's behavior and can influence the course of a rollout.

\ecoparagraph{Uncertainty about agent outcomes.}
A related line of work asks how uncertainty at individual steps translates into uncertainty about the final outcome. SAUP addresses this question by aggregating stepwise uncertainty with situational weights~\citep{zhao2025saup}, while UProp uses trajectory-dependent sampling to estimate uncertainty inherited from earlier decisions~\citep{duan2025uprop}. 
Holistic Trajectory Calibration (HTC) extracts temporal and structural features from token-confidence traces and fits an interpretable calibrator for trajectory success~\citep{zhang2026confidence}. Recent studies show that the effectiveness of generation-level UQ on tool-use trajectories depends on the aggregation method and sampling budget~\citep{bouchard2026trajectory}, and that step-level calibration does not imply trajectory-level calibration~\citep{lin2026agentuq}. Bayesian networks offer another approach to modeling uncertainty in agent workflows. Applications to agentic retrieval report limitations in calibration~\citep{donaldson2026bayesianrag}, while work on multi-agent workflows highlights the difficulty of modeling iterative revisions~\citep{custers2026runtime}.

Our work complements uncertainty propagation across generated decisions by explicitly modeling tool outcomes as evidence of final correctness, which we combine with generation-level uncertainty through Bayesian updates using recorded trajectories without additional rollouts or policy changes, unlike SAUP, UProp, and HTC, which rely on generation-level signals only.

\section{Bayesian Evidence Fusion for Agent Outcomes}
\label{sec:method}

\subsection{Problem Formulation}
\label{sec:problem}

We consider a recorded trajectory $\tau=(o_0,a_1,o_1,\ldots,a_N,o_N)$ generated by an LLM agent whose policy is fixed independently of the uncertainty estimator. At step $t$, the agent produces an action or response $a_t$ and receives feedback $o_t$. The evidence available to the estimator may include generation-level uncertainty signals $\mathbf x_t$ and outcomes from tools or critics. Let $Y\in\{0,1\}$ indicate whether the task is ultimately completed successfully. Here, $N$ denotes the number of steps available for uncertainty estimation. If the environment's terminal action reveals the evaluation outcome, we assess the trajectory using only the preceding steps.

We use a meta-model to map the recorded evidence $\mathcal E(\tau)$ to a score $\widehat p_\theta(\tau)=f_\theta(\mathcal E(\tau))$ that estimates the probability of final task success, $P(Y=1\mid\mathcal E(\tau))$. 
We fit the meta-model using trajectory-level outcome labels from training data and apply it to recorded rollouts without changing the agent's policy. Our evaluation focuses on ranking performance, assessing whether higher failure scores correspond to poorer final outcomes. To prevent outcome leakage, the estimator's inputs exclude the final evaluation label and any observations that directly reveal it. 

\subsection{Bayesian Estimation of Trajectory Success}
\label{sec:belief}

We construct the meta-model using Bayesian inference, treating final task success and failure as competing hypotheses. Starting from a prior probability of success, we update this belief according to how likely the observed generation uncertainty and tool feedback are under each hypothesis.

For brevity, write $\mathcal{E}=\mathcal{E}(\tau)$ for this evidence, and let $\pi=P(Y=1)$ be the prior probability of agent success, with $0<\pi<1$. We define $\ell_0=\log[\pi/(1-\pi)]$ as the \emph{prior log-odds of success}, the logarithm of the ratio of success to failure probabilities before incorporating trajectory evidence. We denote its estimate from training trajectories by $\widehat{\ell}_0$. In log-odds form, Bayes' rule adds the evidence's log-likelihood ratio to this prior term to obtain the posterior log-odds~\citep{bishop2006pattern}:
\begin{equation}
\ell^\star(\mathcal{E})
= \log\frac{P(Y=1 \mid \mathcal{E})}{P(Y=0 \mid \mathcal{E})}
= \ell_0 + \log\frac{p(\mathcal{E} \mid Y=1)}{p(\mathcal{E} \mid Y=0)}.
\label{eq:outcome-logodds}
\end{equation}
The log-likelihood ratio is positive when the observed evidence is more likely under successful outcomes, negative when it is more likely under unsuccessful outcomes, and zero when it is equally likely under both. Tool feedback and generation uncertainty can differ in form and scale: a tool check may return a binary pass/fail outcome, whereas an uncertainty score may be continuous. Their raw values are not directly comparable. Expressing each observation through its log-likelihood ratio converts these different measurements into contributions on the same log-odds scale.

We represent trajectory evidence using multiple signals, such as a tool-check outcome or a generation uncertainty score, and summaries, such as the mean uncertainty over earlier steps. Each signal or summary forms an \emph{evidence channel}. Let $e_j$ denote the observed value of channel $j$. For each outcome $y\in\{0,1\}$, we fit a probability mass or density function $\widehat p_{j,y}(e_j)$ using channel values from training trajectories with label $Y=y$. For example, this means estimating a pass probability for a binary tool check, or a mean and variance for a Gaussian model of a continuous uncertainty score, separately for each outcome class. The fitted channel contribution is $\Lambda_j(e_j)=\log[\widehat p_{j,1}(e_j)/\widehat p_{j,0}(e_j)]$.

If the channels are conditionally independent given $Y$, their joint log-likelihood ratio equals the sum of their individual log-likelihood ratios. Statistical dependence among evidence channels can cause additive fusion to count shared information more than once. Our estimators combine channel contributions additively.
One variant first rescales and shifts the scores accumulated from tool feedback and generation uncertainty over earlier steps. These adjustments, which we call \emph{normalization}, are learned separately for the two history channels from labeled training trajectories.

For a trajectory $\tau$ with resulting score $\ell(\tau)$, we obtain the estimated success probability $\widehat{p}(\tau) = (1+\exp(-\ell(\tau)))^{-1}$. 

\subsection{Evidence Sources and Their Aggregation}

We instantiate the framework using two complementary sources of evidence: (1) feedback from tool calls, and (2) uncertainty in the model's generations. We first describe how tool feedback contributes to the belief in task success, then incorporate generation uncertainty and distinguish the roles of intermediate and final-step evidence.

\ecoparagraph{Tool-only score.} We construct a tool-only score $\ell_C$ by fitting a Bernoulli model for each check type within each outcome class and adding the resulting log-likelihood ratios to the prior log-odds. Each contribution reflects the association between the check outcome and final success, allowing negative check outcomes in trajectories that subsequently recover.

\ecoparagraph{Fusion with generation uncertainty} is done by adding its evidence contribution $\Lambda_U$ to the tool-only score, giving $\ell_F = \ell_C+\Lambda_U$. Let $u_t$ denote the generation uncertainty signal at step $t$. We consider three ways to construct $\Lambda_U$. The \emph{all-steps} variant sums the log-likelihood ratios of $u_t$ over all steps. The \emph{last-only} variant uses only the log-likelihood ratio of the final value $u_N$. The \emph{history-plus-last} variant models the mean of the earlier values $u_1,\ldots,u_{N-1}$ and the final value $u_N$ separately, then adds their log-likelihood ratios. For a one-step trajectory, the history term is omitted.

To compute generation evidence, we estimate the distribution of the observed uncertainty scores $u_t$ separately for successful and failed training trajectories. In the binned all-steps model, we estimate $\widehat p_y(k)$, the probability that a score falls in bin $k$ for trajectories with final outcome $Y=y$, using smoothed counts of training scores. In the Gaussian all-steps model, we estimate a mean $\widehat\mu_y$ and variance $\widehat\sigma_y^2$ for each outcome class, giving the density $\mathcal N(u_t;\widehat\mu_y,\widehat\sigma_y^2)$. The last-only variant fits Gaussian models to final-step scores, while the history-plus-last variant fits separate Gaussian models to history means and final-step scores. The likelihoods under success and failure provide the ratios used in $\Lambda_U$. 
Appendix~\ref{app:likelihood-models} specifies the likelihood models and aggregation rules.

\ecoparagraph{Structured Bayesian Fusion (SBF)} combines three sources of evidence: trajectory length, observations from earlier steps (the history), and observations from the final step. These enter the success score as separate contributions. For a trajectory with $N$ scoring steps, let $K=N-1$ be the number of history steps and $\widehat{\ell}_0$ denote the estimated prior log-odds of success. We~form
\begin{equation}
\ell_{\mathrm{SBF}}
= \widehat{\ell}_0+\Lambda_{\mathrm{len}}
+\Lambda_{H,C}+\Lambda_{H,U}+\Lambda_{\mathrm{last}},
\label{eq:structured-bayesian-fusion}
\end{equation}
where the trajectory-length contribution $\Lambda_{\mathrm{len}}$ reflects whether the rollout length is more typical of successful or failed trajectories. We fit Gaussian distributions to $\log(1+N)$ for each outcome class and compute their log-likelihood ratio for the observed rollout length.

The history contributions $\Lambda_{H,C}$ and $\Lambda_{H,U}$ summarize critic feedback and generation uncertainty over steps $t<N$, respectively. For $\Lambda_{H,C}$, we fit Bernoulli distributions to per-step indicators of observed check failures within each outcome class. The uncertainty signal in $\Lambda_{H,U}$ is computed by combining uncertainty scores at each step using weights learned through logistic regression on training trajectories. For $\Lambda_{H,U}$, we fit Gaussian distributions to these combined scores within each outcome class.
For both history channels $r\in\{C,U\}$, we learn scales $\widehat{\beta}_{H,r}$ and centering values $\widehat{a}_{H,r}$ for affine normalization. They transform the raw accumulated score $\Lambda_{H,r}^{\mathrm{raw}}$ into $\Lambda_{H,r}=\widehat{\beta}_{H,r}(\Lambda_{H,r}^{\mathrm{raw}}-\widehat{a}_{H,r})$. These four coefficients are fitted using trajectory outcome labels and out-of-fold raw scores.
If $K=0$, the trajectory has no history, so both history contributions are 0.

To construct $\Lambda_{\mathrm{last}}$, we use logistic regression to combine generation uncertainty measures at the final step $N$, such as perplexity, token entropy, and verbalized confidence, into a single score. We pair this score with an indicator equal to $1$ if any observed tool check returned a failure outcome at that step and $0$ otherwise. We model this pair using bivariate Gaussian distributions for successful and failed trajectories, allowing correlation within each outcome class.

Appendix~\ref{sec:structured-bayesian-fusion} details how each component is calculated and how its parameters are estimated, and presents an ablation study removing each component separately (\Cref{tab:sbf-components}).

\subsection{Statistical Dependence and History Normalization}

Trajectory length can help predict whether a task succeeds or fails. Having more history steps can also make the average uncertainty score more precise. However, when uncertainty scores across steps are correlated, they contain overlapping information, so each additional step provides less new evidence. Let $h_t$ denote the combined generation uncertainty score at history step $t$. The following proposition uses an idealized Gaussian model conditional on $K=N-1$ to quantify the resulting history evidence and establish a limitation of affine normalization shared across lengths.

\begin{proposition}[Effective history evidence under equi-correlation]
\label{prop:effective-history}
For every positive history length $K$ in the support of both outcome classes, suppose $\mathbf{h}=(h_1,\ldots,h_K)^\top$ satisfies
\begin{equation}
\mathbf{h}\mid(Y=y,K)
\sim \mathcal{N}\left(
\mu_y\mathbf{1}_K,
\sigma^2[(1-\rho)I_K+\rho\mathbf{1}_K\mathbf{1}_K^\top]
\right),
\qquad y\in\{0,1\},
\label{eq:equicorrelated-history}
\end{equation}
where $\mathbf{1}_K$ is the vector of $K$ ones and $I_K$ is the $K\times K$ identity matrix. The population parameters satisfy $\sigma^2>0$ and $0\leq\rho<1$, with $\sigma^2$ and $\rho$ shared across classes. None of $\mu_0,\mu_1,\sigma^2,\rho$ depends on $K$. Writing $\bar h=K^{-1}\sum_{t=1}^{K}h_t$, let
\[
\Lambda_{\mathrm{ind}}(\mathbf{h}\mid K)
=\frac{K(\mu_1-\mu_0)}{\sigma^2}
\left(\bar h-\frac{\mu_1+\mu_0}{2}\right)
\]
denote the log-likelihood ratio obtained under independence using the same class means and variance. The conditional history log-likelihood ratio satisfies
\begin{equation}
\Lambda_{H,U}(\mathbf{h}\mid K)
:=\log\frac{p(\mathbf{h}\mid Y=1,K)}
{p(\mathbf{h}\mid Y=0,K)}
=\frac{\Lambda_{\mathrm{ind}}(\mathbf{h}\mid K)}
{1+(K-1)\rho}.
\label{eq:effective-history}
\end{equation}
If $\mu_1\neq\mu_0$, $\rho>0$, and the common support contains at least two different positive lengths, there are no constants $\beta,\gamma\in\mathbb{R}$ shared across lengths for which $\beta\Lambda_{\mathrm{ind}}(\mathbf{h}\mid K)+\gamma=\Lambda_{H,U}(\mathbf{h}\mid K)$ for every history vector at both lengths.
\end{proposition}

Under the assumptions of \Cref{prop:effective-history}, independent accumulation scales the magnitude of the conditional history log-likelihood ratio by $1+(K-1)\rho$, inflating non-zero evidence when $K>1$ and $\rho>0$. The number of steps may help predict success or failure, but correlated uncertainty scores provide less additional information with each new step. This motivates the separate trajectory-length and history components in SBF, as well as the normalization of its history contributions. The claim about the slope $\beta$ and offset $\gamma$ in \Cref{prop:effective-history} shows that, under its assumptions, a common affine normalization across lengths cannot reproduce the conditional history log-likelihood ratio, even when the population means and variance are known. SBF uses the same rescaling and shift across trajectory lengths, so its history normalization cannot fully correct for dependence between steps under the proposition's assumptions. Nevertheless, SBF with this limited generation-history normalization performs well in our experiments (\Cref{sec:results}).
Appendix~\ref{app:ablations} isolates what the normalization contributes: replacing $\Lambda_{H,C}$ and $\Lambda_{H,U}$ by the raw accumulated scores reduces the average PRR, most strongly on ALFWorld and CRAG (\Cref{tab:sbf-normalization}). The length correction suggested by \Cref{prop:effective-history} improves both ranking and calibration, and SBF performs on par with it (\Cref{tab:length-correction}).
Note that the proposition concerns an idealized model of generation history and does not guarantee that SBF's predicted success probabilities are calibrated. Appendix~\ref{app:effective-history-proof} gives the proof and consequences of \Cref{prop:effective-history}.

\section{Experimental Setup}
\label{sec:setup}

\ecoparagraph{Benchmarks and tasks.}
We consider 15 agentic environments across three task families: (1) \emph{Retrieval} (HotpotQA~\citep{yang2018hotpotqa}, MuSiQue~\citep{trivedi2021musique}, CRAG~\citep{yang2024crag}, BrowseComp-Plus~\citep{chen2025BrowseCompPlus}); (2) \emph{Coding} (LiveCodeBench-Hard (LCB)~\citep{jain2024livecodebench}, BigCodeBench-Hard (BCB)~\citep{zhuo2024bigcodebench}, SWE-bench~\citep{jimenez2024swebench}, CodeContests (CC)~\citep{codecontests2022}, DeepSWE~\citep{huang2026deepswe}); and (3) \emph{Interactive} ($\tau^3$-bench: retail, airline, telecom~\citep{yao2024tau,barres2025tau2}; WebArena-Verified~\citep{hattami2025webarena}, OSWorld-Verified~\citep{xie2024osworld}, ALFWorld~\citep{shridhar2021alfworld}). They differ in actions and tools available to the agent, the meaning of tool feedback, and relationship between intermediate steps and the final answer.
Environment descriptions and dataset statistics (\Cref{tab:retrieval-stats,tab:coding-stats,tab:interactive-stats}) are detailed in Appendix~\ref{app:experiments}.

\ecoparagraph{Models, agent harnesses, and evidence.}
We experiment with five models of different sizes: GPT-OSS-20B and GPT-OSS-120B~\citep{openai2025gptossb}, Qwen2.5-32B-Instruct~\citep{qwen2024qwen}, Qwen3.6-35B-A3B~\citep{qwen36plus}, and Qwen3.8-27B~\citep{qwen38}. Each model is evaluated with several harnesses, including LangGraph ReAct~\citep{yao2023react,langgraph}; CodeAgent and ToolCallingAgent from smolagents~\citep{smolagents}; and benchmark-specific agents for DeepSWE, WebArena, $\tau^3$-bench, and OSWorld. We evaluate 80 benchmark-model-harness configurations in total. The configurations and tools are detailed in Appendix~\ref{app:experiments}.
The critics are lightweight boolean checks based on information that the harness already provides: whether an action is well-formed, whether a tool call executes successfully, and whether it changes the environment state (detailed in Tables \ref{tab:retrieval-critics}, \ref{tab:coding-critics}, and \ref{tab:interactive-critics} in Appendix~\ref{app:experiments}). They follow directly from the information available in each task, and defining them requires no tuning on outcomes and no changes to the agent or the harness.

\ecoparagraph{Evaluation protocol.}
We consider in-distribution (ID) and out-of-distribution (OOD) evaluation with five folds and three split seeds. In ID, estimators are fitted and evaluated within the same benchmark--model--harness configuration. In OOD, they are fitted on a source configuration and applied unchanged to a target configuration. When configurations share tasks, folds are aligned so that no source training task appears in target evaluation. We evaluate only completed trajectories, excluding those stopped by the step limit or environment failures. For tasks with continuous scores (e.g., OSWorld and DeepSWE), we use a binary outcome obtained through a simple binarization rule requiring no manually tuned thresholds for training and stratified fold construction. Appendix~\ref{app:exp-evaluation} details the protocol. Retained trajectory statistics are provided in Appendix~\ref{app:experiments}.

\ecoparagraph{Baselines.}
We compare Bayesian estimators against 11 diverse baselines, including statistical, UQ, unsupervised, and supervised methods. 
All baselines are detailed in Appendix~\ref{app:baselines}. We additionally compare against an LLM-as-judge verifier on ALFWorld and DeepSWE (\Cref{tab:llm-judge} in Appendix~\ref{app:results}). An analysis of trajectory-level aggregation schemes for UQ and Bayes Fused methods is presented in Appendix~\ref{app:ablations}.

\ecoparagraph{Metrics.}
We use prediction rejection ratio (PRR) as our primary evaluation measure~\citep{malinin2021uncertainty,vashurin2025benchmarking}. PRR measures how well an uncertainty score ranks incorrect agent solutions above correct ones. It normalizes the rejection curve between random and oracle rankings, with $1$ corresponding to the oracle, $0$ to random ranking, and negative values indicating performance below random. We report PRR@0.5, considering rejection budgets up to $50\%$, as this better reflects practical settings where only the most uncertain instances are rejected. Importantly, PRR evaluates ranking quality rather than probability calibration, thus reflecting agent failure detection performance. For all tasks, PRR is computed using the raw test-pass fraction, including tasks with continuous scores. A detailed description of PRR is provided in Appendix~\ref{app:prr}.

\section{Results}
\label{sec:results}

\subsection{In-Distribution Performance}
\label{sec:id}
\begin{figure}[t]
    \centering
    \includegraphics[trim={0.cm 0.cm 0.cm 0.cm},clip,width=0.98\linewidth]{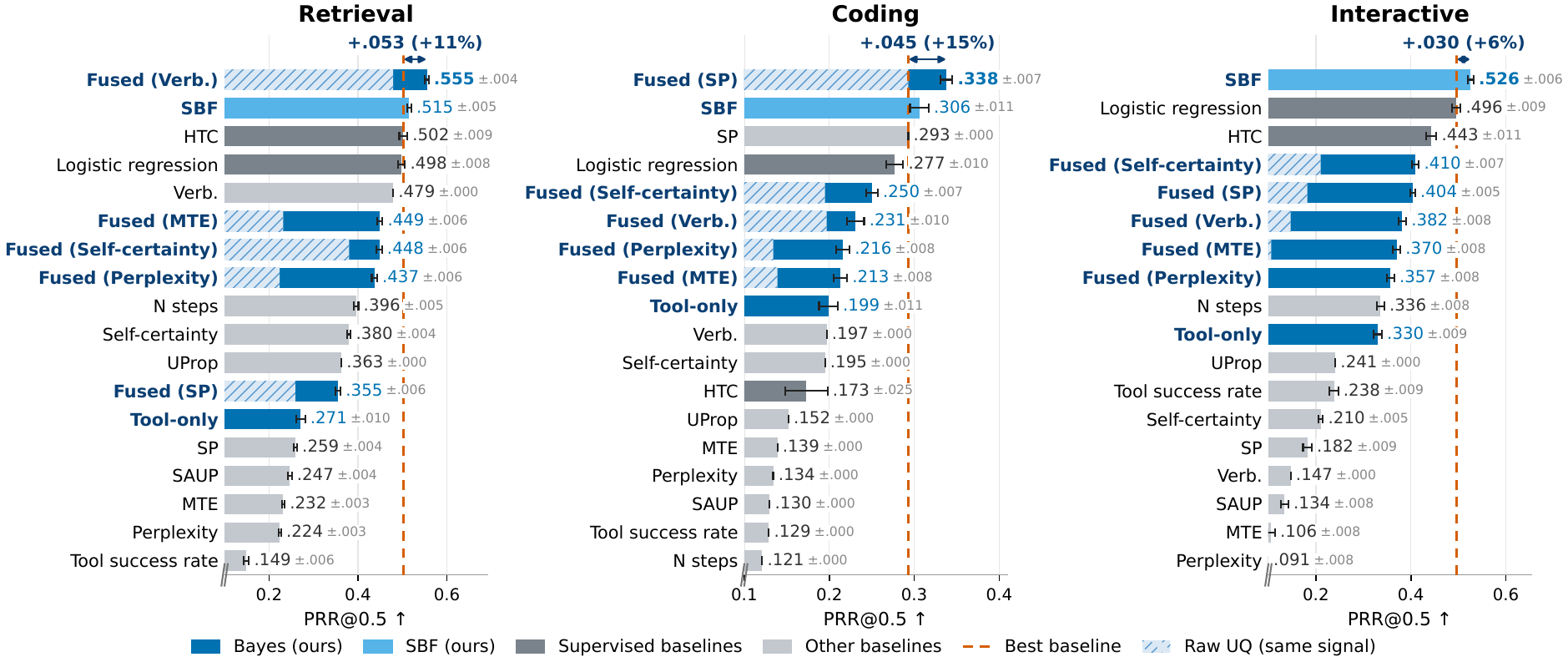}
    \caption{
        \textbf{In-domain results.} Average PRR@0.5$\uparrow$ for each task: (1) retrieval (HotpotQA, MuSiQue, CRAG, BrowseComp-Plus); (2) coding (LCB, BCB, SWE-bench, CodeContests, DeepSWE); (3) and interactive ($\tau^3$-bench retail, airline, and telecom; WebArena, OSWorld, and ALFWorld). Each bar shows the mean over model and harness configurations. 
        Blue bars denote our methods, and gray bars denote baselines. For Bayes Fused, the striped portion of the bar shows the raw UQ score for the same signal, while the solid portion represents the gain from Bayesian fusion. The dashed orange line marks the best baseline. See Appendix~\ref{app:detailed_results_overall} for per-dataset results.
    }
    \label{fig:id_rank}
  \end{figure}

This setting evaluates how well each method detects agent failures when it is fitted on labeled trajectories from the same benchmark, model, and agent as those it scores. 
\Cref{fig:id_rank} reports in-distribution PRR@0.5 for retrieval-based, coding, and interactive tasks, based on the model and harness configurations described in Appendix~\ref{app:experiments}. 
The trajectory-level aggregation used for each task family is specified in the captions of \Cref{tab:prr-dataset-means-rag,tab:prr-dataset-means-coding,tab:prr-dataset-means-other} in Appendix~\ref{app:detailed_results_overall} and detailed in Appendix~\ref{app:exp-evaluation}. Paired task-level bootstrap analysis for ALFWorld, WebArena, OSWorld, and DeepSWE is presented in \Cref{tab:bootstrap} in Appendix~\ref{app:results}.

\ecoparagraph{Retrieval-based tasks.}
Bayes Fused outperforms all corresponding UQ scores, with Bayes Fused (Verb.) achieving the highest average PRR. The gain over raw Verb.\ is particularly high on MuSiQue (see \Cref{tab:prr-dataset-means-rag}), where PRR improves from 0.320 to 0.443. 
Moreover, Bayes Fused (Verb.) achieves the best PRR on HotpotQA and the second-best on MuSiQue and CRAG. The gain from tool feedback comes from evidence about the search process that the final-step uncertainty does not capture. In retrieval-based tasks, critics record whether each search call executed, returned content, and yielded documents or queries not seen earlier (\Cref{tab:retrieval-critics}). An agent whose searches return nothing new, or that cycles through repeated queries, has likely not collected the evidence required for the answer, yet it may still report high confidence. Bayes Fused weights this evidence which lowers the rank of confident answers produced without supporting retrievals. Consistently, the gain is largest where the critics are informative on their own: on MuSiQue and BrowseComp-Plus, Bayes Tool-only reaches PRR of 0.353 and 0.445. 
On CRAG, in contrast, the critics do not check whether new documents were retrieved, so they rank close to random (PRR of $-0.020$), and adding them makes Bayes Fused (Verb.) fall below raw Verb.

\ecoparagraph{Coding tasks.} Bayes Fused improves all UQ scores, with Bayes Fused (SP) achieving the highest average PRR (0.338), followed by Structured Bayesian Fusion (0.306). Both substantially outperform logistic regression and raw UQ methods. The gain over raw SP is largest on DeepSWE (0.017 to 0.207: see \Cref{tab:prr-dataset-means-coding}). On BigCodeBench and CodeContests, the smallest benchmarks, Bayes Fused (SP) improves over raw SP by only 0.008 PRR.
On CodeContests, success rates reach 0.88, leaving only 12\% of failures. On these two small benchmarks, limited training data may make parameter estimation less reliable, particularly for Structured Bayesian Fusion, with more parameters. On BigCodeBench, tool feedback is also weakly informative, which may explain why fusion barely improves on raw SP. On CodeContests, tool feedback is more informative, and both proposed methods remain close to raw SP. 
Overall, our method achieves stronger average performance on coding tasks, with larger training sets and informative tool feedback favoring the fusion of both signals. \Cref{tab:reasoning-effort} in Appendix~\ref{app:results} shows that these gains hold across various reasoning efforts.

  \begin{figure}[t]
    \centering
    \includegraphics[trim={0.cm 0.cm 0.cm 0.cm},clip,width=\linewidth]{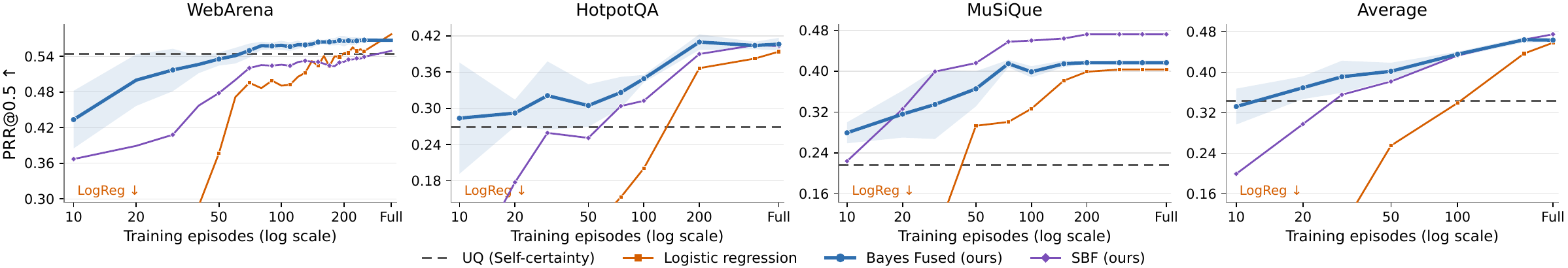}
    \caption{
    \textbf{Training sample efficiency.} PRR@0.5$\uparrow$ by training size for Bayes Fused, logistic regression, and SBF
    averaged over setups, except Qwen2.5-32B-Instruct.
    }
    \label{fig:webarena_hotpotqa_musique_learning_curves}
  \end{figure}

\ecoparagraph{Interactive tasks.}
\Cref{tab:prr-dataset-means-other} presents the results for interactive tasks. Our Structured Bayesian Fusion method achieves the highest average PRR (0.526), followed by logistic regression (0.496), with both substantially outperforming the Bayes Fused variants. Raw UQ methods perform worse on average than simple baselines such as $N$ steps, and all except UProp also fall below tool success rate, while Bayes Fused improves the average PRR of every corresponding UQ score. The further gains from Structured Bayesian Fusion and logistic regression may reflect their more flexible use of trajectory evidence. Logistic regression jointly learns weights for multiple uncertainty and execution features, allowing it to downweight weak or redundant signals. Structured Bayesian Fusion models trajectory length, learns separate projections of uncertainty signals for the history and final step, and normalizes the history contributions using training data. These additional features and learned adjustments may better capture the relationship between execution patterns and task success than the simpler evidence combination used by Bayes Fused.

\ecoparagraph{How many training episodes are needed?}
\Cref{fig:webarena_hotpotqa_musique_learning_curves} shows the dependence of PRR@0.5$\uparrow$ on the number of training episodes for Bayes Fused (Self-certainty), logistic regression, and Structured Bayesian Fusion.
These results highlight an important advantage of the proposed Bayesian framework: strong performance with few training episodes. 
Specifically, even with only 30--50 training episodes, Bayes Fused achieves high performance, while logistic regression is close to random with fewer than 50 episodes.
With around 100 training episodes, Bayes Fused already reaches its near-optimal performance, outperforming other methods, for example, on WebArena.
In contrast, logistic regression typically requires 200 training episodes to approach the performance of Bayes Fused.

Overall, these results highlight that the proposed Bayesian framework can be particularly effective when the number of training episodes is limited, making it well suited for real-world applications where collecting additional training data may be costly or impractical.

Thus, SBF, a single variant used across all domains, outperforms the best baseline in each domain and on average over the 15 benchmarks of \Cref{tab:prr-dataset-means-rag,tab:prr-dataset-means-coding,tab:prr-dataset-means-other} (0.450 vs. 0.424 for logistic regression).

\subsection{Out-of-Distribution Generalization}
\label{sec:ood}

\begin{figure}[t]
    \centering
    \includegraphics[trim={0.cm 0.cm 0.cm 0.cm},clip,width=\linewidth]{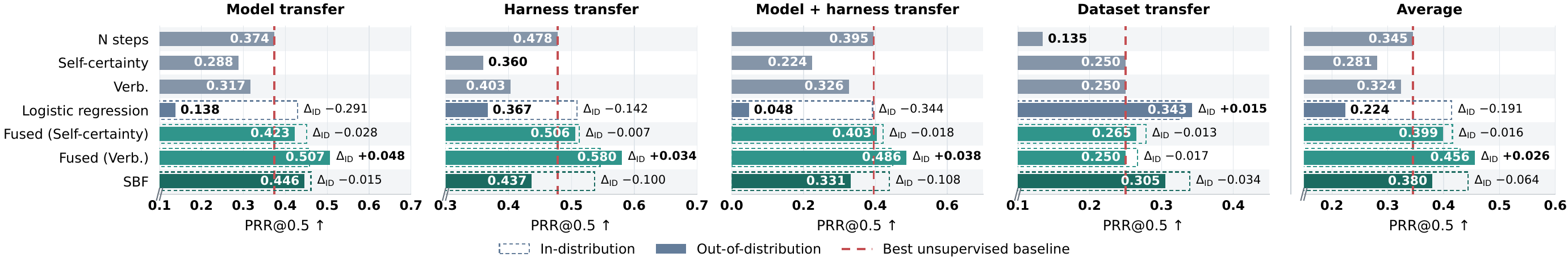}
    \caption{
        \textbf{Within-domain transfer.} Solid bars show OOD PRR@0.5$\uparrow$ for model, harness, joint model--harness, and dataset shifts, while dashed bars show the corresponding in-distribution scores. Each bar represents average performance across the datasets in the transfer setting: (1) model transfer (WebArena, HotpotQA, and MuSiQue); (2) harness transfer (BrowseComp-Plus, HotpotQA, and MuSiQue); (3) joint model--harness transfer (HotpotQA and MuSiQue); and (4) coding dataset transfer (LCB, BCB, SWE-bench, and CodeContests). $\Delta$ denotes OOD performance minus in-distribution performance. Per-transfer results are provided in \Cref{tab:model_transfer,tab:agent_transfer,tab:model_agent_transfer,tab:dataset_transfer} in \Cref{sec:model_transfer}.
    }
    \label{fig:transfer_id_ood_with_avg}
  \end{figure}

\ecoparagraph{Transfer across models, agents, and datasets.}
We consider three transfer settings on a fixed dataset: (i) the same model with different agent harnesses, (ii) the same harness with different models, and (iii) different models and harnesses. We also evaluate cross-dataset transfer with the model and harness held fixed. \Cref{fig:transfer_id_ood_with_avg} summarizes the average out-of-distribution results within the same task domain. Detailed results for all four settings are provided in Appendix~\ref{app:generalization}.

These results indicate that Bayes Fused methods are robust to the evaluated within-domain shifts. Bayes Fused (Verb.) slightly exceeds its in-distribution PRR in every transfer setting except dataset transfer, where PRR decreases by 0.017. Averaged across transfer settings, it shows \textit{no performance degradation}, while Bayes Fused (Self-certainty) incurs a small average PRR decrease of 0.016. By contrast, logistic regression degrades substantially under model and harness shifts, remaining stable only under dataset transfer. Logistic regression and our Structured Bayesian Fusion method show average PRR decreases of 0.191 and 0.064, respectively, suggesting greater overfitting to model- or harness-specific patterns that do not transfer reliably.

\ecoparagraph{Transfer across environments and domains.}
For transfer across domains, we consider two settings: (i) a new environment with a fixed model, and (ii) a new domain with the same model but a different agent. Detailed results for transfer between WebArena and OSWorld, and between ALFWorld and DeepSWE, are provided in \Cref{fig:cross_environment_domain_id_ood} and \Cref{tab:env_transfer,tab:full-shift-alfworld-deepswe} in Appendix~\ref{app:ood_additional_results}.

These results show that all supervised methods struggle to transfer across environments and domains. When the model remains fixed, transfer to a new environment reduces average PRR of the plotted methods by 0.076-0.227, although the methods retain useful ranking performance. Transfer to a new domain is more challenging, with PRR values near zero, indicating performance close to random ranking. These findings highlight the need for adaptation under domain shifts.

    \begin{figure}[t]
    \centering
    \includegraphics[trim={0.cm 0.cm 0.cm 0.cm},clip,width=\linewidth]{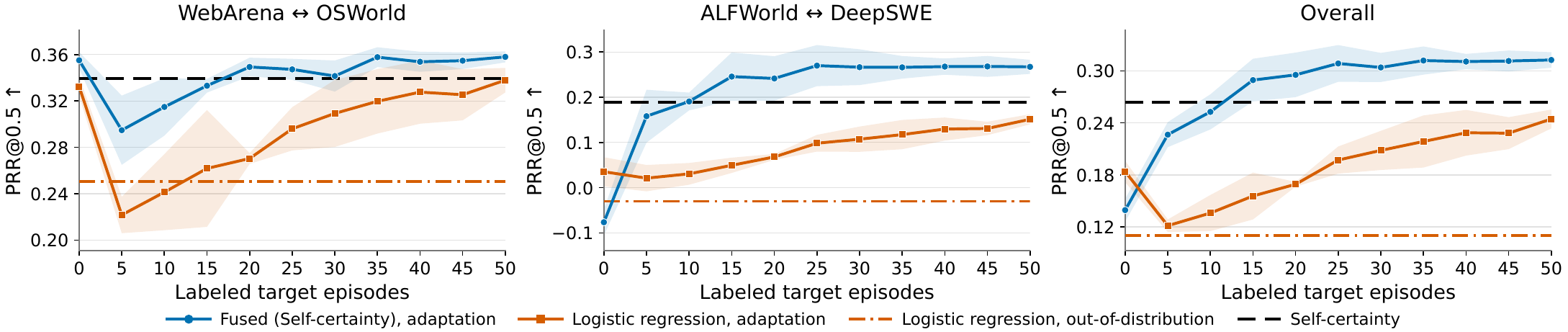}
    \caption{
        \textbf{Adaptation under distribution shift.} PRR@0.5$\uparrow$ by the number of labeled target episodes for Bayes Fused (Self-certainty) and logistic regression, for transfer between WebArena and OSWorld and between ALFWorld and DeepSWE with the same models. Zero corresponds to label-free adaptation. 
        Dashed lines show raw UQ (black) and logistic regression without adaptation.
    }
    \label{fig:domain_adaptation_mean}
  \end{figure}

\ecoparagraph{Label-free and few-shot domain adaptation.}
Label-free adaptation uses unlabeled target trajectories to map generation signals to the source scale for Bayes Fused and to update feature standardization for logistic regression. Few-shot adaptation additionally uses labeled target trajectories to update the signal likelihood means for Bayes Fused or logistic regression coefficients (Appendix~\ref{app:adaptation}). \Cref{fig:domain_adaptation_mean} shows the results for environment and domain shifts, the most challenging setting, where supervised methods struggle to transfer. Zero labeled episodes correspond to label-free adaptation.

Under environment transfer with the same model, label-free adaptation substantially improves both methods, bringing Bayes Fused close to its best observed PRR. Additional labeled episodes from the target environment provide little benefit for Bayes Fused. Under domain transfer, label-free adaptation is insufficient, and both methods remain close to random ranking. We use only the critics shared by both domains, so this drop is not caused by missing critics, although the same check can predict success differently across environments. However, Bayes Fused benefits substantially from labeled episodes, outperforming raw UQ with 10-15 training instances. Logistic regression remains below raw UQ at this sample size.

Overall, Bayes Fused adapts effectively with little or no labeled target data, while logistic regression needs substantially more, making Bayes Fused suited to practical transfer where labels are scarce.

\subsection{Analysis and Ablations}
\label{sec:analysis}

\ecoparagraph{Aggregation of UQ scores over the trajectory.}
We compare the last step, the mean, and the maximum of raw step-level UQ scores over the trajectory (\Cref{tab:raw-aggregation-retrieval,tab:raw-aggregation-coding,tab:raw-aggregation-interactive} in Appendix~\ref{app:ablations}).
On average, the last step works best for retrieval and interactive tasks, while the mean works best for coding. In retrieval and interactive tasks, the last step reflects the final solution, whereas coding involves iterative refinement, where every step contributes to the overall uncertainty.

\ecoparagraph{Components of Bayes Fused}
include a prior, tool feedback and generation uncertainty.
We ablate the two evidence sources independently with SP in \Cref{tab:components-retrieval,tab:components-coding,tab:components-interactive} in Appendix~\ref{app:ablations}. Tool feedback is stronger on retrieval and interactive tasks, generation uncertainty on coding, and combining them improves performance in all domains.
For generation uncertainty, history\,$+$\,last is best on interactive tasks and close to the best (SEP) on retrieval, while continuous aggregation is best on coding.

\ecoparagraph{Components of Structured Bayesian Fusion.}
Removing any SBF contribution lowers the average PRR@0.5 (\Cref{tab:sbf-components} in Appendix~\ref{app:ablations}). The largest drops come from the final-step model ($-.153$ on average), which contains the agent's solution, and from the history normalization, which prevents evidence from accumulating excessively across correlated steps (\Cref{tab:sbf-normalization}). Replacing the joint final-step model by separate evidence models performs on par, so the gains come from the evidence itself rather than from joint modeling. Finally, SBF's normalization reaches the same quality as the length correction suggested by \Cref{prop:effective-history} without requiring an estimate of $\rho$ (\Cref{tab:length-correction}).

\ecoparagraph{Computational overhead.} 
\Cref{tab:runtime-cost} in Appendix~\ref{app:ablations} reports the costs of each estimator. Applied post hoc, our methods add negligible cost: Bayes Fused and Structured Bayesian Fusion add under 0.001\% to per-step inference time, except Bayes Fused (Verb.), which needs an extra LLM call.

\ecoparagraph{Calibration evaluation.} We evaluate the calibration of the proposed methods using the ECE~\citep{pmlr-v70-guo17a} metric. \Cref{tab:ece} in Appendix~\ref{app:results} shows that all learned estimators are better calibrated than raw SP. Additionally, SBF with Platt scaling~\citep{platt2000probabilistic} achieves the lowest average ECE and outperforms logistic regression on three out of five benchmarks, with ranking unchanged within each fold.

\ecoparagraph{Failure prediction at intermediate steps.} The belief update is sequential, allowing the score to be estimated at every step of the trajectory. Each estimator sees only the trajectory up to step $k$, and we test whether it predicts (1) the final outcome of the episode and (2) the state of the task at step $k$. SBF has the best average PRR@0.5 in both setups (\Cref{tab:prefix-final,tab:prefix-state} in Appendix~\ref{app:results}), with the largest margin for the intermediate state (.625 vs. .559 for the second-best method).

\ecoparagraph{Terminated trajectories.} In the main experiments, we evaluate only completed trajectories, since those terminated due to step-budget limits or harness errors are effectively failures and are easy to detect. However, when these trajectories are included as failures, SBF remains the best method on average (\Cref{tab:terminated} in Appendix~\ref{app:results}).

\section{Conclusion}
\label{sec:conclusion}

We presented a Bayesian fusion framework that combines tool-call evidence
with generation uncertainty to assess the correctness of AI agent outcomes.
The framework uses likelihood ratios to quantify how strongly tool feedback
supports task success or failure, then combines this evidence with generation
uncertainty. SBF additionally incorporates trajectory length as evidence,
normalizes accumulated evidence from intermediate steps, and models
final-step evidence separately.
The basic Bayesian fusion achieves the strongest average outcome
rankings on coding and retrieval tasks, while SBF leads on
interactive tasks. Since the belief is updated sequentially, the estimators also score trajectory prefixes, where SBF achieves the best average PRR@0.5 for predicting both the final outcome and the intermediate task state. Our method also requires fewer labeled training instances than logistic regression and maintains its performance under within-domain distribution shifts.
Label-free adaptation improves performance when transferring across
environments, while few-shot adaptation improves performance across domains
using a small amount of labeled target data.
Although no UQ method dominates every setting, these
results highlight the value of incorporating evidence from tool calls and
support Bayesian fusion as a data-efficient approach to UQ for agent outcomes.

\subsection*{AI Use Statement}

We have not used generative AI tools for any task with required disclosure under the ICLR 2027 AI Policy for Authors. We used them to edit the manuscript for readability and organization and to assist in writing code for aggregating experimental results, generating tables, and producing figures. We have reviewed all AI-assisted work. The code was checked and tested by the authors, and all text edits were reviewed by the authors. We take responsibility for the final content of this work, including text, claims or artifacts produced with the aid of generative AI.

\subsection*{Ethics Statement}

This work aims to make LLM agents more reliable by detecting likely failures, which can help decide when an agent's outcome requires additional verification or human review. However, inaccurate uncertainty estimates may lead to over-reliance on incorrect outcomes or unnecessary rejection of correct ones, so the proposed estimators should complement rather than replace human oversight in high-stakes settings. All experiments use publicly available benchmarks and models, and no human subjects or personal data are involved.

\subsection*{Reproducibility Statement}

Section~\ref{sec:method} and Appendix~\ref{app:likelihood-models} specify the proposed estimators, including their likelihood models and fitting procedures, and Appendix~\ref{app:adaptation} describes label-free and few-shot adaptation. Appendix~\ref{app:baselines} details all baselines, and Appendix~\ref{app:exp-evaluation} describes the evaluation protocol, including cross-validation splits, random seeds, and out-of-distribution transfer. Appendices~\ref{app:exp-rag}, \ref{app:exp-coding}, and~\ref{app:exp-interactive} list all environments, models, harnesses, critics, and trajectory statistics. We provide the code and scripts for reproducing all tables and figures as supplementary material.

\bibliography{iclr2027_conference}
\bibliographystyle{iclr2027_conference}

\appendix

\section{Theory and Supporting Method Details}
\label{app:method}

\subsection{Likelihood Models and Simple Fusion Variants}
\label{app:likelihood-models}

Let $c_j\in\{0,1\}$ denote the outcome of a check of type $i_j$, with $c_j=1$ indicating a pass. We model its relationship to the outcome through $\theta_{i,y} = P(c=1 \mid Y=y,i)$, where the observation unit may be an individual check or an episode summary. For each critic type and outcome class, we estimate this probability from the training trajectories using Laplace smoothing,
\begin{equation}
\widehat\theta_{i,y}
= \frac{n_{i,y}^{\mathrm{pass}}+1}{n_{i,y}+2},
\end{equation}
where $n_{i,y}$ counts observed checks of type $i$ in trajectories with final label $y$, and $n_{i,y}^{\mathrm{pass}}$ counts those that pass. Let $\widehat{\pi}\in(0,1)$ denote the prior success probability estimated from the training trajectories and write $\widehat{\ell}_0=\log[\widehat{\pi}/(1-\widehat{\pi})]$. The additive tool score is
\begin{equation}
\ell_C
= \widehat{\ell}_0+\sum_j
\left[
c_j\log\frac{\widehat\theta_{i_j,1}}{\widehat\theta_{i_j,0}}
+(1-c_j)\log\frac{1-\widehat\theta_{i_j,1}}
{1-\widehat\theta_{i_j,0}}
\right].
\label{eq:tool-only}
\end{equation}
A pass supports success when it is more frequent among successful trajectories, and its contribution is small when their ratio is close to one. A failed check can also support success if such failures occur more frequently in trajectories that eventually succeed. The sum uses a conditional independence approximation for the chosen observations. We consider implementations that include each episode summary once, implementations that sum individual check contributions across steps, and implementations that average the per-step totals over the scoring steps. Averaging produces a tempered evidence score. The observation unit and accumulation rule therefore form part of the estimator specification. In all main tables, Tool-only and Bayes Fused sum the log-likelihood ratios of the individual checks over steps when a benchmark records critics at each step, and add one log-likelihood ratio per critic when it records episode-level critics (Appendices~\ref{app:exp-rag}, \ref{app:exp-coding}, and~\ref{app:exp-interactive}). Missing checks are excluded from both the counts and the score, leaving any information carried by their absence unmodeled.

To incorporate generation uncertainty, let $u_t$ denote a scalar signal observed at each scoring step $t=1,\ldots,N$. We consider a variant based on thresholding, denoted SEP, that maps each value to one of two bins through $b_{\mathrm{bin}}:\mathbb{R}\to\{0,1\}$, using a threshold selected from the training data. Let $\widehat{p}_y(k)$ denote the probability of bin $k\in\{0,1\}$ estimated from smoothed frequencies of signal values in training trajectories with outcome $y$. The fused score is
\begin{equation}
\ell_{\mathrm{SEP}}
= \ell_C+\sum_{t=1}^{N}
\log\frac{\widehat{p}_1(b_{\mathrm{bin}}(u_t))}
{\widehat{p}_0(b_{\mathrm{bin}}(u_t))}.
\label{eq:fused-sep}
\end{equation}
The continuous variant models the signal distribution within each outcome class by a Gaussian density, with mean $\widehat{\mu}_y$ and positive variance $\widehat{\sigma}_y^2$ estimated from the training observations. The contribution of a signal value $u$ is
\begin{equation}
\begin{aligned}
G(u)
&= \log\frac{\mathcal{N}(u;\widehat{\mu}_1,\widehat{\sigma}_1^2)}
{\mathcal{N}(u;\widehat{\mu}_0,\widehat{\sigma}_0^2)}\\
&= \log\frac{\widehat{\sigma}_0}{\widehat{\sigma}_1}
+\frac{(u-\widehat{\mu}_0)^2}{2\widehat{\sigma}_0^2}
-\frac{(u-\widehat{\mu}_1)^2}{2\widehat{\sigma}_1^2},
\end{aligned}
\label{eq:gaussian-llr}
\end{equation}
giving
\begin{equation}
\ell_{\mathrm{cont}}
= \ell_C+\sum_{t=1}^{N}G(u_t).
\label{eq:fused-cont}
\end{equation}
Both variants add one contribution per scoring step to the tool score $\ell_C$, which already includes the prior. The sign of each contribution follows from the fitted likelihoods, allowing the same construction to accommodate signals whose larger values indicate either confidence or uncertainty. Discretization discards variation within bins, and continuous fusion depends on the adequacy of the Gaussian approximation. All thresholds, bin probabilities, and Gaussian parameters are estimated using training trajectories and reused unchanged during evaluation.

We also consider estimators based on the final signal value and a summary of the preceding values. The last-only variant uses the Gaussian log-likelihood ratio in \eqref{eq:gaussian-llr}, with its parameters estimated from final-step values in the training trajectories. Denoting this fitted function by $G_{\mathrm{last}}$, the score is
\begin{equation}
\ell_{\mathrm{last}}
= \ell_C+G_{\mathrm{last}}(u_N).
\label{eq:fused-last}
\end{equation}
For $N>1$, the variant based on the joint use of the history and last step summarizes the preceding values by $\bar u_{<N}=(N-1)^{-1}\sum_{t=1}^{N-1}u_t$ and fits separate Gaussian models for the history mean and the final value. Let $r\in\{H,L\}$ index these two roles. Within each role, we estimate class means $\widehat{\mu}_{r,y}$ and a common variance $\widehat{s}_r^2$ pooled within the outcome classes, using one available summary per training trajectory. The history model is fitted using only trajectories with $N>1$. Sharing the variance across classes gives the linear log-likelihood ratio
\begin{equation}
\begin{aligned}
G_r(x)
&= \frac{(x-\widehat{\mu}_{r,0})^2-(x-\widehat{\mu}_{r,1})^2}
{2\widehat{s}_r^2}\\
&= \frac{\widehat{\mu}_{r,1}-\widehat{\mu}_{r,0}}{\widehat{s}_r^2}
\left(x-\frac{\widehat{\mu}_{r,1}+\widehat{\mu}_{r,0}}{2}\right),
~r\in\{H,L\}.
\end{aligned}
\label{eq:pooled-llr}
\end{equation}
The resulting score is
\begin{equation}
\ell_{\mathrm{H+L}}
= \ell_C+G_H(\bar u_{<N})+G_L(u_N).
\label{eq:history-last}
\end{equation}
Each summary contributes once with unit weight. The history variance is estimated across trajectory means, and its contribution has no additional multiplier for the number of steps. When $N=1$, the history term is omitted. 

\subsection{Structured Bayesian Fusion Components}
\label{sec:structured-bayesian-fusion}

Structured Bayesian Fusion includes a separate contribution for trajectory length, measured by the number $N$ of scoring steps. We use the transformed length $d=\log(1+N)$ and approximate its distribution within each outcome class by a Gaussian density. From the training trajectories, we estimate class means $\widehat{\mu}_{d,y}$ and a positive variance $\widehat{s}_d^2$ pooled within the outcome classes, using one length value per trajectory. The resulting contribution is
\begin{equation}
\begin{aligned}
\Lambda_{\mathrm{len}}
&= \log\frac{\mathcal{N}(d;\widehat{\mu}_{d,1},\widehat{s}_d^2)}
{\mathcal{N}(d;\widehat{\mu}_{d,0},\widehat{s}_d^2)}\\
&= \frac{(d-\widehat{\mu}_{d,0})^2-(d-\widehat{\mu}_{d,1})^2}
{2\widehat{s}_d^2}.
\end{aligned}
\label{eq:tb-duration}
\end{equation}
This term appears once per trajectory and captures the association between its length and outcome. Its dependence on length follows from the difference between the fitted class means, allowing longer trajectories to provide evidence towards either success or failure. The association can change across task domains, agent stopping policies, and step budgets, so the predictive value of trajectory length may vary under transfer.

For critic history, let $z_t=1$ when at least one observed check at scoring step $t$ returns a failure outcome, and $z_t=0$ otherwise. The history consists of the $K=N-1$ steps preceding the final step, and $F=\sum_{t=1}^{K}z_t$ counts steps with negative critic feedback. Conditional on the outcome class and history length, we use a working model of independent Bernoulli indicators with a common probability $q_y$ of negative critic feedback for class $y$. We estimate this probability from the training histories using Laplace smoothing, $\widehat{q}_y=(n_y^{\mathrm{fail}}+1)/(n_y^{\mathrm{hist}}+2)$, where $n_y^{\mathrm{hist}}$ counts history steps in training trajectories with outcome $y$, and $n_y^{\mathrm{fail}}$ counts those with $z_t=1$. The accumulated contribution is
\begin{equation}
\Lambda_{H,C}^{\mathrm{raw}}
= F\log\frac{\widehat{q}_1}{\widehat{q}_0}
+(K-F)\log\frac{1-\widehat{q}_1}{1-\widehat{q}_0}.
\label{eq:tb-critic}
\end{equation}
This summary pools critic types and retains one indicator per step, so multiple failure outcomes at the same step are counted once. Steps with no observed checks and steps with only passing checks both have $z_t=0$ and contribute through the $K-F$ term. Thus, zero represents the absence of negative critic feedback, which can include missing feedback. When $K=0$, the history contribution is zero. The superscript $\mathrm{raw}$ identifies the accumulated evidence before the history normalization described later in this subsection.

For generation history, let $\mathbf{x}_t$ collect the available uncertainty signals at step $t$, and let $\widetilde{\mathbf{x}}_t$ denote their standardized and clipped values, using preprocessing parameters learned from the training trajectories. We form a scalar projection $h_t=\widehat{\mathbf{w}}_H^\top\widetilde{\mathbf{x}}_t$, where $\widehat{\mathbf{w}}_H$ is fitted using logistic regression with ridge regularization and the trajectory outcomes as labels. For $K>0$, we use a working model in which the projected history values are independent Gaussian observations conditional on the outcome class and history length. We estimate their class means $\widehat{\mu}_{h,y}$ and a positive variance $\widehat{s}_h^2$ pooled within the outcome classes from projected training history steps. Under this model, the mean $\bar h_{<N}=K^{-1}\sum_{t=1}^{K}h_t$ has variance $\widehat{s}_h^2/K$, giving the accumulated contribution
\begin{equation}
\begin{aligned}
\Lambda_{H,U}^{\mathrm{raw}}
&= \log\frac{\mathcal{N}(\bar{h}_{<N};\widehat{\mu}_{h,1},\widehat{s}_h^2/K)}
{\mathcal{N}(\bar{h}_{<N};\widehat{\mu}_{h,0},\widehat{s}_h^2/K)}\\
&= K\frac{(\bar{h}_{<N}-\widehat{\mu}_{h,0})^2-(\bar{h}_{<N}-\widehat{\mu}_{h,1})^2}
{2\widehat{s}_h^2},
\end{aligned}
\label{eq:tb-history}
\end{equation}
where $\widehat{s}_h^2$ describes variation among projected step values within each outcome class. The history variance $\widehat{s}_H^2$ in \eqref{eq:pooled-llr} instead describes variation among trajectory means of the selected signal. The factor $K$ follows from the working independence assumption and determines how strongly the history mean contributes to the score. Dependence between steps can reduce the information provided by additional observations, motivating the history normalization described later in this subsection. When $K=0$, the history contribution is zero and its mean is not evaluated.

For the final step, we form a separate projection $r_N=\widehat{\mathbf{w}}_{\mathrm{last}}^\top\widetilde{\mathbf{x}}_N$, with coefficients $\widehat{\mathbf{w}}_{\mathrm{last}}$ fitted by logistic regression with ridge regularization using final-step training observations and trajectory outcomes. We combine this projection with the indicator $z_N$ of negative critic feedback at the final step to form $\mathbf{v}_N=(r_N,z_N)^\top$. We fit a bivariate Gaussian model with class mean vectors $\widehat{\boldsymbol{\mu}}_{\mathrm{last},y}$ and a shared positive definite covariance matrix $\widehat{\Sigma}_{\mathrm{last}}$ estimated from variation within the outcome classes. Under this model, the final-step contribution is
\begin{equation}
\Lambda_{\mathrm{last}}
= \left(\widehat{\boldsymbol{\mu}}_{\mathrm{last},1}-\widehat{\boldsymbol{\mu}}_{\mathrm{last},0}\right)^\top
\widehat{\Sigma}_{\mathrm{last}}^{-1}
\left[
\mathbf{v}_N
-\frac{\widehat{\boldsymbol{\mu}}_{\mathrm{last},1}+\widehat{\boldsymbol{\mu}}_{\mathrm{last},0}}{2}
\right].
\label{eq:tb-final}
\end{equation}
The covariance allows the projected uncertainty signal and the critic indicator to be correlated within outcome classes. Because $z_N$ is binary, the Gaussian model serves as a discriminant approximation for this combination of continuous and binary evidence. The contribution appears once per trajectory, including when $N=1$ and both history channels are empty.

We normalize the two history channels separately to adjust the scale and location of their accumulated evidence. For channel $r\in\{C,U\}$, corresponding to critic and generation history, respectively, we use
\begin{equation}
\Lambda_{H,r}
=
\begin{cases}
\widehat{\beta}_{H,r}\bigl(\Lambda_{H,r}^{\mathrm{raw}}-\widehat{a}_{H,r}\bigr), & K>0,\\
0, & K=0,
\end{cases}
\label{eq:tb-normalization}
\end{equation}
where $\widehat{\beta}_{H,r}$ is the fitted slope and $\widehat{a}_{H,r}$ is a centering parameter on the raw score scale. These parameters are estimated separately for each channel using out-of-fold raw scores and the corresponding trajectory outcomes within the training data. This adjustment can temper excessive accumulation when dependence between steps causes the likelihood model to overstate the information in a history. The same fitted transformation applies across positive history lengths, so it provides a common affine adjustment whose adequacy may vary with length. For an empty history, the entire contribution is zero, including the offset. We use $\Lambda_{H,C}$ and $\Lambda_{H,U}$ as adjusted evidence contributions in the combined score. Normalizing them separately does not ensure independence between channels or calibration of the resulting probability.

\subsection{Label-Free and Few-Shot Adaptation}
\label{app:adaptation}

The adaptation experiments extend the source-trained estimators using observations from a target configuration. We describe the transformations used for Bayes Fused and logistic regression.

\ecoparagraph{Label-free adaptation.}
Let $\widehat{\mu}_S$ and $\widehat{s}_S$ denote the mean and standard deviation of a scalar generation signal computed from source training trajectories, and let $\widehat{\mu}_T$ and $\widehat{s}_T$ denote the corresponding quantities computed from unlabeled target trajectories. For Bayes Fused, a target signal value $u_t$ is mapped to the source scale through
\[
\widetilde{u}_t
=\widehat{\mu}_S
+\frac{\widehat{s}_S}{\widehat{s}_T}
(u_t-\widehat{\mu}_T),
\qquad \widehat{s}_T>0.
\]
This transformation matches the target signal's empirical mean and standard deviation to those of the source reference. For logistic regression, input features are standardized using target statistics in place of source statistics. The source likelihood parameters and regression coefficients remain unchanged.

\ecoparagraph{Few-shot adaptation.}
Using labeled target trajectories, Bayes Fused additionally updates the class means of its signal likelihood while retaining the other source parameters. Let $\widehat{\mu}_y$ denote a source class mean. For each target trajectory, average its rescaled signal values over its scoring steps. Within outcome class $y$, let $n_y$ be the number of labeled target trajectories, and let $\bar u_{T,y}$ and $s_{T,y}^2$ be the sample mean and sample variance of these trajectory averages. For $n_y\geq2$, the update is
\[
\begin{aligned}
\widehat{\mu}_y'
&=\widehat{\mu}_y
+\alpha_y(\bar u_{T,y}-\widehat{\mu}_y),\\
\widehat{\tau}_y^2
&=\max\left\{0,
(\bar u_{T,y}-\widehat{\mu}_y)^2
-\frac{s_{T,y}^2}{n_y}
\right\},\\
\alpha_y
&=\frac{\widehat{\tau}_y^2}
{\widehat{\tau}_y^2+s_{T,y}^2/n_y},
\end{aligned}
\]
whenever the denominator defining $\alpha_y$ is positive. The quantity $\widehat{\tau}_y^2$ is a non-negative estimate of the squared discrepancy between the target and source means after subtracting an estimate of the target mean's sampling variance. It determines how far the update moves towards the target estimate. For logistic regression, adaptation refits the coefficients using labeled target trajectories with an $L_2$ penalty on their deviation from the source coefficients.

\subsection{Proof and Consequences of~\texorpdfstring{\Cref{prop:effective-history}}{Proposition~1}}
\label{app:effective-history-proof}

We examine how dependence between steps affects the evidence provided by the scalar generation history $\mathbf{h}=(h_1,\ldots,h_K)^\top$, where $K=N-1$. For each positive history length with $P(K\mid Y=y)>0$ for both $y\in\{0,1\}$, assume
\[
\begin{aligned}
\mathbf{h}\mid(Y=y,K)
&\sim \mathcal{N}(\mu_y\mathbf{1}_K,\Sigma_K),\\
\Sigma_K
&=\sigma^2 [(1-\rho)I_K+\rho\mathbf{1}_K\mathbf{1}_K^\top ],
\end{aligned}
\]
where $\mathbf{1}_K$ is the vector of $K$ ones and $I_K$ is the $K\times K$ identity matrix. Each history value has class mean $\mu_y$ and variance $\sigma^2>0$, and every pair of different history values has correlation $\rho\in[0,1)$ conditional on the outcome and length. The covariance is shared across outcome classes, and $\mu_0,\mu_1,\sigma^2,\rho$ do not depend on $K$. The parameters are treated as population quantities, and the calculation therefore ignores uncertainty arising from their estimation. Write $\Delta=\mu_1-\mu_0$, $m=(\mu_1+\mu_0)/2$, and $\bar h=K^{-1}\sum_{t=1}^{K}h_t$, so that $\bar h$ is the history mean denoted by $\bar h_{<N}$ in the estimator. The Gaussian assumption concerns the distribution conditional on the realized history length. When length is determined by an adaptive stopping rule, the assumption must hold after conditioning on the resulting length.

\begin{proof}[Proof of~\Cref{prop:effective-history}]
Consider any positive history length $K$ in the support of both outcome classes. For every nonzero $\mathbf{v}\in\mathbb{R}^K$,
\[
\mathbf{v}^\top\Sigma_K\mathbf{v}
=\sigma^2(1-\rho)\|\mathbf{v}\|_2^2
+\sigma^2\rho(\mathbf{1}_K^\top\mathbf{v})^2
>0,
\]
since $\sigma^2>0$ and $0\leq\rho<1$. Thus, $\Sigma_K$ is positive definite, and both conditional Gaussian distributions have positive densities on $\mathbb{R}^K$. Their shared covariance makes the normalizing factors cancel in the density ratio. Expanding the quadratic forms gives
\[
\begin{aligned}
\Lambda_{H,U}(\mathbf{h}\mid K)
&:=\log\frac{p(\mathbf{h}\mid Y=1,K)}
{p(\mathbf{h}\mid Y=0,K)}\\
&=\frac{1}{2}(\mathbf{h}-\mu_0\mathbf{1}_K)^\top
\Sigma_K^{-1}(\mathbf{h}-\mu_0\mathbf{1}_K)\\
&\quad-\frac{1}{2}(\mathbf{h}-\mu_1\mathbf{1}_K)^\top
\Sigma_K^{-1}(\mathbf{h}-\mu_1\mathbf{1}_K)\\
&=\Delta\,\mathbf{1}_K^\top\Sigma_K^{-1}
(\mathbf{h}-m\mathbf{1}_K).
\end{aligned}
\]

The covariance satisfies
\[
\Sigma_K\mathbf{1}_K
=\sigma^2[1+(K-1)\rho]\mathbf{1}_K,
\]
and hence
\[
\Sigma_K^{-1}\mathbf{1}_K
=\frac{\mathbf{1}_K}{\sigma^2[1+(K-1)\rho]}.
\]
Using the symmetry of $\Sigma_K^{-1}$ in the preceding expression for the log-likelihood ratio yields
\[
\begin{aligned}
\Lambda_{H,U}(\mathbf{h}\mid K)
&=\frac{\Delta}{\sigma^2[1+(K-1)\rho]}
\mathbf{1}_K^\top(\mathbf{h}-m\mathbf{1}_K)\\
&=\frac{K}{1+(K-1)\rho}
\frac{\Delta}{\sigma^2}(\bar h-m)\\
&=K_{\mathrm{eff}}\frac{\mu_1-\mu_0}{\sigma^2}
\left(\bar h-\frac{\mu_1+\mu_0}{2}\right),
\end{aligned}
\]
where $K_{\mathrm{eff}}=K/[1+(K-1)\rho]$. Since $\Lambda_{\mathrm{ind}}(\mathbf{h}\mid K)=K\Delta(\bar h-m)/\sigma^2$, this establishes
\[
\Lambda_{H,U}(\mathbf{h}\mid K)
=\frac{\Lambda_{\mathrm{ind}}(\mathbf{h}\mid K)}
{1+(K-1)\rho}.
\]

For the second claim, suppose $\Delta\neq0$, $\rho>0$, and two different positive lengths belong to the support of both outcome classes. If constants $\beta,\gamma$ satisfy $\beta\Lambda_{\mathrm{ind}}(\mathbf{h}\mid K)+\gamma=\Lambda_{H,U}(\mathbf{h}\mid K)$ for every history vector at both lengths, then at each of these lengths,
\[
\beta\,\frac{K\Delta}{\sigma^2}(\bar h-m)+\gamma
=\frac{K\Delta}{\sigma^2[1+(K-1)\rho]}(\bar h-m)
\]
for every $\bar h\in\mathbb{R}$. Indeed, the conditional Gaussian distributions have support $\mathbb{R}^K$, and every real history mean is attained by some vector in that support. Setting $\bar h=m$ gives $\gamma=0$. Equating the coefficients of $\bar h$ and using $K\Delta\neq0$ then gives
\[
\beta=\frac{1}{1+(K-1)\rho}.
\]
The right-hand side takes different values at the two lengths because $\rho>0$. Thus, no common pair $(\beta,\gamma)$ satisfies the required equality, proving the second claim.
\end{proof}

The effective history size $K_{\mathrm{eff}}$ describes the precision of the history mean under the assumed model, since
\[
\operatorname{Var}(\bar h\mid Y=y,K)
=\frac{\mathbf{1}_K^\top\Sigma_K\mathbf{1}_K}{K^2}
=\sigma^2\left(\rho+\frac{1-\rho}{K}\right)
=\frac{\sigma^2}{K_{\mathrm{eff}}}.
\]
Thus, $K_{\mathrm{eff}}$ is the equivalent number of independent observations with variance $\sigma^2$ in terms of the variance of their mean. It equals $K$ when $\rho=0$ and equals one when $K=1$. For any $\rho\in(0,1)$, it increases towards $1/\rho$ as $K$ grows, while the variance of the mean approaches $\sigma^2\rho$. Additional steps therefore provide diminishing gains in precision under this dependence model. Length can also carry evidence about the outcome through differences in its distribution between classes. For the evidence pair $(K,\mathbf{h})$, Bayes' rule and the probability chain rule give
\[
\begin{aligned}
\ell^\star(K,\mathbf{h})
&:=\log\frac{P(Y=1\mid K,\mathbf{h})}
{P(Y=0\mid K,\mathbf{h})}\\
&=\ell_0+\log\frac{P(K\mid Y=1)}{P(K\mid Y=0)}
+\Lambda_{H,U}(\mathbf{h}\mid K).
\end{aligned}
\]
This decomposition requires no independence assumption between length and history values. The length ratio describes the association between trajectory length and outcome, while the conditional history ratio measures the evidence in the observed values at that length. Since $N=K+1$, either length variable carries the same information. The fitted length and history components of Structured Bayesian Fusion serve these two roles, and their agreement with the population terms depends on the adequacy of their models and estimation.

The affine limitation applies to the form of the generation-history normalization in \eqref{eq:tb-normalization}. For positive history lengths, that transformation can be written as $\beta\Lambda_{H,U}^{\mathrm{raw}}+\gamma$, with $\beta=\widehat{\beta}_{H,U}$ and $\gamma=-\widehat{\beta}_{H,U}\widehat{a}_{H,U}$. Even when the raw score uses the population means and variance, unequal class means and positive correlation prevent a common pair of coefficients from recovering the conditional history likelihood ratio across different lengths under the assumed model. The required rescaling varies with length, and a common offset cannot compensate for that variation. This limitation concerns exact recovery of the generation-history likelihood ratio. The usefulness of fitted affine normalization for outcome ranking and probability calibration remains an empirical question.

\section{Additional Details of Experimental Setup}
\label{app:experiments}

\subsection{Detailed Description of Baselines}
\label{app:baselines}

We compare the Bayesian methods from Section~\ref{sec:method} with trajectory length, tool success rate, five generation-level signals, three trajectory-level UQ methods, and logistic regression. As in Section~\ref{sec:problem}, $N$ denotes the number of scoring steps of a trajectory $\tau$, with action $a_t$ and feedback $o_t$ at step $t$. Let $n_t$ be the number of generated tokens at step $t$. For the $j$-th generated token of step $t$, $j=1,\dots,n_t$, let $p_{tj}(v)$ denote the estimated probability of vocabulary token $v\in V$, and $p_{tj}$ the probability of the generated token. For raw generation-level UQ methods, we consider three aggregation techniques: the average and the maximum over the trajectory, and the value at the last step alone. All learnable baselines are trained on the source training folds only, with the binary outcome $Y$ as the label.

\ecoparagraph{N steps.}
Shorter trajectories indicate simpler tasks for the agent and therefore receive higher confidence:
\[
    \uv_{N_{\mathrm{steps}}}=-N.
\]

\ecoparagraph{Tool success rate.}
Fraction of passed tool checks among the $M$ observed checks $c_j\in\{0,1\}$ in the trajectory, with missing tools skipped:
\[
    \uv_{\mathrm{TSR}}=\frac{1}{M}\sum_{j=1}^{M}c_j.
\]

\ecoparagraph{Sequence probability (SP).}
Negative sum of the token-level log-probabilities of the generated sequence:
\[
    \uv_{\mathrm{SP}}=-\sum_{j=1}^{n_t}\log p_{tj}.
\]

\ecoparagraph{Perplexity.}
Negative average token-level log-probability, commonly referred to as perplexity~\citep{fomicheva-etal-2020-unsupervised}:
\[
    \uv_{\mathrm{PPL}}=
    \exp\left(-\frac{1}{n_t}
    \sum_{j=1}^{n_t}\log p_{tj}\right).
\]

\ecoparagraph{Mean token entropy (MTE).}
Average entropy of the token distributions~\citep{fomicheva-etal-2020-unsupervised}. We approximate each distribution by its top-$k$ tokens ($k=20$), renormalized to sum to one:
\[
    \uv_{\mathrm{MTE}}=-\frac{1}{n_t}
    \sum_{j=1}^{n_t}\sum_{v\in V} p_{tj}(v)\log p_{tj}(v).
\]

\ecoparagraph{Self-certainty.}
Self-certainty~\citep{kang2025scalable} measures the KL divergence from the uniform distribution $\mathcal U$ over the top-$k$ tokens. We approximate it with raw top-$k$ log-probabilities and reverse it to obtain an uncertainty score:
\[
\uv_{\mathrm{SC}}
= \frac{1}{n_t}\sum_{j=1}^{n_t}
\mathrm{KL}(\mathcal U|p_{tj})
= \frac{1}{n_t}\sum_{j=1}^{n_t}
\left(
-\frac{1}{k}\sum_{v\in\mathrm{top}-k}\log p_{tj}(v)
- \log k
\right).
\]

\ecoparagraph{Verbalized confidence (Verb.).}
We additionally explicitly ask the agent to report its confidence at each step and use its response as the uncertainty estimate~\citep{tian2023calibration}:
\[
    \uv_{\mathrm{Verb.}}= 1 - \widehat p_t^{\,\mathrm{verb.}}\
\]

\ecoparagraph{Situation-Aware Uncertainty Propagation (SAUP).}
SAUP~\citep{zhao2025saup} propagates step uncertainty with larger weight on late steps and on steps that differs from the task context. Let $\uv_t$ be the MTE of step $t$ and $w_t$ its situational weight:
\[
    \uv_{\mathrm{SAUP}}(\tau)
    =
    \sqrt{\frac{1}{N}\sum_{t=1}^{N}(w_t\uv_t)^2}.
\]
We consider five choices of the weights $w_t$, following the variants of the original method: uniform, positional ($t/N$), based on embedding distances between the task, the step and its observation, their sum, or predicted by a Gaussian HMM over those distances. We use the HMM variant without labels, since the original work relies on manually annotated states that are unavailable for our trajectories.

\ecoparagraph{Uncertainty Propagation (UProp).}
UProp~\citep{duan2025uprop} splits the uncertainty of each step into the intrinsic uncertainty of the current decision and the extrinsic uncertainty inherited from earlier decisions. At each step $t$, we resample $S=10$ alternative actions $a_t^{(1)},\dots,a_t^{(S)}$ from the same model and prompt $h_t$. Let $|a|$ denote the number of tokens in action $a$, $\mathrm{dist}$ a string distance, and $\kappa$ a Gaussian kernel. The intrinsic uncertainty (IU) is the mean length-normalized negative log-likelihood of the sampled actions, and the pointwise mutual information (PMI) measures how unusual the action $a_t$ is among sampled alternatives:
\[
    \mathrm{IU}_t
    =
    -\frac{1}{S}\sum_{s=1}^{S}\frac{1}{|a_t^{(s)}|}\log p\big(a_t^{(s)}\mid h_t\big),
    \qquad
    \mathrm{PMI}_t
    =
    -\log\frac{1}{S}\sum_{s=1}^{S}\kappa\big(\mathrm{dist}(a_t^{(s)},a_t)\big).
\]
Finally, the extrinsic uncertainty (EU) accumulates the PMI of earlier steps, $\mathrm{EU}_t=\sum_{t'<t}\mathrm{PMI}_{t'}$. Given the step uncertainty $H_t=\mathrm{IU}_t+\mathrm{EU}_t$, the trajectory uncertainty is
\[
    \uv_{\mathrm{UProp}}(\tau)=
    \frac{\sum_{t=1}^{N}H_t}{N+\sum_{t=1}^{N}\mathrm{EU}_t/\mathrm{IU}_t}.
\]

\ecoparagraph{Holistic Trajectory Calibration (HTC).}
HTC~\citep{zhang2026confidence} is a supervised calibrator that fits a logistic regression on 48 trajectory-level features of the per-token confidence trace, where token confidence is the probability of the top-1 token and the mean probability of the top-$k$ tokens. The features are grouped into dynamics (how confidence changes across steps), position (confidence statistics at the first and last steps), stability (variability of confidence within and across steps), and structure (step count and tokens per step). Following the original work, we report HTC-Reduced, which fits the regression with an $L_1$ penalty.

\ecoparagraph{LLM-as-judge.} At the final step, a judge model (GPT-OSS-20B) receives the task, the agent's final response, and the last 10 steps of its trajectory, and predicts the probability that the agent successfully completed the task. Unlike the other baselines, it requires an additional LLM call. We evaluate it on ALFWorld and DeepSWE using the same protocol as the other methods.

\ecoparagraph{Logistic regression.}
Additionally, we train an $L_2$-penalized logistic regression that predicts task success from trajectory-level features. The candidate features are aggregates of each generation-level signal over steps (e.g., first, last, mean, maximum, and variability), trajectory-level tool outcomes, and per-step pass rates. A small subset of features is selected greedily on the training folds. 

\subsection{Detailed Description of PRR}
\label{app:prr}

The Prediction Rejection Ratio (PRR)~\citep{malinin2021uncertainty,vashurin2025benchmarking} measures how effectively an uncertainty score ranks agent trajectories by the quality of their solutions. It is computed from the rejection curve, which shows the average quality of the remaining trajectories as an increasing fraction of the most uncertain trajectories is rejected. PRR normalizes the area under this curve for the evaluated method with the areas for random and oracle rankings:
\[
    \mathrm{PRR}=\frac{\mathrm{AUC}_{\mathrm{method}}-\mathrm{AUC}_{\mathrm{random}}}{\mathrm{AUC}_{\mathrm{oracle}}-\mathrm{AUC}_{\mathrm{random}}}.
\]
PRR is upper-bounded by one, which corresponds to the oracle ranking. A value of zero corresponds to random ranking, and negative values indicate performance below random. Following prior work~\citep{vashurin2025benchmarking}, we compute PRR only over the first 50\% of the rejection curve (PRR@0.5), since rejecting more than half of the trajectories is impractical.

PRR is analogous to ROC-AUC, but unlike it, applies to both binary and continuous outcome measures. As the final solution quality, we use the native score of each benchmark:
\begin{compactitem}
    \item \textbf{Retrieval:} the judge's verdict on the final answer;
    \item \textbf{Coding:} whether the final solution passes all hidden tests; for DeepSWE, the fraction of passed hidden tests;
    \item \textbf{$\tau^3$-bench:} whether the benchmark reward equals one;
    \item \textbf{WebArena-Verified, OSWorld-Verified, and ALFWorld} \citep{hattami2025webarena, xie2024osworld, shridhar2021alfworld}: the benchmark evaluator's score, including partial credit on OSWorld-Verified.
\end{compactitem}

\subsection{Evaluation Protocol}
\label{app:exp-evaluation}

\ecoparagraph{Cross-validation.}
For each configuration, we split the episodes into five stratified folds. Folds are stratified by the binary outcome. Each estimator is fitted on four folds and applied to the held-out fold. We pool the held-out predictions of all five folds and compute a single PRR@0.5 over the pooled set, so every episode is scored exactly once by a model that has not seen it.

\ecoparagraph{Random seeds and statistical robustness.}
We repeat the fold splitting with three seeds, $s\in\{0,1,2\}$. For each configuration, we compute PRR@0.5 for every seed and report the mean and standard deviation $\sigma_i$ over seeds. Unsupervised methods, such as raw generation-level signals, do not depend on the fold split and therefore have zero standard deviation. The small standard deviations observed for some of these methods arise from random tie-breaking between equal scores. When a table column averages $n$ configurations, we report the mean of their PRR@0.5 values, and its standard deviation is obtained by propagating the per-configuration standard deviations, treating configurations as independent with $\sigma_{\mathrm{avg}}=\frac{1}{n}\sqrt{\sum_{i=1}^{n}\sigma_i^2}$. Across benchmarks, the task sets are essentially independent. Within a benchmark, we treat configurations as independent only to keep the SD pipeline uniform; even under perfect correlation between configurations, the SDs stay of the same order of magnitude. To address task-sampling uncertainty, we additionally compute paired task-level bootstrap intervals, which account for tasks shared across configurations (\Cref{tab:bootstrap} in Appendix~\ref{app:results}).

\ecoparagraph{Trajectory-level aggregation.}
Aggregation techniques vary by the task, as specified in the captions of \Cref{tab:prr-dataset-means-rag,tab:prr-dataset-means-coding,tab:prr-dataset-means-other}. For coding, Bayes Fused uses continuous aggregation, while raw UQ scores are averaged over the trajectory. For retrieval-based and interactive tasks, Bayes Fused uses history + last aggregation, while raw UQ uses only the last-step score. These aggregations are supported by the task formulation itself. Coding tasks require iterative refinement, where each step matters equally, while for other tasks, only the last step provides the final solution and should therefore be treated differently. A detailed analysis of different aggregation methods for Bayes Fused and UQ is presented in Appendix~\ref{app:ablations}.

\ecoparagraph{Out-of-distribution evaluation.}
For the out-of-distribution experiments, all parameters of an estimator, including likelihoods, thresholds, standardization statistics, and regression weights, are fitted on the four training folds of the source configuration. The fitted estimator is then applied unchanged to the corresponding held-out fold of the target configuration, and predictions are pooled over the five folds as in the in-distribution setting. Folds are assigned per task and shared by all agents on a benchmark, stratified by the joint vector of the configurations' binary outcomes on each task. The held-out fold of the target configuration therefore contains exactly the tasks excluded from source training, so no task-level information leaks between training and testing. We use the same three random seeds and, for reference, also report the in-distribution PRR@0.5 of the target configuration.

\ecoparagraph{Data filtering.}
We evaluate only episodes that the agent completed itself, excluding those stopped by the step budget, the context window, or an environment failure. Such externally terminated episodes are failures by construction, so estimating their uncertainty is uninformative. \Cref{tab:terminated} in Appendix~\ref{app:results} reports the results when these episodes are included as failures.

\ecoparagraph{Label binarization.}
For tasks with continuous scores (e.g., OSWorld and DeepSWE), fold stratification and training require a binary label. For OSWorld, every episode with a score below 1 is considered a failure. For DeepSWE, no task is fully solved, so we binarize the test-pass fraction at the median: the median of the whole cohort for fold stratification and the median of the training folds of each split for training, which gives balanced training classes. The held-out fold is therefore never used to set the threshold for training. At test time, PRR is computed on the raw continuous test-pass fraction.

\subsection{Retrieval-Based Environments}
\label{app:exp-rag}

\begin{table}[ht]
\centering
\begingroup
\renewcommand{\arraystretch}{0.92}%
\caption{\textbf{Retrieval task statistics.} For each model and harness configuration, \emph{Episodes} reports the total number of scored episodes and the completed episodes retained for evaluation. \emph{Steps} reports the maximum and median number of scored steps per episode. \emph{Success rate} is the fraction of successful episodes.}
\label{tab:retrieval-stats}
\resizebox{\linewidth}{!}{%
\begin{tabular}{lllrrrrr}
\toprule
\multirow{2}{*}{\textbf{Benchmark}} & \multirow{2}{*}{\textbf{Model}} & \multirow{2}{*}{\textbf{Harness}}
  & \multicolumn{2}{c}{\textbf{Episodes}} & \multicolumn{2}{c}{\textbf{Steps}} & \multirow{2}{*}{\textbf{Success rate}} \\
\cmidrule(lr){4-5}\cmidrule(lr){6-7}
 & & & all & completed & max & median & \\
\midrule
\multirow{6}{*}{HotpotQA}
  & \multirow{2}{*}{Qwen2.5-32B-Instruct} & smolagents CodeAgent & 700 & 700 & 13 & 2 & 0.836 \\
  &                              & LangGraph ReAct      & 700 & 700 & 7  & 1 & 0.737 \\
  & \multirow{2}{*}{GPT-OSS-20B} & smolagents CodeAgent & 700 & 646 & 14 & 1 & 0.804 \\
  &                              & LangGraph ReAct      & 700 & 700 & 12 & 2 & 0.824 \\
  & \multirow{2}{*}{Qwen3.6-35B-A3B} & smolagents CodeAgent & 700 & 669 & 36 & 2 & 0.891 \\
  &                              & LangGraph ReAct      & 700 & 700 & 20 & 2 & 0.876 \\
\midrule
\multirow{6}{*}{MuSiQue}
  & \multirow{2}{*}{Qwen2.5-32B-Instruct} & smolagents CodeAgent & 300 & 299 & 16 & 3 & 0.550 \\
  &                              & LangGraph ReAct      & 300 & 300 & 12 & 2 & 0.450 \\
  & \multirow{2}{*}{GPT-OSS-20B} & smolagents CodeAgent & 300 & 235 & 17 & 2 & 0.477 \\
  &                              & LangGraph ReAct      & 300 & 300 & 12 & 3 & 0.507 \\
  & \multirow{2}{*}{Qwen3.6-35B-A3B} & smolagents CodeAgent & 300 & 213 & 36 & 6 & 0.680 \\
  &                              & LangGraph ReAct      & 300 & 300 & 24 & 5 & 0.600 \\
\midrule
\multirow{6}{*}{CRAG}
  & \multirow{2}{*}{Qwen2.5-32B-Instruct} & smolagents CodeAgent & 300 & 296 & 8004 & 4 & 0.323 \\
  &                              & LangGraph ReAct      & 300 & 297 & 13   & 3 & 0.263 \\
  & \multirow{2}{*}{GPT-OSS-20B} & smolagents CodeAgent & 300 & 249 & 32   & 2 & 0.357 \\
  &                              & LangGraph ReAct      & 300 & 198 & 13   & 6 & 0.313 \\
  & \multirow{2}{*}{Qwen3.6-35B-A3B} & smolagents CodeAgent & 300 & 264 & 602  & 6 & 0.543 \\
  &                              & LangGraph ReAct      & 300 & 237 & 13   & 8 & 0.460 \\
\midrule
\multirow{2}{*}{BrowseComp-Plus}
  & \multirow{2}{*}{Qwen3.6-35B-A3B} & LangGraph ReAct        & 830 & 696 & 55 & 23   & 0.391 \\
  &                              & smolagents ToolCalling & 830 & 734 & 51 & 27.5 & 0.347 \\
\bottomrule
\end{tabular}}
\endgroup
\end{table}

\begin{table}[ht]
\centering
\begingroup
\renewcommand{\arraystretch}{0.92}%
\caption{\textbf{Retrieval task critics.} Definitions of the critics used to assess retrieval actions and their outcomes.}
\label{tab:retrieval-critics}
\resizebox{\linewidth}{!}{%
\begin{tabular}{lll}
\toprule
\textbf{Benchmark} & \textbf{Critic} & \textbf{Description} \\
\midrule
\multirow{2}{*}{HotpotQA, MuSiQue}
  & \texttt{retrieved\_any} & the search returned at least one document \\
  & \texttt{retrieved\_new} & the search returned a document not retrieved earlier \\
\midrule
\multirow{3}{*}{CRAG}
  & \texttt{tool\_ok}               & the tool call executed without an error \\
  & \texttt{tool\_returned\_content} & the tool returned non-empty content \\
  & \texttt{query\_is\_new}          & the query differs from all earlier queries \\
\midrule
\multirow{4}{*}{BrowseComp-Plus}
  & \texttt{action\_format\_valid}  & all actions in the episode are well-formed \\
  & \texttt{ui\_execution\_success} & all tool calls in the episode executed successfully \\
  & \texttt{state\_change}          & at least one tool call returned new text \\
  & \texttt{no\_stagnation}         & the agent does not end in a repeated cycle of actions \\
\bottomrule
\end{tabular}}
\endgroup
\end{table}

This subsection describes the HotpotQA, MuSiQue, CRAG, and BrowseComp-Plus environments in detail.

\ecoparagraph{Models and harnesses.}
We generate trajectories by Qwen2.5-32B-Instruct~\citep{qwen2024qwen}, GPT-OSS-20B~\citep{openai2025gptossb}, and Qwen3.6-35B-A3B~\citep{qwen36plus}, each served locally through vLLM~\citep{kwon2023efficient}. On HotpotQA, MuSiQue, and CRAG, each model runs with two harnesses: smolagents CodeAgent~\citep{smolagents}, which writes Python code that calls the search tool, and LangGraph ReAct~\citep{yao2023react,langgraph}, which includes free-text thoughts followed by actions. BrowseComp-Plus is run with Qwen3.6-35B-A3B~\citep{qwen36plus} only, under LangGraph ReAct~\citep{yao2023react,langgraph} and smolagents ToolCallingAgent~\citep{smolagents}, since the other models achieve near-zero success on this benchmark. For the reasoning models, verbalized confidence is elicited with disabled thinking. We evaluate only episodes that the agent completed itself. \Cref{tab:retrieval-stats} presents all 20 configurations with their numbers of episodes, step statistics, and success rates.

\ecoparagraph{Critics.} Critics differ across benchmarks and are listed in \Cref{tab:retrieval-critics}. A critic is a boolean check applied at each step by the harness itself, without additional model calls. Critics check the mechanics of a step, such as whether a tool was executed and returned content, and never inspect the answer itself. Therefore, a step that retrieves irrelevant documents can still pass, so the informativeness of critics varies across benchmarks. Since critics are recorded at each step, Tool-only and Bayes Fused sum their log-likelihood ratios over steps.

\ecoparagraph{Hardware configuration.}
We use four NVIDIA A100-SXM4-80GB GPUs for HotpotQA, MuSiQue, and CRAG, and two NVIDIA RTX PRO 6000 Blackwell GPUs for BrowseComp-Plus.

\subsection{Coding Environments}
\label{app:exp-coding}

\begin{table}[ht]
\centering
\begingroup
\renewcommand{\arraystretch}{0.92}%
\caption{\textbf{Coding task statistics.} For each model and harness configuration, \emph{Episodes} reports the total and completed episode counts. \emph{Steps} reports the maximum and median number of scored steps per episode. \emph{Success rate} is the fraction of completed episodes passing all hidden tests, except for DeepSWE, where it is the mean fraction of hidden tests passed. Episodes with no candidate solution because every generation ended within the reasoning block are excluded.}
    \label{tab:coding-stats}
    \resizebox{\linewidth}{!}{%
    \begin{tabular}{lllrrrrr}
    \toprule
    \multirow{2}{*}{\textbf{Benchmark}} & \multirow{2}{*}{\textbf{Model}} & \multirow{2}{*}{\textbf{Harness}}
      & \multicolumn{2}{c}{\textbf{Episodes}} & \multicolumn{2}{c}{\textbf{Steps}} & \multirow{2}{*}{\textbf{Success rate}} \\
    \cmidrule(lr){4-5}\cmidrule(lr){6-7}
     & & & all & completed & max & median & \\
    \midrule
    \multirow{8}{*}{LiveCodeBench-Hard}
      & \multirow{2}{*}{GPT-OSS-20B}  & smolagents CodeAgent & 350 & 309 & 13 & 7 & 0.49 \\
      &                               & ReAct                & 350 & 176 & 18 & 10 & 0.56 \\
      & \multirow{2}{*}{GPT-OSS-120B} & smolagents CodeAgent & 350 & 304 & 23 & 13 & 0.65 \\
      &                               & ReAct                & 350 & 237 & 20 & 13 & 0.61 \\
      & \multirow{2}{*}{Qwen3.6-35B-A3B}  & smolagents CodeAgent & 350 & 285 & 23 & 16 & 0.71 \\
      &                               & ReAct                & 350 & 288 & 23 & 17 & 0.73 \\
      & \multirow{2}{*}{Qwen3.8-27B}  & smolagents CodeAgent & 348 & 314 & 23 & 15 & 0.77 \\
      &                               & ReAct                & 350 & 322 & 23 & 15 & 0.74 \\
    \midrule
    \multirow{8}{*}{BigCodeBench-Hard}
      & \multirow{2}{*}{GPT-OSS-20B}  & smolagents CodeAgent & 117 & 102 & 11 & 7 & 0.39 \\
      &                               & ReAct                & 118 & 51  & 20 & 10 & 0.45 \\
      & \multirow{2}{*}{GPT-OSS-120B} & smolagents CodeAgent & 117 & 71  & 20 & 13 & 0.58 \\
      &                               & ReAct                & 119 & 61  & 16 & 12 & 0.52 \\
      & \multirow{2}{*}{Qwen3.6-35B-A3B}  & smolagents CodeAgent & 119 & 78  & 20 & 12 & 0.47 \\
      &                               & ReAct                & 119 & 78  & 23 & 17 & 0.47 \\
      & \multirow{2}{*}{Qwen3.8-27B}  & smolagents CodeAgent & 119 & 73  & 23 & 12 & 0.58 \\
      &                               & ReAct                & 119 & 79  & 22 & 15 & 0.54 \\
    \midrule
    \multirow{8}{*}{SWE-bench}
      & \multirow{2}{*}{GPT-OSS-20B}  & smolagents CodeAgent & 234 & 215 & 20 & 7 & 0.59 \\
      &                               & ReAct                & 232 & 178 & 24 & 8 & 0.61 \\
      & \multirow{2}{*}{GPT-OSS-120B} & smolagents CodeAgent & 232 & 199 & 23 & 7 & 0.67 \\
      &                               & ReAct                & 233 & 197 & 23 & 8 & 0.67 \\
      & \multirow{2}{*}{Qwen3.6-35B-A3B}  & smolagents CodeAgent & 234 & 217 & 24 & 7 & 0.66 \\
      &                               & ReAct                & 233 & 211 & 24 & 7 & 0.67 \\
      & \multirow{2}{*}{Qwen3.8-27B}  & smolagents CodeAgent & 234 & 213 & 24 & 7 & 0.74 \\
      &                               & ReAct                & 234 & 222 & 24 & 7 & 0.72 \\
    \midrule
    \multirow{8}{*}{CodeContests}
      & \multirow{2}{*}{GPT-OSS-20B}  & smolagents CodeAgent & 60 & 54 & 12 & 7 & 0.50 \\
      &                               & ReAct                & 60 & 29 & 14 & 10 & 0.48 \\
      & \multirow{2}{*}{GPT-OSS-120B} & smolagents CodeAgent & 60 & 53 & 22 & 12 & 0.72 \\
      &                               & ReAct                & 60 & 36 & 21 & 12 & 0.78 \\
      & \multirow{2}{*}{Qwen3.6-35B-A3B}  & smolagents CodeAgent & 60 & 52 & 22 & 13.5 & 0.88 \\
      &                               & ReAct                & 60 & 52 & 22 & 17 & 0.88 \\
      & \multirow{2}{*}{Qwen3.8-27B}  & smolagents CodeAgent & 60 & 58 & 23 & 15 & 0.84 \\
      &                               & ReAct                & 60 & 55 & 23 & 15 & 0.85 \\
    \midrule
    \multirow{3}{*}{DeepSWE}
      & GPT-OSS-20B  & mini-swe-agent & 113 & 99 & 155 & 59  & 0.48 \\
      & GPT-OSS-120B & mini-swe-agent & 113 & 87 & 158 & 57  & 0.58 \\
      & Qwen3.6-35B-A3B  & mini-swe-agent & 113 & 65 & 192 & 125 & 0.78 \\
    \bottomrule
    \end{tabular}}
    \endgroup
    \end{table}
\begin{table}[ht]
\centering
\caption{\textbf{Coding task critics.} Definitions of the critics used to assess candidate solutions and coding trajectories.}
\label{tab:coding-critics}
\resizebox{\linewidth}{!}{%
\begin{tabular}{lll}
\toprule
\textbf{Benchmark} & \textbf{Critic} & \textbf{Description} \\
\midrule
\multirow{4}{*}{\shortstack[l]{LiveCodeBench-Hard, BigCodeBench-Hard,\\SWE-bench, CodeContests}}
  & \texttt{critic\_L0} & the candidate parses without syntax errors \\
  & \texttt{critic\_L1} & static linting finds no undefined names or redefinitions \\
  & \texttt{critic\_L2} & the candidate passes the public tests \\
  & \texttt{critic\_L3} & a reviewer model accepts the candidate \\
\midrule
\multirow{4}{*}{DeepSWE}
  & \texttt{ran\_tests}         & the agent ran the tests at least once \\
  & \texttt{last\_test\_passed} & the last test run passed \\
  & \texttt{test\_flipped}      & a test run passed after an earlier failure \\
  & \texttt{no\_format\_errors} & the agent produced no malformed responses \\
\bottomrule
\end{tabular}}
\end{table}

This subsection describes the LiveCodeBench-Hard, BigCodeBench-Hard, SWE-bench, CodeContests, and DeepSWE environments in detail.

\ecoparagraph{Models and harnesses.}
On LiveCodeBench-Hard, BigCodeBench-Hard, SWE-bench, and CodeContests, we generate trajectories with GPT-OSS-20B, GPT-OSS-120B~\citep{openai2025gptossb}, Qwen3.6-35B-A3B~\citep{qwen36plus}, and Qwen3.8-27B~\citep{qwen38} in thinking mode, each served locally through vLLM~\citep{kwon2023efficient}. Each model runs with two harnesses: smolagents CodeAgent~\citep{smolagents} and ReAct~\citep{yao2023react}. On DeepSWE, GPT-OSS-20B, GPT-OSS-120B~\citep{openai2025gptossb}, and Qwen3.6-35B-A3B~\citep{qwen36plus} run with mini-swe-agent over 113 tasks with a 200-command budget. An episode is successful if the final solution passes all hidden tests; on DeepSWE, where no model resolves a task outright, we use the fraction of passed hidden tests. We evaluate only episodes that the agent completed itself. \Cref{tab:coding-stats} presents all 35 configurations with their numbers of episodes, step statistics, and success rates.

\ecoparagraph{Critics.} Critics differ across benchmarks and are listed in \Cref{tab:coding-critics}. On LiveCodeBench-Hard, BigCodeBench-Hard, SWE-bench, and CodeContests, critics are tools that the agent may invoke on its current candidate solution; SWE-bench provides no public tests, so \texttt{critic\_L2} is unavailable there. On DeepSWE, critics summarize the episode's use of tests and version control. Tool-only and Bayes Fused therefore sum the log-likelihood ratios over steps on the first four benchmarks and count each DeepSWE critic once per episode.

\ecoparagraph{Hardware configuration.}
We use NVIDIA H100-80GB GPUs for LiveCodeBench-Hard, BigCodeBench-Hard, SWE-bench, and CodeContests, and NVIDIA A100-80GB GPUs for DeepSWE.

\subsection{Interactive Environments}
\label{app:exp-interactive}

\begin{table}[ht]
\centering
\begingroup
\renewcommand{\arraystretch}{0.92}%
\caption{\textbf{Interactive task statistics.} For each model and harness configuration, \emph{Episodes} reports the total and completed episode counts. \emph{Steps} reports the maximum and median number of scored steps per episode. \emph{Success rate} is the fraction of successful episodes.}
\label{tab:interactive-stats}
\resizebox{\linewidth}{!}{%
\begin{tabular}{lllrrrrr}
\toprule
\multirow{2}{*}{\textbf{Benchmark}} & \multirow{2}{*}{\textbf{Model}} & \multirow{2}{*}{\textbf{Harness}}
  & \multicolumn{2}{c}{\textbf{Episodes}} & \multicolumn{2}{c}{\textbf{Steps}} & \multirow{2}{*}{\textbf{Success rate}} \\
\cmidrule(lr){4-5}\cmidrule(lr){6-7}
 & & & all & completed & max & median & \\
\midrule
\multirow{6}{*}{$\tau^3$-retail}
  & \multirow{2}{*}{Qwen2.5-32B-Instruct} & LLM agent      & 456 & 445 & 30 & 14 & 0.401 \\
  &                              & Planning agent & 456 & 447 & 30 & 14 & 0.408 \\
  & \multirow{2}{*}{GPT-OSS-20B} & LLM agent      & 456 & 455 & 30 & 12 & 0.417 \\
  &                              & Planning agent & 454 & 451 & 30 & 13 & 0.467 \\
  & \multirow{2}{*}{Qwen3.6-35B-A3B} & LLM agent      & 456 & 451 & 30 & 13 & 0.739 \\
  &                              & Planning agent & 456 & 438 & 30 & 13 & 0.669 \\
\midrule
\multirow{6}{*}{$\tau^3$-airline}
  & \multirow{2}{*}{Qwen2.5-32B-Instruct} & LLM agent      & 200 & 194 & 30 & 10 & 0.305 \\
  &                              & Planning agent & 197 & 191 & 30 & 13 & 0.249 \\
  & \multirow{2}{*}{GPT-OSS-20B} & LLM agent      & 200 & 200 & 27 & 8  & 0.525 \\
  &                              & Planning agent & 200 & 199 & 30 & 9  & 0.525 \\
  & \multirow{2}{*}{Qwen3.6-35B-A3B} & LLM agent      & 200 & 165 & 30 & 13 & 0.640 \\
  &                              & Planning agent & 200 & 175 & 30 & 13 & 0.650 \\
\midrule
\multirow{6}{*}{$\tau^3$-telecom}
  & \multirow{2}{*}{Qwen2.5-32B-Instruct} & LLM agent      & 228 & 154 & 22 & 13 & 0.246 \\
  &                              & Planning agent & 228 & 146 & 25 & 16 & 0.206 \\
  & \multirow{2}{*}{GPT-OSS-20B} & LLM agent      & 228 & 127 & 26 & 18 & 0.329 \\
  &                              & Planning agent & 228 & 111 & 26 & 19 & 0.215 \\
  & \multirow{2}{*}{Qwen3.6-35B-A3B} & LLM agent      & 456 & 225 & 30 & 18 & 0.314 \\
  &                              & Planning agent & 456 & 214 & 30 & 20 & 0.241 \\
\midrule
\multirow{2}{*}{WebArena-Verified}
  & GPT-OSS-20B & GenericAgent & 812 & 315 & 96  & 12 & 0.441 \\
  & Qwen3.6-35B-A3B & GenericAgent & 812 & 662 & 104 & 13 & 0.462 \\
\midrule
OSWorld-Verified
  & Qwen3.6-35B-A3B & Qwen OSWorld agent & 361 & 275 & 100 & 16 & 0.590 \\
\midrule
\multirow{4}{*}{ALFWorld}
  & \multirow{2}{*}{GPT-OSS-20B} & ReAct                & 140 & 97  & 48 & 13 & 0.340 \\
  &                              & smolagents CodeAgent & 140 & 139 & 75 & 17 & 0.353 \\
  & \multirow{2}{*}{Qwen3.6-35B-A3B} & ReAct                & 140 & 86  & 45 & 12 & 0.826 \\
  &                              & smolagents CodeAgent & 140 & 122 & 48 & 9  & 0.893 \\
\bottomrule
\end{tabular}}
\endgroup
\end{table}

\begin{table}[ht]
\centering
\begingroup
\renewcommand{\arraystretch}{0.92}%
\caption{\textbf{Interactive task critics.} Definitions of the critics used to assess agent actions and their effects on the environment.}
\label{tab:interactive-critics}
\resizebox{\linewidth}{!}{%
\begin{tabular}{lll}
\toprule
\textbf{Benchmark} & \textbf{Critic} & \textbf{Description} \\
\midrule
\multirow{3}{*}{$\tau^3$-bench}
  & \texttt{is\_tool\_call}          & the step issues a tool call \\
  & \texttt{tool\_ok}                & the tool call executed without an error \\
  & \texttt{tool\_returned\_content} & the tool returned non-empty content \\
\midrule
\multirow{6}{*}{\shortstack[l]{WebArena-Verified,\\OSWorld-Verified}}
  & \texttt{action\_format\_valid}    & all actions in the episode are well-formed \\
  & \texttt{ui\_execution\_success}   & all actions executed without an error \\
  & \texttt{state\_change}            & at least one action visibly changed the screen \\
  & \texttt{no\_stagnation}           & the agent does not end in a repeated cycle of ineffective actions \\
  & \texttt{llm\_judge\_pass}         & the last LLM verifier call accepted the state \\
  & \texttt{agent\_reported\_success} & the agent reports success (WebArena-Verified only) \\
\midrule
\multirow{4}{*}{ALFWorld}
  & \texttt{format\_valid}          & the response follows the required format \\
  & \texttt{action\_valid}          & the proposed action is admissible \\
  & \texttt{no\_repeated\_fallback} & the harness did not replace the action for exceeding the repeat limit \\
  & \texttt{state\_changed}         & the observation or the set of admissible actions changed \\
\bottomrule
\end{tabular}}
\endgroup
\end{table}

This subsection describes the $\tau^3$-bench, WebArena-Verified, OSWorld-Verified, and ALFWorld environments in detail.

\ecoparagraph{Models and harnesses.}
On $\tau^3$-bench retail, airline, and telecom, we generate trajectories with Qwen2.5-32B-Instruct~\citep{qwen2024qwen}, GPT-OSS-20B~\citep{openai2025gptossb}, and Qwen3.6-35B-A3B~\citep{qwen36plus}, each running with the two $\tau$-bench agents: a plain LLM agent and a planning agent. On WebArena-Verified, GPT-OSS-20B~\citep{openai2025gptossb} and Qwen3.6-35B-A3B~\citep{qwen36plus} run with WebArenaAgent from the benchmark itself; Qwen3.6-35B-A3B observes the accessibility tree together with a screenshot, and GPT-OSS observes the accessibility tree only. OSWorld-Verified is run with Qwen3.6-35B-A3B~\citep{qwen36plus} using the Qwen OSWorld agent. On ALFWorld, GPT-OSS-20B~\citep{openai2025gptossb} and Qwen3.6-35B-A3B~\citep{qwen36plus} run 140 tasks of the \texttt{valid\_seen} split with a ReAct~\citep{yao2023react} scaffold and smolagents CodeAgent under a 50-step budget. On $\tau^3$-bench, an episode is successful if the benchmark reward equals one; on the other environments, success is determined by the benchmark evaluator. We evaluate only episodes that the agent completed itself. \Cref{tab:interactive-stats} presents all 25 configurations with their numbers of episodes, step statistics, and success rates.

\ecoparagraph{Critics.} Critics differ across benchmarks and are listed in \Cref{tab:interactive-critics}. On $\tau^3$-bench and ALFWorld, critics are recorded at each step. On WebArena-Verified and OSWorld-Verified, step-level checks are aggregated into episode-level critics, and \texttt{llm\_judge\_pass} is the verdict of an LLM verifier that the agent may call as a tool. Tool-only and Bayes Fused therefore sum the log-likelihood ratios over steps on $\tau^3$-bench and ALFWorld and count each episode-level critic once on WebArena-Verified and OSWorld-Verified.

\ecoparagraph{Hardware configuration.}
We use four NVIDIA A100-SXM4-80GB GPUs for $\tau^3$-bench, NVIDIA A100-80GB GPUs for ALFWorld, and two NVIDIA RTX PRO 6000 Blackwell GPUs for WebArena-Verified and OSWorld-Verified.

\clearpage
\section{Detailed Results}
\label{app:detailed_results}

\subsection{Average Results}
\label{app:detailed_results_overall}

\begin{table*}[h]
\caption{\textbf{In-domain retrieval results.} PRR@0.5$\uparrow$, averaged over available model and harness configurations on HotpotQA, MuSiQue, CRAG, and BrowseComp-Plus. Raw UQ uses the last-step score, while Bayes Fused uses History + Last aggregation. Avg is the mean across available dataset columns. The best result in each column is \textbf{bold}, and the second-best is \underline{underlined}.}
\label{tab:prr-dataset-means-rag}
\centering
\begingroup
\resizebox{\textwidth}{!}{%
\begin{tabular}{ll|cccc|c}
\toprule
\textbf{Group} & \textbf{Method} & \textbf{HotpotQA} & \textbf{MuSiQue} & \textbf{CRAG} & \textbf{BrowseComp} & \textbf{Average} \\

\midrule
\multirow{2}{*}{\rotatebox[origin=c]{90}{\scriptsize Stat.}} & $N$ steps & .218 {\scriptsize $\pm$ .010} & .370 {\scriptsize $\pm$ .011} & .185 {\scriptsize $\pm$ .014} & .810 {\scriptsize $\pm$ .000} & .396 {\scriptsize $\pm$ .005} \\
 & Tool success rate & .293 {\scriptsize $\pm$ .010} & .340 {\scriptsize $\pm$ .007} & -.031 {\scriptsize $\pm$ .021} & -.005 {\scriptsize $\pm$ .000} & .149 {\scriptsize $\pm$ .006} \\
\cmidrule(lr){2-7}
\multirow{7}{*}{\rotatebox[origin=c]{90}{\scriptsize UQ}} & SP & .129 {\scriptsize $\pm$ .010} & .126 {\scriptsize $\pm$ .005} & .267 {\scriptsize $\pm$ .010} & .515 {\scriptsize $\pm$ .000} & .259 {\scriptsize $\pm$ .004} \\
 & Perplexity & -.045 {\scriptsize $\pm$ .011} & -.004 {\scriptsize $\pm$ .002} & .044 {\scriptsize $\pm$ .005} & .900 {\scriptsize $\pm$ .000} & .224 {\scriptsize $\pm$ .003} \\
 & MTE & -.057 {\scriptsize $\pm$ .010} & .002 {\scriptsize $\pm$ .005} & .068 {\scriptsize $\pm$ .005} & .915 {\scriptsize $\pm$ .000} & .232 {\scriptsize $\pm$ .003} \\
 & Self-certainty & .230 {\scriptsize $\pm$ .011} & .173 {\scriptsize $\pm$ .007} & .219 {\scriptsize $\pm$ .007} & .900 {\scriptsize $\pm$ .000} & .380 {\scriptsize $\pm$ .004} \\
 & Verb. & \underline{.408} {\scriptsize $\pm$ .002} & .320 {\scriptsize $\pm$ .000} & \textbf{.474} {\scriptsize $\pm$ .000} & .715 {\scriptsize $\pm$ .000} & .479 {\scriptsize $\pm$ .000} \\
 & SAUP & .174 {\scriptsize $\pm$ .002} & .190 {\scriptsize $\pm$ .009} & .225 {\scriptsize $\pm$ .013} & .400 {\scriptsize $\pm$ .000} & .247 {\scriptsize $\pm$ .004} \\
 & UProp & .357 {\scriptsize $\pm$ .000} & .208 {\scriptsize $\pm$ .000} & .070 {\scriptsize $\pm$ .000} & .815 {\scriptsize $\pm$ .000} & .363 {\scriptsize $\pm$ .000} \\
\cmidrule(lr){2-7}
\multirow{2}{*}{\rotatebox[origin=c]{90}{\scriptsize\shortstack{Super-\\vised}}} & HTC & .371 {\scriptsize $\pm$ .016} & \textbf{.469} {\scriptsize $\pm$ .020} & .224 {\scriptsize $\pm$ .025} & \underline{.945} {\scriptsize $\pm$ .005} & .502 {\scriptsize $\pm$ .009} \\
 & \shortstack[l]{Logistic regression} & .307 {\scriptsize $\pm$ .018} & .350 {\scriptsize $\pm$ .010} & .358 {\scriptsize $\pm$ .027} & \textbf{.975} {\scriptsize $\pm$ .005} & .498 {\scriptsize $\pm$ .008} \\
\cmidrule(lr){2-7}
\multirow{7}{*}{\rotatebox[origin=c]{90}{\scriptsize \textbf{Bayes (ours)}}} & Tool-only & .307 {\scriptsize $\pm$ .011} & .353 {\scriptsize $\pm$ .009} & -.020 {\scriptsize $\pm$ .025} & .445 {\scriptsize $\pm$ .030} & .271 {\scriptsize $\pm$ .010} \\
 & Fused (SP) & .329 {\scriptsize $\pm$ .013} & .376 {\scriptsize $\pm$ .009} & .129 {\scriptsize $\pm$ .019} & .585 {\scriptsize $\pm$ .000} & .355 {\scriptsize $\pm$ .006} \\
 & Fused (Perplexity) & .323 {\scriptsize $\pm$ .010} & .374 {\scriptsize $\pm$ .012} & .177 {\scriptsize $\pm$ .021} & .875 {\scriptsize $\pm$ .000} & .437 {\scriptsize $\pm$ .006} \\
 & Fused (MTE) & .325 {\scriptsize $\pm$ .010} & .376 {\scriptsize $\pm$ .007} & .198 {\scriptsize $\pm$ .022} & .895 {\scriptsize $\pm$ .000} & .449 {\scriptsize $\pm$ .006} \\
 & Fused (Self-certainty) & .319 {\scriptsize $\pm$ .009} & .384 {\scriptsize $\pm$ .008} & .213 {\scriptsize $\pm$ .021} & .875 {\scriptsize $\pm$ .000} & .448 {\scriptsize $\pm$ .006} \\
 & Fused (Verb.) & \textbf{.447} {\scriptsize $\pm$ .005} & \underline{.443} {\scriptsize $\pm$ .006} & \underline{.391} {\scriptsize $\pm$ .013} & .940 {\scriptsize $\pm$ .005} & \textbf{.555} {\scriptsize $\pm$ .004} \\
 & SBF & .334 {\scriptsize $\pm$ .011} & .403 {\scriptsize $\pm$ .009} & \underline{.391} {\scriptsize $\pm$ .015} & .930 {\scriptsize $\pm$ .000} & \underline{.515} {\scriptsize $\pm$ .005} \\
\bottomrule
\end{tabular}%
}
\endgroup
\end{table*}

\begin{table*}[h]
\caption{\textbf{In-domain coding results.} PRR@0.5$\uparrow$, averaged over available model and harness configurations on LiveCodeBench-Hard, BigCodeBench-Hard, SWE-bench, CodeContests, and DeepSWE. Raw UQ uses the trajectory mean. Bayes Fused uses Continuous aggregation after a rank-Gaussian transformation fitted to training-fold signal values without using labels or target data. Class-conditional Gaussian distributions are fitted to the transformed values. Avg is the mean across available dataset columns. The best result in each column is \textbf{bold}, and the second-best is \underline{underlined}.}
\label{tab:prr-dataset-means-coding}
\centering
\begingroup
\resizebox{\textwidth}{!}{%
\begin{tabular}{ll|ccccc|c}
\toprule
\textbf{Group} & \textbf{Method} & \textbf{LCB} & \textbf{BCB} & \textbf{SWE} & \textbf{CC} & \textbf{DeepSWE} & \textbf{Average} \\
\midrule
\multirow{2}{*}{\rotatebox[origin=c]{90}{\scriptsize Stat.}} & $N$ steps & .082 {\scriptsize $\pm$ .000} & .073 {\scriptsize $\pm$ .000} & .133 {\scriptsize $\pm$ .000} & .262 {\scriptsize $\pm$ .000} & .053 {\scriptsize $\pm$ .000} & .121 {\scriptsize $\pm$ .000} \\
 & Tool success rate & .171 {\scriptsize $\pm$ .000} & .128 {\scriptsize $\pm$ .000} & .088 {\scriptsize $\pm$ .000} & .422 {\scriptsize $\pm$ .000} & -.163 {\scriptsize $\pm$ .000} & .129 {\scriptsize $\pm$ .000} \\
\cmidrule(lr){2-8}
\multirow{7}{*}{\rotatebox[origin=c]{90}{\scriptsize UQ}} & SP & .278 {\scriptsize $\pm$ .000} & .254 {\scriptsize $\pm$ .000} & \underline{.364} {\scriptsize $\pm$ .000} & \underline{.553} {\scriptsize $\pm$ .000} & .017 {\scriptsize $\pm$ .000} & .293 {\scriptsize $\pm$ .000} \\
 & Perplexity & .047 {\scriptsize $\pm$ .000} & .192 {\scriptsize $\pm$ .000} & .218 {\scriptsize $\pm$ .000} & .204 {\scriptsize $\pm$ .000} & .007 {\scriptsize $\pm$ .000} & .134 {\scriptsize $\pm$ .000} \\
 & MTE & .058 {\scriptsize $\pm$ .000} & .187 {\scriptsize $\pm$ .000} & .221 {\scriptsize $\pm$ .000} & .207 {\scriptsize $\pm$ .000} & .023 {\scriptsize $\pm$ .000} & .139 {\scriptsize $\pm$ .000} \\
 & Self-certainty & .173 {\scriptsize $\pm$ .000} & .174 {\scriptsize $\pm$ .000} & .268 {\scriptsize $\pm$ .000} & .387 {\scriptsize $\pm$ .000} & -.030 {\scriptsize $\pm$ .000} & .195 {\scriptsize $\pm$ .000} \\
 & Verb. & .111 {\scriptsize $\pm$ .000} & .157 {\scriptsize $\pm$ .000} & .318 {\scriptsize $\pm$ .000} & .406 {\scriptsize $\pm$ .000} & -.007 {\scriptsize $\pm$ .000} & .197 {\scriptsize $\pm$ .000} \\
 & SAUP & .070 {\scriptsize $\pm$ .000} & .137 {\scriptsize $\pm$ .000} & .188 {\scriptsize $\pm$ .000} & .226 {\scriptsize $\pm$ .000} & .030 {\scriptsize $\pm$ .000} & .130 {\scriptsize $\pm$ .000} \\
 & UProp & .066 {\scriptsize $\pm$ .000} & \textbf{.267} {\scriptsize $\pm$ .000} & .169 {\scriptsize $\pm$ .000} & .268 {\scriptsize $\pm$ .000} & -.010 {\scriptsize $\pm$ .000} & .152 {\scriptsize $\pm$ .000} \\
\cmidrule(lr){2-8}
\multirow{2}{*}{\rotatebox[origin=c]{90}{\scriptsize \shortstack{Super-\\vised}}} & HTC & .184 {\scriptsize $\pm$ .018} & .114 {\scriptsize $\pm$ .049} & .355 {\scriptsize $\pm$ .015} & .094 {\scriptsize $\pm$ .106} & .120 {\scriptsize $\pm$ .038} & .173 {\scriptsize $\pm$ .025} \\
 & \shortstack[l]{Logistic regression} & .280 {\scriptsize $\pm$ .004} & .187 {\scriptsize $\pm$ .033} & .354 {\scriptsize $\pm$ .005} & .491 {\scriptsize $\pm$ .024} & .073 {\scriptsize $\pm$ .032} & .277 {\scriptsize $\pm$ .010} \\
\cmidrule(lr){2-8}
\multirow{7}{*}{\rotatebox[origin=c]{90}{\scriptsize \textbf{Bayes (ours)}}} & Tool-only & .191 {\scriptsize $\pm$ .008} & .135 {\scriptsize $\pm$ .034} & .169 {\scriptsize $\pm$ .010} & .383 {\scriptsize $\pm$ .020} & .120 {\scriptsize $\pm$ .038} & .199 {\scriptsize $\pm$ .011} \\
 & Fused (SP) & \underline{.297} {\scriptsize $\pm$ .002} & \underline{.262} {\scriptsize $\pm$ .026} & .362 {\scriptsize $\pm$ .003} & \textbf{.561} {\scriptsize $\pm$ .009} & \underline{.207} {\scriptsize $\pm$ .021} & \textbf{.338} {\scriptsize $\pm$ .007} \\
 & Fused (Perplexity) & .156 {\scriptsize $\pm$ .011} & .205 {\scriptsize $\pm$ .012} & .204 {\scriptsize $\pm$ .016} & .297 {\scriptsize $\pm$ .018} & \textbf{.217} {\scriptsize $\pm$ .026} & .216 {\scriptsize $\pm$ .008} \\
 & Fused (MTE) & .160 {\scriptsize $\pm$ .011} & .208 {\scriptsize $\pm$ .009} & .207 {\scriptsize $\pm$ .015} & .323 {\scriptsize $\pm$ .018} & .167 {\scriptsize $\pm$ .030} & .213 {\scriptsize $\pm$ .008} \\
 & Fused (Self-certainty) & .226 {\scriptsize $\pm$ .004} & .204 {\scriptsize $\pm$ .008} & .248 {\scriptsize $\pm$ .010} & .435 {\scriptsize $\pm$ .015} & .137 {\scriptsize $\pm$ .031} & .250 {\scriptsize $\pm$ .007} \\
 & Fused (Verb.) & .257 {\scriptsize $\pm$ .004} & .146 {\scriptsize $\pm$ .025} & .336 {\scriptsize $\pm$ .006} & .328 {\scriptsize $\pm$ .017} & .090 {\scriptsize $\pm$ .039} & .231 {\scriptsize $\pm$ .010} \\
 & SBF & \textbf{.308} {\scriptsize $\pm$ .010} & .132 {\scriptsize $\pm$ .022} & \textbf{.371} {\scriptsize $\pm$ .009} & .546 {\scriptsize $\pm$ .029} & .173 {\scriptsize $\pm$ .036} & \underline{.306} {\scriptsize $\pm$ .011} \\
\bottomrule
\end{tabular}%
}
\endgroup
\end{table*}

\begin{table*}[h]
\caption{\textbf{In-domain interactive results.} PRR@0.5$\uparrow$, averaged over available model and harness configurations on $\tau^3$-bench, WebArena-Verified, OSWorld-Verified, and ALFWorld. Raw UQ uses the last-step score, while Bayes Fused uses History + Last aggregation. Avg is the mean across available dataset columns. The best result in each column is \textbf{bold}, and the second-best is \underline{underlined}.}
\label{tab:prr-dataset-means-other}
\centering
\begingroup
\resizebox{\textwidth}{!}{%
\begin{tabular}{ll|cccccc|c}
\toprule
\textbf{Group} & \textbf{Method} & \textbf{$\tau^3$-retail} & \textbf{$\tau^3$-airline} & \textbf{$\tau^3$-telecom} & \textbf{WebArena} & \textbf{OSWorld} & \textbf{ALFWorld} & \textbf{Average} \\
\midrule
\multirow{2}{*}{\rotatebox[origin=c]{90}{\scriptsize Stat.}} & $N$ steps & .131 {\scriptsize $\pm$ .015} & .357 {\scriptsize $\pm$ .030} & .501 {\scriptsize $\pm$ .032} & .290 {\scriptsize $\pm$ .000} & .350 {\scriptsize $\pm$ .000} & .388 {\scriptsize $\pm$ .000} & .336 {\scriptsize $\pm$ .008} \\
  & Tool success rate & .142 {\scriptsize $\pm$ .015} & .145 {\scriptsize $\pm$ .048} & .206 {\scriptsize $\pm$ .022} & .400 {\scriptsize $\pm$ .000} & .000 {\scriptsize $\pm$ .000} & .538 {\scriptsize $\pm$ .000} & .238 {\scriptsize $\pm$ .009} \\
\cmidrule(lr){2-9}
\multirow{7}{*}{\rotatebox[origin=c]{90}{\scriptsize UQ}} & SP & .097 {\scriptsize $\pm$ .023} & .077 {\scriptsize $\pm$ .046} & -.058 {\scriptsize $\pm$ .017} & .435 {\scriptsize $\pm$ .000} & .220 {\scriptsize $\pm$ .000} & .323 {\scriptsize $\pm$ .000} & .182 {\scriptsize $\pm$ .009} \\
  & Perplexity & -.026 {\scriptsize $\pm$ .010} & -.126 {\scriptsize $\pm$ .040} & -.060 {\scriptsize $\pm$ .020} & .470 {\scriptsize $\pm$ .000} & .040 {\scriptsize $\pm$ .000} & .250 {\scriptsize $\pm$ .000} & .091 {\scriptsize $\pm$ .008} \\
  & MTE & -.021 {\scriptsize $\pm$ .009} & -.111 {\scriptsize $\pm$ .041} & -.080 {\scriptsize $\pm$ .018} & .480 {\scriptsize $\pm$ .000} & .060 {\scriptsize $\pm$ .000} & .310 {\scriptsize $\pm$ .000} & .106 {\scriptsize $\pm$ .008} \\
  & Self-certainty & .077 {\scriptsize $\pm$ .015} & .042 {\scriptsize $\pm$ .027} & -.026 {\scriptsize $\pm$ .011} & .545 {\scriptsize $\pm$ .000} & .280 {\scriptsize $\pm$ .000} & .340 {\scriptsize $\pm$ .000} & .210 {\scriptsize $\pm$ .005} \\
  & Verb. & .170 {\scriptsize $\pm$ .000} & -.225 {\scriptsize $\pm$ .000} & .184 {\scriptsize $\pm$ .000} & .285 {\scriptsize $\pm$ .000} & .280 {\scriptsize $\pm$ .000} & .188 {\scriptsize $\pm$ .000} & .147 {\scriptsize $\pm$ .000} \\
  & SAUP & .136 {\scriptsize $\pm$ .006} & -.332 {\scriptsize $\pm$ .010} & .419 {\scriptsize $\pm$ .018} & .170 {\scriptsize $\pm$ .005} & .100 {\scriptsize $\pm$ .010} & .313 {\scriptsize $\pm$ .043} & .134 {\scriptsize $\pm$ .008} \\
  & UProp & .231 {\scriptsize $\pm$ .000} & .088 {\scriptsize $\pm$ .000} & .254 {\scriptsize $\pm$ .000} & .420 {\scriptsize $\pm$ .000} & .050 {\scriptsize $\pm$ .000} & .405 {\scriptsize $\pm$ .000} & .241 {\scriptsize $\pm$ .000} \\
\cmidrule(lr){2-9}
\multirow{2}{*}{\rotatebox[origin=c]{90}{\scriptsize \shortstack{Super-\\vised}}} & HTC & \underline{.278} {\scriptsize $\pm$ .033} & \underline{.364} {\scriptsize $\pm$ .019} & .574 {\scriptsize $\pm$ .029} & .555 {\scriptsize $\pm$ .025} & .360 {\scriptsize $\pm$ .020} & .528 {\scriptsize $\pm$ .029} & .443 {\scriptsize $\pm$ .011} \\
  & \shortstack[l]{Logistic regression} & .222 {\scriptsize $\pm$ .017} & .305 {\scriptsize $\pm$ .030} & \textbf{.646} {\scriptsize $\pm$ .022} & \textbf{.575} {\scriptsize $\pm$ .005} & \textbf{.450} {\scriptsize $\pm$ .010} & \underline{.775} {\scriptsize $\pm$ .033} & \underline{.496} {\scriptsize $\pm$ .009} \\
\cmidrule(lr){2-9}
\multirow{7}{*}{\rotatebox[origin=c]{90}{\scriptsize \textbf{Bayes (ours)}}} & Tool-only & .154 {\scriptsize $\pm$ .013} & .186 {\scriptsize $\pm$ .041} & .434 {\scriptsize $\pm$ .012} & .345 {\scriptsize $\pm$ .021} & .210 {\scriptsize $\pm$ .010} & .648 {\scriptsize $\pm$ .017} & .330 {\scriptsize $\pm$ .009} \\
  & Fused (SP) & \textbf{.286} {\scriptsize $\pm$ .010} & .230 {\scriptsize $\pm$ .022} & .518 {\scriptsize $\pm$ .011} & .505 {\scriptsize $\pm$ .005} & .360 {\scriptsize $\pm$ .010} & .528 {\scriptsize $\pm$ .015} & .404 {\scriptsize $\pm$ .005} \\
  & Fused (Perplexity) & .215 {\scriptsize $\pm$ .013} & .213 {\scriptsize $\pm$ .025} & .475 {\scriptsize $\pm$ .013} & .545 {\scriptsize $\pm$ .000} & .170 {\scriptsize $\pm$ .030} & .523 {\scriptsize $\pm$ .018} & .357 {\scriptsize $\pm$ .008} \\
  & Fused (MTE) & .206 {\scriptsize $\pm$ .018} & .303 {\scriptsize $\pm$ .032} & .480 {\scriptsize $\pm$ .017} & .540 {\scriptsize $\pm$ .000} & .160 {\scriptsize $\pm$ .020} & .530 {\scriptsize $\pm$ .017} & .370 {\scriptsize $\pm$ .008} \\
  & Fused (Self-certainty) & .213 {\scriptsize $\pm$ .013} & .306 {\scriptsize $\pm$ .024} & .530 {\scriptsize $\pm$ .016} & \underline{.570} {\scriptsize $\pm$ .000} & .300 {\scriptsize $\pm$ .020} & .540 {\scriptsize $\pm$ .023} & .410 {\scriptsize $\pm$ .007} \\
  & Fused (Verb.) & .256 {\scriptsize $\pm$ .012} & .187 {\scriptsize $\pm$ .041} & .536 {\scriptsize $\pm$ .011} & .505 {\scriptsize $\pm$ .005} & .270 {\scriptsize $\pm$ .020} & .540 {\scriptsize $\pm$ .003} & .382 {\scriptsize $\pm$ .008} \\
  & SBF & \underline{.278} {\scriptsize $\pm$ .019} & \textbf{.479} {\scriptsize $\pm$ .013} & \underline{.626} {\scriptsize $\pm$ .018} & .545 {\scriptsize $\pm$ .007} & \underline{.400} {\scriptsize $\pm$ .010} & \textbf{.825} {\scriptsize $\pm$ .010} & \textbf{.526} {\scriptsize $\pm$ .006} \\
\bottomrule
\end{tabular}%
}
\endgroup
\end{table*}

\clearpage
\section{Generalization across Models, Agents, Environments, and Domains}
\label{app:generalization}

\subsection{Average Cross-Environment and Cross-Domain Results}
\label{app:ood_additional_results}

  \begin{figure}[h]
    \centering
    \includegraphics[trim={0.cm 0.cm 0.cm 0.cm},clip,width=\linewidth]{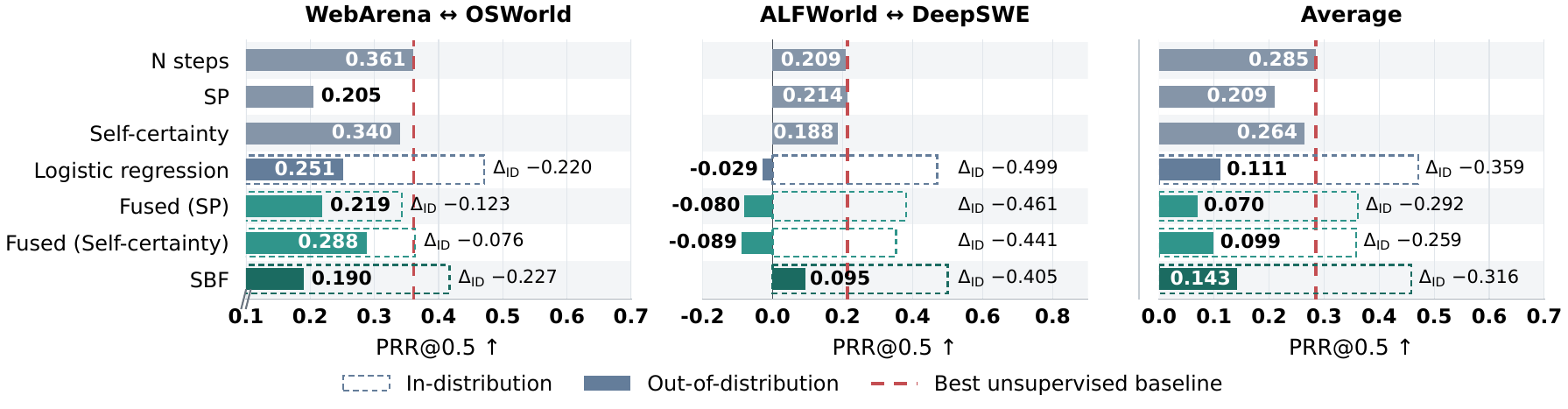}
    \caption{
        \textbf{Cross-environment and cross-domain transfer.} Filled bars show out-of-distribution PRR@0.5$\uparrow$, while dashed bars show the corresponding in-distribution scores. Bars average performance between WebArena and OSWorld (environment shift) or between ALFWorld and DeepSWE (domain shift). $\Delta$ denotes out-of-distribution performance minus in-distribution performance. Detailed per-transfer results are provided in \Cref{tab:env_transfer,tab:full-shift-alfworld-deepswe}.
    }
    \label{fig:cross_environment_domain_id_ood}
  \end{figure}

\subsection{Model Transfer}
\label{sec:model_transfer}

\begin{table*}[h]
\caption{\textbf{Model transfer.} PRR@0.5$\uparrow$ with the dataset and harness held fixed. Column headers indicate the source $\to$ target model. GPT and Qwen denote GPT-OSS-20B and Qwen3.6-35B-A3B, respectively. CodeAgent is the smolagents harness. $\Delta_{\mathrm{ID}}$ is Avg minus the mean in-domain PRR for the same targets. In-domain scores are reused for $N$ steps, tool success rate, and raw UQ. The best result in each column is \textbf{bold}, and the second-best is \underline{underlined}.}
\label{tab:model_transfer}
\centering
\setlength{\tabcolsep}{2pt}
\resizebox{\textwidth}{!}{%
\begin{tabular}{ll|cc|cccc|cccc|cc}
\toprule
\multirow{3}{*}{\textbf{Group}} &
\multirow{3}{*}{\textbf{Method}} &
\multicolumn{2}{c|}{\textbf{WebArena}} &
\multicolumn{4}{c|}{\textbf{HotpotQA}} &
\multicolumn{4}{c|}{\textbf{MuSiQue}} &
\multirow{3}{*}{\textbf{Avg}} &
\multirow{3}{*}{$\boldsymbol{\Delta_{\mathrm{ID}}}$} \\
\cmidrule(lr){3-4}\cmidrule(lr){5-8}\cmidrule(lr){9-12}
& &
\textbf{Qwen $\to$ GPT} &
\textbf{GPT $\to$ Qwen} &
\multicolumn{2}{c|}{\textbf{GPT $\to$ Qwen}} &
\multicolumn{2}{c|}{\textbf{Qwen $\to$ GPT}} &
\multicolumn{2}{c|}{\textbf{GPT $\to$ Qwen}} &
\multicolumn{2}{c|}{\textbf{Qwen $\to$ GPT}} &
& \\
\cmidrule(lr){3-3}\cmidrule(lr){4-4}
\cmidrule(lr){5-6}\cmidrule(lr){7-8}
\cmidrule(lr){9-10}\cmidrule(lr){11-12}
& &
WebArenaAgent & WebArenaAgent &
CodeAgent & ReAct &
CodeAgent & ReAct &
CodeAgent & ReAct &
CodeAgent & ReAct &
& \\
\midrule
\multirow{2}{*}{\rotatebox[origin=c]{90}{\scriptsize Stat.}} & $N$ steps & .207 {\scriptsize $\pm$ .000} & .369 {\scriptsize $\pm$ .000} & .380 {\scriptsize $\pm$ .000} & .465 {\scriptsize $\pm$ .020} & .088 {\scriptsize $\pm$ .000} & .341 {\scriptsize $\pm$ .020} & .377 {\scriptsize $\pm$ .010} & \textbf{.662} {\scriptsize $\pm$ .010} & .192 {\scriptsize $\pm$ .050} & \underline{.656} {\scriptsize $\pm$ .000} & .374 {\scriptsize $\pm$ .006} & \textendash \\
  & Tool success rate & .494 {\scriptsize $\pm$ .000} & .306 {\scriptsize $\pm$ .000} & .369 {\scriptsize $\pm$ .030} & .515 {\scriptsize $\pm$ .020} & .154 {\scriptsize $\pm$ .010} & .503 {\scriptsize $\pm$ .010} & .336 {\scriptsize $\pm$ .010} & .493 {\scriptsize $\pm$ .000} & .225 {\scriptsize $\pm$ .030} & .545 {\scriptsize $\pm$ .010} & .394 {\scriptsize $\pm$ .005} & \textendash \\
\cmidrule(lr){2-14}
\multirow{7}{*}{\rotatebox[origin=c]{90}{\scriptsize UQ}} & SP & .680 {\scriptsize $\pm$ .000} & .185 {\scriptsize $\pm$ .000} & .387 {\scriptsize $\pm$ .000} & .072 {\scriptsize $\pm$ .000} & -.060 {\scriptsize $\pm$ .000} & .106 {\scriptsize $\pm$ .000} & .257 {\scriptsize $\pm$ .000} & .118 {\scriptsize $\pm$ .000} & .030 {\scriptsize $\pm$ .000} & .010 {\scriptsize $\pm$ .000} & .179 {\scriptsize $\pm$ .000} & \textendash \\
  & Perplexity & .584 {\scriptsize $\pm$ .000} & .364 {\scriptsize $\pm$ .000} & -.220 {\scriptsize $\pm$ .000} & -.116 {\scriptsize $\pm$ .000} & -.067 {\scriptsize $\pm$ .000} & -.211 {\scriptsize $\pm$ .000} & -.021 {\scriptsize $\pm$ .000} & -.160 {\scriptsize $\pm$ .000} & -.051 {\scriptsize $\pm$ .000} & -.213 {\scriptsize $\pm$ .000} & -.011 {\scriptsize $\pm$ .000} & \textendash \\
  & MTE & .590 {\scriptsize $\pm$ .000} & .373 {\scriptsize $\pm$ .000} & -.265 {\scriptsize $\pm$ .000} & -.135 {\scriptsize $\pm$ .000} & -.068 {\scriptsize $\pm$ .000} & -.213 {\scriptsize $\pm$ .000} & -.099 {\scriptsize $\pm$ .000} & -.083 {\scriptsize $\pm$ .000} & -.018 {\scriptsize $\pm$ .000} & -.227 {\scriptsize $\pm$ .000} & -.014 {\scriptsize $\pm$ .000} & \textendash \\
  & Self-certainty & \underline{.692} {\scriptsize $\pm$ .000} & .397 {\scriptsize $\pm$ .000} & \underline{.418} {\scriptsize $\pm$ .000} & .331 {\scriptsize $\pm$ .000} & .095 {\scriptsize $\pm$ .000} & .275 {\scriptsize $\pm$ .000} & .188 {\scriptsize $\pm$ .000} & .275 {\scriptsize $\pm$ .000} & -.010 {\scriptsize $\pm$ .000} & .222 {\scriptsize $\pm$ .000} & .288 {\scriptsize $\pm$ .000} & \textendash \\
  & Verb. & .217 {\scriptsize $\pm$ .000} & .346 {\scriptsize $\pm$ .000} & .396 {\scriptsize $\pm$ .000} & .577 {\scriptsize $\pm$ .000} & .205 {\scriptsize $\pm$ .000} & .374 {\scriptsize $\pm$ .000} & .245 {\scriptsize $\pm$ .000} & .394 {\scriptsize $\pm$ .000} & .074 {\scriptsize $\pm$ .000} & .342 {\scriptsize $\pm$ .000} & .317 {\scriptsize $\pm$ .000} & \textendash \\
  & SAUP & .323 {\scriptsize $\pm$ .007} & .156 {\scriptsize $\pm$ .005} & .341 {\scriptsize $\pm$ .000} & .104 {\scriptsize $\pm$ .000} & .100 {\scriptsize $\pm$ .000} & .224 {\scriptsize $\pm$ .000} & .159 {\scriptsize $\pm$ .000} & .106 {\scriptsize $\pm$ .000} & .040 {\scriptsize $\pm$ .000} & .303 {\scriptsize $\pm$ .000} & .186 {\scriptsize $\pm$ .001} & {+.019} \\
  & UProp & .535 {\scriptsize $\pm$ .000} & .315 {\scriptsize $\pm$ .000} & \textbf{.485} {\scriptsize $\pm$ .000} & \underline{.586} {\scriptsize $\pm$ .000} & \underline{.249} {\scriptsize $\pm$ .000} & .344 {\scriptsize $\pm$ .000} & .340 {\scriptsize $\pm$ .000} & .250 {\scriptsize $\pm$ .000} & .199 {\scriptsize $\pm$ .000} & .368 {\scriptsize $\pm$ .000} & .367 {\scriptsize $\pm$ .000} & \textendash \\
\cmidrule(lr){2-14}
\multirow{2}{*}{\rotatebox[origin=c]{90}{\scriptsize \shortstack{Super-\\vised}}} & HTC & .460 {\scriptsize $\pm$ .071} & .176 {\scriptsize $\pm$ .024} & .370 {\scriptsize $\pm$ .000} & -.067 {\scriptsize $\pm$ .000} & .128 {\scriptsize $\pm$ .000} & .327 {\scriptsize $\pm$ .000} & \underline{.384} {\scriptsize $\pm$ .000} & .582 {\scriptsize $\pm$ .000} & .270 {\scriptsize $\pm$ .000} & .548 {\scriptsize $\pm$ .000} & .318 {\scriptsize $\pm$ .007} & -.128 \\
  & Logistic regression & .613 {\scriptsize $\pm$ .027} & .278 {\scriptsize $\pm$ .034} & .231 {\scriptsize $\pm$ .000} & .135 {\scriptsize $\pm$ .000} & -.034 {\scriptsize $\pm$ .000} & -.058 {\scriptsize $\pm$ .000} & .137 {\scriptsize $\pm$ .000} & .092 {\scriptsize $\pm$ .000} & .030 {\scriptsize $\pm$ .000} & -.040 {\scriptsize $\pm$ .000} & .138 {\scriptsize $\pm$ .004} & -.291 \\
\cmidrule(lr){2-14}
\multirow{7}{*}{\rotatebox[origin=c]{90}{\scriptsize \textbf{Bayes (ours)}}} & Tool-only & .464 {\scriptsize $\pm$ .049} & .064 {\scriptsize $\pm$ .022} & .401 {\scriptsize $\pm$ .000} & .553 {\scriptsize $\pm$ .000} & .177 {\scriptsize $\pm$ .000} & \underline{.524} {\scriptsize $\pm$ .000} & .375 {\scriptsize $\pm$ .000} & .555 {\scriptsize $\pm$ .000} & .246 {\scriptsize $\pm$ .000} & .610 {\scriptsize $\pm$ .000} & .397 {\scriptsize $\pm$ .005} & {+.007} \\
  & Fused (SP) & .642 {\scriptsize $\pm$ .006} & .269 {\scriptsize $\pm$ .003} & .353 {\scriptsize $\pm$ .000} & .543 {\scriptsize $\pm$ .000} & .217 {\scriptsize $\pm$ .000} & .316 {\scriptsize $\pm$ .000} & .380 {\scriptsize $\pm$ .000} & .576 {\scriptsize $\pm$ .000} & \underline{.284} {\scriptsize $\pm$ .000} & .399 {\scriptsize $\pm$ .000} & .398 {\scriptsize $\pm$ .001} & -.034 \\
  & Fused (Perplexity) & .682 {\scriptsize $\pm$ .007} & \textbf{.428} {\scriptsize $\pm$ .001} & .324 {\scriptsize $\pm$ .000} & .530 {\scriptsize $\pm$ .000} & .102 {\scriptsize $\pm$ .000} & .413 {\scriptsize $\pm$ .000} & .336 {\scriptsize $\pm$ .000} & .570 {\scriptsize $\pm$ .000} & .186 {\scriptsize $\pm$ .000} & .468 {\scriptsize $\pm$ .000} & .404 {\scriptsize $\pm$ .001} & -.024 \\
  & Fused (MTE) & .661 {\scriptsize $\pm$ .007} & .411 {\scriptsize $\pm$ .003} & .308 {\scriptsize $\pm$ .000} & .534 {\scriptsize $\pm$ .000} & .120 {\scriptsize $\pm$ .000} & .487 {\scriptsize $\pm$ .000} & .284 {\scriptsize $\pm$ .000} & .501 {\scriptsize $\pm$ .000} & .257 {\scriptsize $\pm$ .000} & .610 {\scriptsize $\pm$ .000} & .417 {\scriptsize $\pm$ .001} & -.022 \\
  & Fused (Self-certainty) & \textbf{.708} {\scriptsize $\pm$ .002} & \underline{.414} {\scriptsize $\pm$ .002} & .340 {\scriptsize $\pm$ .000} & .526 {\scriptsize $\pm$ .000} & .131 {\scriptsize $\pm$ .000} & .424 {\scriptsize $\pm$ .000} & .350 {\scriptsize $\pm$ .000} & .509 {\scriptsize $\pm$ .000} & .250 {\scriptsize $\pm$ .000} & .576 {\scriptsize $\pm$ .000} & .423 {\scriptsize $\pm$ .000} & -.028 \\
  & Fused (Verb.) & .589 {\scriptsize $\pm$ .010} & .368 {\scriptsize $\pm$ .005} & \textbf{.485} {\scriptsize $\pm$ .000} & \textbf{.639} {\scriptsize $\pm$ .000} & \textbf{.308} {\scriptsize $\pm$ .000} & \textbf{.604} {\scriptsize $\pm$ .000} & .361 {\scriptsize $\pm$ .000} & .563 {\scriptsize $\pm$ .000} & \textbf{.474} {\scriptsize $\pm$ .000} & \textbf{.674} {\scriptsize $\pm$ .000} & \textbf{.507} {\scriptsize $\pm$ .001} & {+.048} \\
  & SBF & .573 {\scriptsize $\pm$ .015} & .375 {\scriptsize $\pm$ .013} & .408 {\scriptsize $\pm$ .000} & \underline{.586} {\scriptsize $\pm$ .000} & .146 {\scriptsize $\pm$ .000} & .429 {\scriptsize $\pm$ .000} & \textbf{.399} {\scriptsize $\pm$ .000} & \underline{.647} {\scriptsize $\pm$ .000} & .250 {\scriptsize $\pm$ .000} & .652 {\scriptsize $\pm$ .000} & \underline{.446} {\scriptsize $\pm$ .002} & -.015 \\
\bottomrule
\end{tabular}
}
\end{table*}

\subsection{Agent Transfer}
\label{sec:agent_transfer}

\begin{table*}[h]
\caption{\textbf{Harness transfer.} PRR@0.5$\uparrow$ with the dataset and model held fixed. Column headers indicate the source $\to$ target harness. Code and Tool denote CodeAgent and ToolAgent from smolagents, respectively. $\Delta_{\mathrm{ID}}$ is Avg minus the mean in-domain PRR for the same targets. In-domain scores are reused for $N$ steps, tool success rate, and raw UQ. The best result in each column is \textbf{bold}, and the second-best is \underline{underlined}.}
\label{tab:agent_transfer}
\centering
\setlength{\tabcolsep}{2pt}
\resizebox{\textwidth}{!}{%
\begin{tabular}{ll|cc|cccc|cccc|cc}
\toprule
\multirow{4}{*}{\textbf{Group}} &
\multirow{4}{*}{\textbf{Method}} &
\multicolumn{2}{c|}{\textbf{BrowseComp-Plus}} &
\multicolumn{4}{c|}{\textbf{HotpotQA}} &
\multicolumn{4}{c|}{\textbf{MuSiQue}} &
\multirow{4}{*}{\textbf{Avg}} &
\multirow{4}{*}{$\boldsymbol{\Delta_{\mathrm{ID}}}$} \\
\cmidrule(lr){3-4}\cmidrule(lr){5-8}\cmidrule(lr){9-12}
& &
\multicolumn{2}{c|}{\textbf{Qwen3.6-35B-A3B}} &
\multicolumn{2}{c|}{\textbf{GPT-OSS-20B}} &
\multicolumn{2}{c|}{\textbf{Qwen3.6-35B-A3B}} &
\multicolumn{2}{c|}{\textbf{GPT-OSS-20B}} &
\multicolumn{2}{c|}{\textbf{Qwen3.6-35B-A3B}} &
& \\
\cmidrule(lr){3-4}\cmidrule(lr){5-6}\cmidrule(lr){7-8}
\cmidrule(lr){9-10}\cmidrule(lr){11-12}
& &
Tool & ReAct &
Code & ReAct &
Code & ReAct &
Code & ReAct &
Code & ReAct &
& \\
& &
$\to$ ReAct & $\to$ Tool &
$\to$ ReAct & $\to$ Code &
$\to$ ReAct & $\to$ Code &
$\to$ ReAct & $\to$ Code &
$\to$ ReAct & $\to$ Code &
& \\
\midrule
\multirow{2}{*}{\rotatebox[origin=c]{90}{\scriptsize Stat.}} & $N$ steps & .745 {\scriptsize $\pm$ .000} & .873 {\scriptsize $\pm$ .000} & .341 {\scriptsize $\pm$ .020} & .088 {\scriptsize $\pm$ .000} & .465 {\scriptsize $\pm$ .020} & .380 {\scriptsize $\pm$ .000} & .656 {\scriptsize $\pm$ .000} & .192 {\scriptsize $\pm$ .050} & \textbf{.662} {\scriptsize $\pm$ .010} & \textbf{.377} {\scriptsize $\pm$ .010} & .478 {\scriptsize $\pm$ .006} & \textendash \\
  & Tool success rate & .006 {\scriptsize $\pm$ .000} & -.015 {\scriptsize $\pm$ .000} & .503 {\scriptsize $\pm$ .010} & .154 {\scriptsize $\pm$ .010} & .515 {\scriptsize $\pm$ .020} & .369 {\scriptsize $\pm$ .030} & .545 {\scriptsize $\pm$ .010} & .225 {\scriptsize $\pm$ .030} & .493 {\scriptsize $\pm$ .000} & \underline{.336} {\scriptsize $\pm$ .010} & .313 {\scriptsize $\pm$ .005} & \textendash \\
\cmidrule(lr){2-14}
\multirow{7}{*}{\rotatebox[origin=c]{90}{\scriptsize UQ}} & SP & .553 {\scriptsize $\pm$ .000} & .481 {\scriptsize $\pm$ .000} & .106 {\scriptsize $\pm$ .000} & -.060 {\scriptsize $\pm$ .000} & .072 {\scriptsize $\pm$ .000} & .387 {\scriptsize $\pm$ .000} & .010 {\scriptsize $\pm$ .000} & .030 {\scriptsize $\pm$ .000} & .118 {\scriptsize $\pm$ .000} & .257 {\scriptsize $\pm$ .000} & .195 {\scriptsize $\pm$ .000} & \textendash \\
  & Perplexity & .920 {\scriptsize $\pm$ .000} & .877 {\scriptsize $\pm$ .000} & -.211 {\scriptsize $\pm$ .000} & -.067 {\scriptsize $\pm$ .000} & -.116 {\scriptsize $\pm$ .000} & -.220 {\scriptsize $\pm$ .000} & -.213 {\scriptsize $\pm$ .000} & -.051 {\scriptsize $\pm$ .000} & -.160 {\scriptsize $\pm$ .000} & -.021 {\scriptsize $\pm$ .000} & .074 {\scriptsize $\pm$ .000} & \textendash \\
  & MTE & .943 {\scriptsize $\pm$ .000} & .886 {\scriptsize $\pm$ .000} & -.213 {\scriptsize $\pm$ .000} & -.068 {\scriptsize $\pm$ .000} & -.135 {\scriptsize $\pm$ .000} & -.265 {\scriptsize $\pm$ .000} & -.227 {\scriptsize $\pm$ .000} & -.018 {\scriptsize $\pm$ .000} & -.083 {\scriptsize $\pm$ .000} & -.099 {\scriptsize $\pm$ .000} & .072 {\scriptsize $\pm$ .000} & \textendash \\
  & Self-certainty & .924 {\scriptsize $\pm$ .000} & .883 {\scriptsize $\pm$ .000} & .275 {\scriptsize $\pm$ .000} & .095 {\scriptsize $\pm$ .000} & .331 {\scriptsize $\pm$ .000} & .418 {\scriptsize $\pm$ .000} & .222 {\scriptsize $\pm$ .000} & -.010 {\scriptsize $\pm$ .000} & .275 {\scriptsize $\pm$ .000} & .188 {\scriptsize $\pm$ .000} & .360 {\scriptsize $\pm$ .000} & \textendash \\
  & Verb. & .739 {\scriptsize $\pm$ .000} & .687 {\scriptsize $\pm$ .000} & .374 {\scriptsize $\pm$ .000} & .205 {\scriptsize $\pm$ .000} & .577 {\scriptsize $\pm$ .000} & .396 {\scriptsize $\pm$ .000} & .342 {\scriptsize $\pm$ .000} & .074 {\scriptsize $\pm$ .000} & .394 {\scriptsize $\pm$ .000} & .245 {\scriptsize $\pm$ .000} & .403 {\scriptsize $\pm$ .000} & \textendash \\
  & SAUP & .359 {\scriptsize $\pm$ .003} & .437 {\scriptsize $\pm$ .004} & .326 {\scriptsize $\pm$ .000} & .133 {\scriptsize $\pm$ .000} & .120 {\scriptsize $\pm$ .000} & .267 {\scriptsize $\pm$ .000} & .242 {\scriptsize $\pm$ .000} & .262 {\scriptsize $\pm$ .000} & .165 {\scriptsize $\pm$ .000} & .069 {\scriptsize $\pm$ .000} & .238 {\scriptsize $\pm$ .001} & {+.025} \\
  & UProp & .813 {\scriptsize $\pm$ .000} & .817 {\scriptsize $\pm$ .000} & .344 {\scriptsize $\pm$ .000} & \underline{.249} {\scriptsize $\pm$ .000} & .586 {\scriptsize $\pm$ .000} & \textbf{.485} {\scriptsize $\pm$ .000} & .368 {\scriptsize $\pm$ .000} & .199 {\scriptsize $\pm$ .000} & .250 {\scriptsize $\pm$ .000} & .109 {\scriptsize $\pm$ .000} & .422 {\scriptsize $\pm$ .000} & \textendash \\
\cmidrule(lr){2-14}
\multirow{2}{*}{\rotatebox[origin=c]{90}{\scriptsize \shortstack{Super-\\vised}}} & HTC & \underline{.957} {\scriptsize $\pm$ .004} & .910 {\scriptsize $\pm$ .005} & .251 {\scriptsize $\pm$ .000} & -.015 {\scriptsize $\pm$ .000} & .506 {\scriptsize $\pm$ .000} & .412 {\scriptsize $\pm$ .000} & .539 {\scriptsize $\pm$ .000} & .056 {\scriptsize $\pm$ .000} & .139 {\scriptsize $\pm$ .000} & .051 {\scriptsize $\pm$ .000} & .381 {\scriptsize $\pm$ .001} & -.144 \\
  & Logistic regression & \textbf{.988} {\scriptsize $\pm$ .003} & \textbf{.951} {\scriptsize $\pm$ .005} & .216 {\scriptsize $\pm$ .000} & .117 {\scriptsize $\pm$ .000} & .088 {\scriptsize $\pm$ .000} & \underline{.477} {\scriptsize $\pm$ .000} & .555 {\scriptsize $\pm$ .000} & .262 {\scriptsize $\pm$ .000} & -.115 {\scriptsize $\pm$ .000} & .134 {\scriptsize $\pm$ .000} & .367 {\scriptsize $\pm$ .001} & -.142 \\
\cmidrule(lr){2-14}
\multirow{7}{*}{\rotatebox[origin=c]{90}{\scriptsize \textbf{Bayes (ours)}}} & Tool-only & .615 {\scriptsize $\pm$ .038} & -.153 {\scriptsize $\pm$ .028} & \underline{.536} {\scriptsize $\pm$ .000} & .178 {\scriptsize $\pm$ .000} & .554 {\scriptsize $\pm$ .000} & .358 {\scriptsize $\pm$ .000} & .640 {\scriptsize $\pm$ .000} & .248 {\scriptsize $\pm$ .000} & .564 {\scriptsize $\pm$ .000} & .132 {\scriptsize $\pm$ .000} & .367 {\scriptsize $\pm$ .005} & -.043 \\
  & Fused (SP) & .574 {\scriptsize $\pm$ .005} & .572 {\scriptsize $\pm$ .005} & .529 {\scriptsize $\pm$ .000} & .217 {\scriptsize $\pm$ .000} & .545 {\scriptsize $\pm$ .000} & .475 {\scriptsize $\pm$ .000} & .658 {\scriptsize $\pm$ .000} & .276 {\scriptsize $\pm$ .000} & .566 {\scriptsize $\pm$ .000} & .127 {\scriptsize $\pm$ .000} & .454 {\scriptsize $\pm$ .001} & {+.007} \\
  & Fused (Perplexity) & .901 {\scriptsize $\pm$ .001} & .849 {\scriptsize $\pm$ .002} & .526 {\scriptsize $\pm$ .000} & .179 {\scriptsize $\pm$ .000} & .533 {\scriptsize $\pm$ .000} & .451 {\scriptsize $\pm$ .000} & .657 {\scriptsize $\pm$ .000} & .256 {\scriptsize $\pm$ .000} & .552 {\scriptsize $\pm$ .000} & .160 {\scriptsize $\pm$ .000} & .506 {\scriptsize $\pm$ .000} & {+.011} \\
  & Fused (MTE) & .926 {\scriptsize $\pm$ .001} & .863 {\scriptsize $\pm$ .002} & .532 {\scriptsize $\pm$ .000} & .194 {\scriptsize $\pm$ .000} & .529 {\scriptsize $\pm$ .000} & .464 {\scriptsize $\pm$ .000} & \underline{.676} {\scriptsize $\pm$ .000} & \underline{.288} {\scriptsize $\pm$ .000} & .551 {\scriptsize $\pm$ .000} & .164 {\scriptsize $\pm$ .000} & \underline{.519} {\scriptsize $\pm$ .000} & {+.008} \\
  & Fused (Self-certainty) & .890 {\scriptsize $\pm$ .002} & .863 {\scriptsize $\pm$ .001} & .522 {\scriptsize $\pm$ .000} & .236 {\scriptsize $\pm$ .000} & .466 {\scriptsize $\pm$ .000} & \textbf{.485} {\scriptsize $\pm$ .000} & .675 {\scriptsize $\pm$ .000} & .285 {\scriptsize $\pm$ .000} & .512 {\scriptsize $\pm$ .000} & .125 {\scriptsize $\pm$ .000} & .506 {\scriptsize $\pm$ .000} & -.007 \\
  & Fused (Verb.) & .938 {\scriptsize $\pm$ .002} & \underline{.932} {\scriptsize $\pm$ .011} & \textbf{.588} {\scriptsize $\pm$ .000} & \textbf{.321} {\scriptsize $\pm$ .000} & \textbf{.639} {\scriptsize $\pm$ .000} & .474 {\scriptsize $\pm$ .000} & \textbf{.725} {\scriptsize $\pm$ .000} & \textbf{.442} {\scriptsize $\pm$ .000} & \underline{.596} {\scriptsize $\pm$ .000} & .144 {\scriptsize $\pm$ .000} & \textbf{.580} {\scriptsize $\pm$ .001} & {+.034} \\
  & SBF & .951 {\scriptsize $\pm$ .001} & .894 {\scriptsize $\pm$ .001} & .401 {\scriptsize $\pm$ .000} & .129 {\scriptsize $\pm$ .000} & \underline{.603} {\scriptsize $\pm$ .000} & .475 {\scriptsize $\pm$ .000} & .663 {\scriptsize $\pm$ .000} & .280 {\scriptsize $\pm$ .000} & -.117 {\scriptsize $\pm$ .000} & .097 {\scriptsize $\pm$ .000} & .437 {\scriptsize $\pm$ .000} & -.100 \\
\bottomrule
\end{tabular}
}
\end{table*}

\subsection{Joint Model and Agent Transfer}
\label{sec:joint_transfer}

\begin{table*}[h]
\caption{\textbf{Joint model and harness transfer.} PRR@0.5$\uparrow$ with the dataset held fixed. Column headers specify the source $\to$ target model/harness pair. GPT and Qwen denote GPT-OSS-20B and Qwen3.6-35B-A3B, respectively, and Code denotes smolagents CodeAgent. $\Delta_{\mathrm{ID}}$ is Avg minus the mean in-domain PRR for the same targets. In-domain scores are reused for $N$ steps, tool success rate, and raw UQ. The best result in each column is \textbf{bold}, and the second-best is \underline{underlined}.}
\label{tab:model_agent_transfer}
\centering
\setlength{\tabcolsep}{2pt}
\resizebox{\textwidth}{!}{%
\begin{tabular}{ll|cccc|cccc|cc}
\toprule
\multirow{4}{*}{\textbf{Group}} & \multirow{4}{*}{\textbf{Method}} & \multicolumn{4}{c|}{\textbf{HotpotQA}} & \multicolumn{4}{c|}{\textbf{MuSiQue}} & \multirow{4}{*}{\shortstack{\textbf{Avg}}} & \multirow{4}{*}{$\boldsymbol{\Delta_{\mathrm{ID}}}$} \\
\cmidrule(lr){3-6}\cmidrule(lr){7-10}
& & \shortstack{GPT / Code\\$\to$\\Qwen / ReAct} & \shortstack{Qwen / ReAct\\$\to$\\GPT / Code} & \shortstack{GPT / ReAct\\$\to$\\Qwen / Code} & \shortstack{Qwen / Code\\$\to$\\GPT / ReAct} & \shortstack{GPT / Code\\$\to$\\Qwen / ReAct} & \shortstack{Qwen / ReAct\\$\to$\\GPT / Code} & \shortstack{GPT / ReAct\\$\to$\\Qwen / Code} & \shortstack{Qwen / Code\\$\to$\\GPT / ReAct} & & \\
\midrule
\multirow{2}{*}{\rotatebox[origin=c]{90}{\scriptsize Stat.}} & $N$ steps & .465 {\scriptsize $\pm$ .020} & .088 {\scriptsize $\pm$ .000} & .380 {\scriptsize $\pm$ .000} & .341 {\scriptsize $\pm$ .020} & \textbf{.662} {\scriptsize $\pm$ .010} & .192 {\scriptsize $\pm$ .050} & \textbf{.377} {\scriptsize $\pm$ .010} & \underline{.656} {\scriptsize $\pm$ .000} & .395 {\scriptsize $\pm$ .007} & \textendash \\
  & Tool success rate & .515 {\scriptsize $\pm$ .020} & .154 {\scriptsize $\pm$ .010} & .369 {\scriptsize $\pm$ .030} & .503 {\scriptsize $\pm$ .010} & .493 {\scriptsize $\pm$ .000} & .225 {\scriptsize $\pm$ .030} & \underline{.336} {\scriptsize $\pm$ .010} & .545 {\scriptsize $\pm$ .010} & .392 {\scriptsize $\pm$ .006} & \textendash \\
\cmidrule(lr){2-12}
\multirow{7}{*}{\rotatebox[origin=c]{90}{\scriptsize UQ}} & SP & .072 {\scriptsize $\pm$ .000} & -.060 {\scriptsize $\pm$ .000} & .387 {\scriptsize $\pm$ .000} & .106 {\scriptsize $\pm$ .000} & .118 {\scriptsize $\pm$ .000} & .030 {\scriptsize $\pm$ .000} & .257 {\scriptsize $\pm$ .000} & .010 {\scriptsize $\pm$ .000} & .115 {\scriptsize $\pm$ .000} & \textendash \\
  & Perplexity & -.116 {\scriptsize $\pm$ .000} & -.067 {\scriptsize $\pm$ .000} & -.220 {\scriptsize $\pm$ .000} & -.211 {\scriptsize $\pm$ .000} & -.160 {\scriptsize $\pm$ .000} & -.051 {\scriptsize $\pm$ .000} & -.021 {\scriptsize $\pm$ .000} & -.213 {\scriptsize $\pm$ .000} & -.132 {\scriptsize $\pm$ .000} & \textendash \\
  & MTE & -.135 {\scriptsize $\pm$ .000} & -.068 {\scriptsize $\pm$ .000} & -.265 {\scriptsize $\pm$ .000} & -.213 {\scriptsize $\pm$ .000} & -.083 {\scriptsize $\pm$ .000} & -.018 {\scriptsize $\pm$ .000} & -.099 {\scriptsize $\pm$ .000} & -.227 {\scriptsize $\pm$ .000} & -.138 {\scriptsize $\pm$ .000} & \textendash \\
  & Self-certainty & .331 {\scriptsize $\pm$ .000} & .095 {\scriptsize $\pm$ .000} & .418 {\scriptsize $\pm$ .000} & .275 {\scriptsize $\pm$ .000} & .275 {\scriptsize $\pm$ .000} & -.010 {\scriptsize $\pm$ .000} & .188 {\scriptsize $\pm$ .000} & .222 {\scriptsize $\pm$ .000} & .224 {\scriptsize $\pm$ .000} & \textendash \\
  & Verb. & .577 {\scriptsize $\pm$ .000} & .205 {\scriptsize $\pm$ .000} & .396 {\scriptsize $\pm$ .000} & .374 {\scriptsize $\pm$ .000} & .394 {\scriptsize $\pm$ .000} & .074 {\scriptsize $\pm$ .000} & .245 {\scriptsize $\pm$ .000} & .342 {\scriptsize $\pm$ .000} & .326 {\scriptsize $\pm$ .000} & \textendash \\
  & SAUP & .156 {\scriptsize $\pm$ .000} & .153 {\scriptsize $\pm$ .000} & .263 {\scriptsize $\pm$ .000} & .285 {\scriptsize $\pm$ .000} & .176 {\scriptsize $\pm$ .000} & .106 {\scriptsize $\pm$ .000} & .097 {\scriptsize $\pm$ .000} & .234 {\scriptsize $\pm$ .000} & .184 {\scriptsize $\pm$ .000} & {+.018} \\
  & UProp & \underline{.586} {\scriptsize $\pm$ .000} & \underline{.249} {\scriptsize $\pm$ .000} & \textbf{.485} {\scriptsize $\pm$ .000} & .344 {\scriptsize $\pm$ .000} & .250 {\scriptsize $\pm$ .000} & .199 {\scriptsize $\pm$ .000} & .109 {\scriptsize $\pm$ .000} & .368 {\scriptsize $\pm$ .000} & .324 {\scriptsize $\pm$ .000} & \textendash \\
\cmidrule(lr){2-12}
\multirow{2}{*}{\rotatebox[origin=c]{90}{\scriptsize \shortstack{Super-\\vised}}} & HTC & .450 {\scriptsize $\pm$ .000} & .205 {\scriptsize $\pm$ .000} & .420 {\scriptsize $\pm$ .000} & .256 {\scriptsize $\pm$ .000} & .081 {\scriptsize $\pm$ .000} & -.164 {\scriptsize $\pm$ .000} & .102 {\scriptsize $\pm$ .000} & .077 {\scriptsize $\pm$ .000} & .178 {\scriptsize $\pm$ .000} & -.241 \\
  & Logistic regression & .135 {\scriptsize $\pm$ .000} & -.034 {\scriptsize $\pm$ .000} & .209 {\scriptsize $\pm$ .000} & -.058 {\scriptsize $\pm$ .000} & .092 {\scriptsize $\pm$ .000} & .030 {\scriptsize $\pm$ .000} & .050 {\scriptsize $\pm$ .000} & -.040 {\scriptsize $\pm$ .000} & .048 {\scriptsize $\pm$ .000} & -.344 \\
\cmidrule(lr){2-12}
\multirow{7}{*}{\rotatebox[origin=c]{90}{\scriptsize \textbf{Bayes (ours)}}} & Tool-only & .577 {\scriptsize $\pm$ .000} & .177 {\scriptsize $\pm$ .000} & .358 {\scriptsize $\pm$ .000} & \underline{.515} {\scriptsize $\pm$ .000} & .600 {\scriptsize $\pm$ .000} & .247 {\scriptsize $\pm$ .000} & .126 {\scriptsize $\pm$ .000} & .623 {\scriptsize $\pm$ .000} & \underline{.403} {\scriptsize $\pm$ .000} & {+.002} \\
  & Fused (SP) & .542 {\scriptsize $\pm$ .000} & .143 {\scriptsize $\pm$ .000} & .367 {\scriptsize $\pm$ .000} & .506 {\scriptsize $\pm$ .000} & .602 {\scriptsize $\pm$ .000} & \underline{.294} {\scriptsize $\pm$ .000} & .143 {\scriptsize $\pm$ .000} & .599 {\scriptsize $\pm$ .000} & .399 {\scriptsize $\pm$ .000} & -.014 \\
  & Fused (Perplexity) & .530 {\scriptsize $\pm$ .000} & .106 {\scriptsize $\pm$ .000} & .323 {\scriptsize $\pm$ .000} & .405 {\scriptsize $\pm$ .000} & .608 {\scriptsize $\pm$ .000} & .153 {\scriptsize $\pm$ .000} & .180 {\scriptsize $\pm$ .000} & .506 {\scriptsize $\pm$ .000} & .351 {\scriptsize $\pm$ .000} & -.048 \\
  & Fused (MTE) & .529 {\scriptsize $\pm$ .000} & .126 {\scriptsize $\pm$ .000} & .336 {\scriptsize $\pm$ .000} & .505 {\scriptsize $\pm$ .000} & .620 {\scriptsize $\pm$ .000} & .255 {\scriptsize $\pm$ .000} & .170 {\scriptsize $\pm$ .000} & .645 {\scriptsize $\pm$ .000} & .398 {\scriptsize $\pm$ .000} & -.016 \\
  & Fused (Self-certainty) & .528 {\scriptsize $\pm$ .000} & .123 {\scriptsize $\pm$ .000} & .382 {\scriptsize $\pm$ .000} & .505 {\scriptsize $\pm$ .000} & .611 {\scriptsize $\pm$ .000} & .254 {\scriptsize $\pm$ .000} & .167 {\scriptsize $\pm$ .000} & \underline{.656} {\scriptsize $\pm$ .000} & \underline{.403} {\scriptsize $\pm$ .000} & -.018 \\
  & Fused (Verb.) & \textbf{.648} {\scriptsize $\pm$ .000} & \textbf{.319} {\scriptsize $\pm$ .000} & \underline{.474} {\scriptsize $\pm$ .000} & \textbf{.603} {\scriptsize $\pm$ .000} & .599 {\scriptsize $\pm$ .000} & \textbf{.404} {\scriptsize $\pm$ .000} & .166 {\scriptsize $\pm$ .000} & \textbf{.675} {\scriptsize $\pm$ .000} & \textbf{.486} {\scriptsize $\pm$ .000} & {+.038} \\
  & SBF & .537 {\scriptsize $\pm$ .000} & .149 {\scriptsize $\pm$ .000} & .473 {\scriptsize $\pm$ .000} & .511 {\scriptsize $\pm$ .000} & \underline{.652} {\scriptsize $\pm$ .000} & .174 {\scriptsize $\pm$ .000} & .015 {\scriptsize $\pm$ .000} & .141 {\scriptsize $\pm$ .000} & .331 {\scriptsize $\pm$ .000} & -.108 \\
\bottomrule
\end{tabular}
}
\end{table*}

\subsection{Environment Transfer}
\label{sec:env_transfer}

\begin{table*}[h]
\caption{\textbf{Cross-dataset transfer in coding.} PRR@0.5$\uparrow$ with the model and harness held fixed. Methods are trained on the other coding datasets and evaluated on the held-out dataset named in each column. $\Delta_{\mathrm{ID}}$ is the average PRR change relative to the corresponding in-domain results on the target datasets. The best result in each column is \textbf{bold}, and the second-best is \underline{underlined}.}
\label{tab:dataset_transfer}
\centering
\resizebox{\textwidth}{!}{%
\begin{tabular}{ll|cccc|cc}
\toprule
\textbf{Group} & \textbf{Method} & \textbf{LCB} & \textbf{BCB} & \textbf{SWE} & \textbf{CC} & \textbf{Avg (OOD)} & $\boldsymbol{\Delta_{\mathrm{ID}}}$ \\
\midrule
\multirow{2}{*}{\rotatebox[origin=c]{90}{\scriptsize Stat.}} & $N$ steps & .080 {\scriptsize $\pm$ .000} & .070 {\scriptsize $\pm$ .000} & .130 {\scriptsize $\pm$ .000} & .260 {\scriptsize $\pm$ .000} & .135 {\scriptsize $\pm$ .000} & \textendash \\
 & Tool success rate & .170 {\scriptsize $\pm$ .000} & .130 {\scriptsize $\pm$ .000} & .090 {\scriptsize $\pm$ .000} & .420 {\scriptsize $\pm$ .000} & .203 {\scriptsize $\pm$ .000} & \textendash \\
\cmidrule(lr){2-8}
\multirow{7}{*}{\rotatebox[origin=c]{90}{\scriptsize UQ}} & SP & \textbf{.280} {\scriptsize $\pm$ .000} & \underline{.250} {\scriptsize $\pm$ .000} & \textbf{.360} {\scriptsize $\pm$ .000} & \underline{.550} {\scriptsize $\pm$ .000} & \textbf{.360} {\scriptsize $\pm$ .000} & \textendash \\
 & Perplexity & .050 {\scriptsize $\pm$ .000} & .190 {\scriptsize $\pm$ .000} & .220 {\scriptsize $\pm$ .000} & .200 {\scriptsize $\pm$ .000} & .165 {\scriptsize $\pm$ .000} & \textendash \\
 & MTE & .060 {\scriptsize $\pm$ .000} & .190 {\scriptsize $\pm$ .000} & .220 {\scriptsize $\pm$ .000} & .210 {\scriptsize $\pm$ .000} & .170 {\scriptsize $\pm$ .000} & \textendash \\
 & Self-certainty & .170 {\scriptsize $\pm$ .000} & .170 {\scriptsize $\pm$ .000} & .270 {\scriptsize $\pm$ .000} & .390 {\scriptsize $\pm$ .000} & .250 {\scriptsize $\pm$ .000} & \textendash \\
 & Verbalized confidence & .110 {\scriptsize $\pm$ .000} & .160 {\scriptsize $\pm$ .000} & .320 {\scriptsize $\pm$ .000} & .410 {\scriptsize $\pm$ .000} & .250 {\scriptsize $\pm$ .000} & \textendash \\
 & SAUP & .070 {\scriptsize $\pm$ .000} & .140 {\scriptsize $\pm$ .000} & .190 {\scriptsize $\pm$ .000} & .230 {\scriptsize $\pm$ .000} & .158 {\scriptsize $\pm$ .000} & {+.002} \\
 & UProp & .070 {\scriptsize $\pm$ .000} & \textbf{.270} {\scriptsize $\pm$ .000} & .170 {\scriptsize $\pm$ .000} & .270 {\scriptsize $\pm$ .000} & .195 {\scriptsize $\pm$ .000} & \textendash \\
\cmidrule(lr){2-8}
\multirow{2}{*}{\rotatebox[origin=c]{90}{\scriptsize \shortstack{Super-\\vised}}} & HTC & .080 {\scriptsize $\pm$ .020} & -.080 {\scriptsize $\pm$ .050} & .170 {\scriptsize $\pm$ .020} & .170 {\scriptsize $\pm$ .070} & .085 {\scriptsize $\pm$ .023} & -.101 \\
 & Logistic regression & \underline{.270} {\scriptsize $\pm$ .010} & .220 {\scriptsize $\pm$ .010} & \underline{.340} {\scriptsize $\pm$ .010} & .540 {\scriptsize $\pm$ .020} & \underline{.343} {\scriptsize $\pm$ .007} & {+.015} \\
\cmidrule(lr){2-8}
\multirow{7}{*}{\rotatebox[origin=c]{90}{\scriptsize \textbf{Bayes (ours)}}} & Tool-only & .190 {\scriptsize $\pm$ .010} & .190 {\scriptsize $\pm$ .010} & .160 {\scriptsize $\pm$ .010} & .380 {\scriptsize $\pm$ .030} & .230 {\scriptsize $\pm$ .009} & {+.013} \\
 & Fused (SP) & \textbf{.280} {\scriptsize $\pm$ .000} & .160 {\scriptsize $\pm$ .010} & .270 {\scriptsize $\pm$ .000} & \textbf{.580} {\scriptsize $\pm$ .010} & .323 {\scriptsize $\pm$ .004} & -.048 \\
 & Fused (Perplexity) & .130 {\scriptsize $\pm$ .000} & .180 {\scriptsize $\pm$ .010} & .240 {\scriptsize $\pm$ .010} & .260 {\scriptsize $\pm$ .000} & .203 {\scriptsize $\pm$ .004} & -.013 \\
 & Fused (MTE) & .140 {\scriptsize $\pm$ .000} & .180 {\scriptsize $\pm$ .000} & .250 {\scriptsize $\pm$ .000} & .270 {\scriptsize $\pm$ .010} & .210 {\scriptsize $\pm$ .003} & -.015 \\
 & Fused (Self-certainty) & .210 {\scriptsize $\pm$ .000} & .170 {\scriptsize $\pm$ .000} & .280 {\scriptsize $\pm$ .010} & .400 {\scriptsize $\pm$ .010} & .265 {\scriptsize $\pm$ .004} & -.013 \\
 & Fused (Verb.) & .120 {\scriptsize $\pm$ .000} & .160 {\scriptsize $\pm$ .000} & \underline{.340} {\scriptsize $\pm$ .010} & .380 {\scriptsize $\pm$ .010} & .250 {\scriptsize $\pm$ .004} & -.017 \\
 & SBF & .200 {\scriptsize $\pm$ .010} & .190 {\scriptsize $\pm$ .010} & \underline{.340} {\scriptsize $\pm$ .000} & .490 {\scriptsize $\pm$ .010} & .305 {\scriptsize $\pm$ .004} & -.034 \\
\bottomrule
\end{tabular}
}
\end{table*}

\begin{table*}[h]
\caption{\textbf{Cross-environment transfer.} PRR@0.5$\uparrow$ for WebArena $\leftrightarrow$ OSWorld using Qwen3.6-35B-A3B. $\Delta_{\mathrm{ID}}$ is the average PRR change relative to the corresponding in-domain results. The best result in each column is \textbf{bold}, and the second-best is \underline{underlined}.}
\label{tab:env_transfer}
\centering
\resizebox{0.7\textwidth}{!}{%
\begin{tabular}{ll|cc|cc}
\toprule
\multirow{2}{*}{\textbf{Group}} &
\multirow{2}{*}{\textbf{Method}} &
\textbf{WebArena} &
\textbf{OSWorld} &
\multirow{2}{*}{\textbf{Avg}} &
\multirow{2}{*}{$\boldsymbol{\Delta_{\mathrm{ID}}}$} \\
& &
$\to$ \textbf{OSWorld} &
$\to$ \textbf{WebArena} &
& \\
\midrule
\multirow{2}{*}{\rotatebox[origin=c]{90}{\scriptsize Stat.}} & $N$ steps & \textbf{.352} {\scriptsize $\pm$ .000} & .369 {\scriptsize $\pm$ .000} & \textbf{.361} {\scriptsize $\pm$ .000} & \textendash \\
 & Tool success rate & -.005 {\scriptsize $\pm$ .000} & .306 {\scriptsize $\pm$ .000} & .151 {\scriptsize $\pm$ .000} & \textendash \\
\cmidrule(lr){2-6}
\multirow{7}{*}{\rotatebox[origin=c]{90}{\scriptsize UQ}} & SP & .224 {\scriptsize $\pm$ .000} & .185 {\scriptsize $\pm$ .000} & .205 {\scriptsize $\pm$ .000} & \textendash \\
 & Perplexity & .037 {\scriptsize $\pm$ .000} & .364 {\scriptsize $\pm$ .000} & .200 {\scriptsize $\pm$ .000} & \textendash \\
 & MTE & .058 {\scriptsize $\pm$ .000} & \underline{.373} {\scriptsize $\pm$ .000} & .215 {\scriptsize $\pm$ .000} & \textendash \\
 & Self-certainty & .282 {\scriptsize $\pm$ .000} & \textbf{.397} {\scriptsize $\pm$ .000} & \underline{.340} {\scriptsize $\pm$ .000} & \textendash \\
 & Verb. & .283 {\scriptsize $\pm$ .000} & .346 {\scriptsize $\pm$ .000} & .315 {\scriptsize $\pm$ .000} & \textendash \\
 & SAUP & .239 {\scriptsize $\pm$ .007} & .100 {\scriptsize $\pm$ .029} & .169 {\scriptsize $\pm$ .015} & {+.046} \\
 & UProp & .047 {\scriptsize $\pm$ .000} & .315 {\scriptsize $\pm$ .000} & .181 {\scriptsize $\pm$ .000} & \textendash \\
\cmidrule(lr){2-6}
\multirow{2}{*}{\rotatebox[origin=c]{90}{\scriptsize \shortstack{Super-\\vised}}} & HTC & .073 {\scriptsize $\pm$ .046} & .206 {\scriptsize $\pm$ .005} & .140 {\scriptsize $\pm$ .023} & -.258 \\
 & Logistic regression & .281 {\scriptsize $\pm$ .025} & .220 {\scriptsize $\pm$ .011} & .251 {\scriptsize $\pm$ .014} & -.220 \\
\cmidrule(lr){2-6}
\multirow{7}{*}{\rotatebox[origin=c]{90}{\scriptsize \textbf{Bayes (ours)}}} & Tool-only & .073 {\scriptsize $\pm$ .063} & .065 {\scriptsize $\pm$ .017} & .069 {\scriptsize $\pm$ .033} & -.174 \\
 & Fused (SP) & .213 {\scriptsize $\pm$ .003} & .225 {\scriptsize $\pm$ .004} & .219 {\scriptsize $\pm$ .003} & -.123 \\
 & Fused (Perplexity) & .170 {\scriptsize $\pm$ .028} & .289 {\scriptsize $\pm$ .040} & .230 {\scriptsize $\pm$ .024} & -.062 \\
 & Fused (MTE) & .212 {\scriptsize $\pm$ .005} & .337 {\scriptsize $\pm$ .009} & .275 {\scriptsize $\pm$ .005} & {-.013} \\
 & Fused (Self-certainty) & \underline{.319} {\scriptsize $\pm$ .013} & .256 {\scriptsize $\pm$ .063} & .288 {\scriptsize $\pm$ .032} & -.076 \\
 & Fused (Verb.) & .314 {\scriptsize $\pm$ .020} & .307 {\scriptsize $\pm$ .011} & .311 {\scriptsize $\pm$ .011} & -.017 \\
 & SBF & .020 {\scriptsize $\pm$ .046} & .360 {\scriptsize $\pm$ .003} & .190 {\scriptsize $\pm$ .023} & -.227 \\
\bottomrule
\end{tabular}
}
\end{table*}

\subsection{Domain Transfer}
\label{sec:domain_transfer}

\begin{table*}[h]
\caption{\textbf{Cross-domain transfer.} PRR@0.5$\uparrow$ for ALFWorld $\leftrightarrow$ DeepSWE. Differences from in-domain performance also reflect changes in the available features: transfer uses three shared critics, while in-domain evaluation uses all environment-specific critics. The best result in each column is \textbf{bold}, and the second-best is \underline{underlined}.}
\label{tab:full-shift-alfworld-deepswe}
\label{tab:full-shift-deepswe-alfworld}
\centering
\setlength{\tabcolsep}{2pt}
\resizebox{\textwidth}{!}{%
\begin{tabular}{ll|cc|cc|cc|cc|cc}
\toprule
\multirow{4}{*}{\textbf{Group}} & \multirow{4}{*}{\textbf{Method}} & \multicolumn{4}{c|}{\textbf{ALFWorld $\to$ DeepSWE}} & \multicolumn{4}{c|}{\textbf{DeepSWE $\to$ ALFWorld}} & \multirow{4}{*}{\shortstack{\textbf{Avg}}} & \multirow{4}{*}{$\boldsymbol{\Delta_{\mathrm{ID}}}$} \\
\cmidrule(lr){3-6}\cmidrule(lr){7-10}
& & \multicolumn{2}{c|}{\textbf{GPT-OSS-20B}} & \multicolumn{2}{c|}{\textbf{Qwen3.6-35B-A3B}} & \multicolumn{2}{c|}{\textbf{GPT-OSS-20B}} & \multicolumn{2}{c|}{\textbf{Qwen3.6-35B-A3B}} & & \\
\cmidrule(lr){3-4}\cmidrule(lr){5-6}\cmidrule(lr){7-8}\cmidrule(lr){9-10}
 &  & \shortstack{ReAct $\to$\\mini-swe-agent} & \shortstack{smolagents $\to$\\mini-swe-agent} & \shortstack{ReAct $\to$\\mini-swe-agent} & \shortstack{smolagents $\to$\\mini-swe-agent} & \shortstack{mini-swe-agent\\$\to$ ReAct} & \shortstack{mini-swe-agent\\$\to$ smolagents} & \shortstack{mini-swe-agent\\$\to$ ReAct} & \shortstack{mini-swe-agent\\$\to$ smolagents} &  &  \\
\midrule
\multirow{2}{*}{\rotatebox[origin=c]{90}{\scriptsize Stat.}} & $N$ steps & .170 {\scriptsize $\pm$ .000} & .170 {\scriptsize $\pm$ .000} & -.110 {\scriptsize $\pm$ .000} & -.110 {\scriptsize $\pm$ .000} & \underline{.640} {\scriptsize $\pm$ .000} & .570 {\scriptsize $\pm$ .000} & \textbf{.490} {\scriptsize $\pm$ .000} & -.150 {\scriptsize $\pm$ .000} & .209 {\scriptsize $\pm$ .000} & \textendash \\
  & Tool success rate & -.080 {\scriptsize $\pm$ .000} & -.080 {\scriptsize $\pm$ .000} & .120 {\scriptsize $\pm$ .000} & .120 {\scriptsize $\pm$ .000} & .620 {\scriptsize $\pm$ .000} & \textbf{.930} {\scriptsize $\pm$ .000} & .120 {\scriptsize $\pm$ .000} & .070 {\scriptsize $\pm$ .000} & \textbf{.228} {\scriptsize $\pm$ .000} & \textendash \\
\cmidrule(lr){2-12}
\multirow{7}{*}{\rotatebox[origin=c]{90}{\scriptsize UQ}} & SP & -.030 {\scriptsize $\pm$ .000} & -.030 {\scriptsize $\pm$ .000} & \underline{.240} {\scriptsize $\pm$ .000} & \underline{.240} {\scriptsize $\pm$ .000} & -.080 {\scriptsize $\pm$ .000} & .570 {\scriptsize $\pm$ .000} & \underline{.450} {\scriptsize $\pm$ .000} & .350 {\scriptsize $\pm$ .000} & \underline{.214} {\scriptsize $\pm$ .000} & \textendash \\
  & Perplexity & -.070 {\scriptsize $\pm$ .000} & -.070 {\scriptsize $\pm$ .000} & \textbf{.250} {\scriptsize $\pm$ .000} & \textbf{.250} {\scriptsize $\pm$ .000} & -.050 {\scriptsize $\pm$ .000} & .510 {\scriptsize $\pm$ .000} & -.010 {\scriptsize $\pm$ .000} & .550 {\scriptsize $\pm$ .000} & .170 {\scriptsize $\pm$ .000} & \textendash \\
  & MTE & -.020 {\scriptsize $\pm$ .000} & -.020 {\scriptsize $\pm$ .000} & .210 {\scriptsize $\pm$ .000} & .210 {\scriptsize $\pm$ .000} & -.150 {\scriptsize $\pm$ .000} & .750 {\scriptsize $\pm$ .000} & .030 {\scriptsize $\pm$ .000} & \textbf{.610} {\scriptsize $\pm$ .000} & .203 {\scriptsize $\pm$ .000} & \textendash \\
  & Self-certainty & .030 {\scriptsize $\pm$ .000} & .030 {\scriptsize $\pm$ .000} & .040 {\scriptsize $\pm$ .000} & .040 {\scriptsize $\pm$ .000} & -.030 {\scriptsize $\pm$ .000} & \textbf{.930} {\scriptsize $\pm$ .000} & -.130 {\scriptsize $\pm$ .000} & \underline{.590} {\scriptsize $\pm$ .000} & .188 {\scriptsize $\pm$ .000} & \textendash \\
  & Verb. confidence & \underline{.200} {\scriptsize $\pm$ .000} & \underline{.200} {\scriptsize $\pm$ .000} & -.180 {\scriptsize $\pm$ .000} & -.180 {\scriptsize $\pm$ .000} & .190 {\scriptsize $\pm$ .000} & \underline{.760} {\scriptsize $\pm$ .000} & -.160 {\scriptsize $\pm$ .000} & -.040 {\scriptsize $\pm$ .000} & .099 {\scriptsize $\pm$ .000} & \textendash \\
  & SAUP & -.080 {\scriptsize $\pm$ .070} & -.070 {\scriptsize $\pm$ .040} & .120 {\scriptsize $\pm$ .120} & .180 {\scriptsize $\pm$ .020} & -.100 {\scriptsize $\pm$ .080} & .540 {\scriptsize $\pm$ .140} & -.190 {\scriptsize $\pm$ .080} & .000 {\scriptsize $\pm$ .040} & .050 {\scriptsize $\pm$ .029} & -.144 \\
  & UProp & .180 {\scriptsize $\pm$ .000} & .180 {\scriptsize $\pm$ .000} & -.170 {\scriptsize $\pm$ .000} & -.170 {\scriptsize $\pm$ .000} & \textbf{.710} {\scriptsize $\pm$ .000} & .430 {\scriptsize $\pm$ .000} & .140 {\scriptsize $\pm$ .000} & .340 {\scriptsize $\pm$ .000} & .205 {\scriptsize $\pm$ .000} & \textendash \\
\cmidrule(lr){2-12}
\multirow{2}{*}{\rotatebox[origin=c]{90}{\scriptsize \shortstack{Super-\\vised}}} & HTC & .040 {\scriptsize $\pm$ .070} & -.040 {\scriptsize $\pm$ .050} & .020 {\scriptsize $\pm$ .070} & -.090 {\scriptsize $\pm$ .070} & .170 {\scriptsize $\pm$ .070} & .470 {\scriptsize $\pm$ .170} & .050 {\scriptsize $\pm$ .050} & .030 {\scriptsize $\pm$ .170} & .081 {\scriptsize $\pm$ .036} & -.230 \\
  & Logistic regression & .020 {\scriptsize $\pm$ .040} & -.030 {\scriptsize $\pm$ .030} & -.070 {\scriptsize $\pm$ .130} & -.040 {\scriptsize $\pm$ .030} & -.070 {\scriptsize $\pm$ .080} & -.210 {\scriptsize $\pm$ .080} & .130 {\scriptsize $\pm$ .030} & .040 {\scriptsize $\pm$ .040} & -.029 {\scriptsize $\pm$ .024} & -.499 \\
\cmidrule(lr){2-12}
\multirow{7}{*}{\rotatebox[origin=c]{90}{\scriptsize \textbf{Bayes (ours)}}} & Tool-only & .030 {\scriptsize $\pm$ .100} & -.190 {\scriptsize $\pm$ .020} & .110 {\scriptsize $\pm$ .040} & -.110 {\scriptsize $\pm$ .050} & -.170 {\scriptsize $\pm$ .060} & -.700 {\scriptsize $\pm$ .070} & -.390 {\scriptsize $\pm$ .020} & -.210 {\scriptsize $\pm$ .030} & -.204 {\scriptsize $\pm$ .019} & -.638 \\
  & Fused (SP) & .030 {\scriptsize $\pm$ .040} & -.020 {\scriptsize $\pm$ .010} & -.120 {\scriptsize $\pm$ .040} & -.030 {\scriptsize $\pm$ .030} & -.090 {\scriptsize $\pm$ .060} & -.310 {\scriptsize $\pm$ .140} & -.090 {\scriptsize $\pm$ .120} & -.010 {\scriptsize $\pm$ .060} & -.080 {\scriptsize $\pm$ .027} & -.461 \\
  & Fused (Perplexity) & .020 {\scriptsize $\pm$ .080} & -.070 {\scriptsize $\pm$ .010} & -.090 {\scriptsize $\pm$ .010} & -.040 {\scriptsize $\pm$ .060} & -.110 {\scriptsize $\pm$ .040} & -.340 {\scriptsize $\pm$ .190} & -.040 {\scriptsize $\pm$ .140} & -.010 {\scriptsize $\pm$ .060} & -.085 {\scriptsize $\pm$ .033} & -.469 \\
  & Fused (MTE) & \textbf{.270} {\scriptsize $\pm$ .020} & -.070 {\scriptsize $\pm$ .000} & -.010 {\scriptsize $\pm$ .030} & .150 {\scriptsize $\pm$ .090} & -.120 {\scriptsize $\pm$ .040} & -.420 {\scriptsize $\pm$ .210} & -.050 {\scriptsize $\pm$ .100} & .230 {\scriptsize $\pm$ .090} & -.003 {\scriptsize $\pm$ .034} & {-.368} \\
  & Fused (Self-certainty) & -.060 {\scriptsize $\pm$ .080} & .040 {\scriptsize $\pm$ .030} & .000 {\scriptsize $\pm$ .040} & -.020 {\scriptsize $\pm$ .000} & -.060 {\scriptsize $\pm$ .100} & -.560 {\scriptsize $\pm$ .100} & .010 {\scriptsize $\pm$ .040} & -.060 {\scriptsize $\pm$ .080} & -.089 {\scriptsize $\pm$ .024} & -.441 \\
  & Fused (Verb.) & .190 {\scriptsize $\pm$ .020} & \textbf{.210} {\scriptsize $\pm$ .000} & -.100 {\scriptsize $\pm$ .040} & -.050 {\scriptsize $\pm$ .020} & .090 {\scriptsize $\pm$ .050} & -.050 {\scriptsize $\pm$ .110} & .120 {\scriptsize $\pm$ .050} & .110 {\scriptsize $\pm$ .000} & .065 {\scriptsize $\pm$ .017} & {-.263} \\
  & SBF & .050 {\scriptsize $\pm$ .020} & -.010 {\scriptsize $\pm$ .050} & -.020 {\scriptsize $\pm$ .000} & -.030 {\scriptsize $\pm$ .030} & .340 {\scriptsize $\pm$ .080} & .380 {\scriptsize $\pm$ .040} & -.180 {\scriptsize $\pm$ .030} & .230 {\scriptsize $\pm$ .020} & .095 {\scriptsize $\pm$ .014} & -.405 \\
\bottomrule
\end{tabular}
}
\end{table*}

\clearpage
\section{Ablations}
\label{app:ablations}

\ecoparagraph{Aggregation of UQ scores over the trajectory.} 
\Cref{tab:raw-aggregation-retrieval,tab:raw-aggregation-coding,tab:raw-aggregation-interactive} compare three ways to aggregate step-level raw UQ scores over the trajectory: the last step, the mean, and the maximum over steps. No aggregation is uniformly best, but the preferred choice is consistent within a task family. On interactive and retrieval tasks, the last step is usually best, since the final step typically contains the answer or terminating action, e.g., on WebArena it is best for every UQ method. On coding, the mean is best on average, as errors can arise at any step that writes or edits code. The maximum is typically the weakest choice outside coding (e.g., SP drops to 0.007 on retrieval). We therefore use the last step for retrieval and interactive tasks and the mean for coding.

\begin{table}[h]
\caption{\textbf{Aggregation of raw UQ scores: retrieval tasks.} In-domain PRR@0.5$\uparrow$ of raw uncertainty scores aggregated over the trajectory by the last step, the mean, or the maximum over steps, averaged over the model and harness configurations of each benchmark except Qwen2.5-32B-Instruct. Avg is the mean over the benchmarks. The best aggregation for each score and column is \textbf{bold}.}
\label{tab:raw-aggregation-retrieval}
\centering
\small
\setlength{\tabcolsep}{4pt}
\begin{tabular}{ll|cccc|c}
\toprule
\textbf{Score} & \textbf{Agg.} & \textbf{HotpotQA} & \textbf{MuSiQue} & \textbf{CRAG} & \textbf{BrowseComp-Plus} & \textbf{Avg} \\
\midrule
\multirow{3}{*}{SP} & last & .126 & .104 & \textbf{.174} & .515 & .230 \\
 & mean & \textbf{.160} & \textbf{.128} & .161 & \textbf{.540} & \textbf{.247} \\
 & max & -.041 & -.026 & .029 & .065 & .007 \\
\cmidrule(lr){1-7}
\multirow{3}{*}{Perplexity} & last & \textbf{-.154} & \textbf{-.111} & \textbf{-.166} & \textbf{.900} & \textbf{.117} \\
 & mean & -.157 & -.199 & -.223 & .730 & .038 \\
 & max & -.162 & -.281 & -.249 & .710 & .005 \\
\cmidrule(lr){1-7}
\multirow{3}{*}{MTE} & last & -.170 & \textbf{-.106} & \textbf{-.150} & \textbf{.915} & \textbf{.122} \\
 & mean & \textbf{-.162} & -.174 & -.173 & .725 & .054 \\
 & max & -.167 & -.257 & -.242 & .715 & .012 \\
\cmidrule(lr){1-7}
\multirow{3}{*}{Self-certainty} & last & .280 & .169 & \textbf{.134} & \textbf{.900} & \textbf{.371} \\
 & mean & \textbf{.284} & \textbf{.218} & .121 & .695 & .330 \\
 & max & .072 & -.019 & .021 & .245 & .080 \\
\cmidrule(lr){1-7}
\multirow{3}{*}{Verb.} & last & \textbf{.388} & .264 & \textbf{.469} & .715 & .459 \\
 & mean & .306 & \textbf{.289} & .410 & .905 & \textbf{.478} \\
 & max & .298 & .227 & .400 & \textbf{.930} & .464 \\
\bottomrule
\end{tabular}
\end{table}

\begin{table}[h]
\caption{\textbf{Aggregation of raw UQ scores: coding tasks.} In-domain PRR@0.5$\uparrow$ of raw uncertainty scores aggregated over the trajectory by the last step, the mean, or the maximum over steps, averaged over the model and harness configurations of each benchmark. Avg is the mean over the benchmarks. The best aggregation for each score and column is \textbf{bold}.}
\label{tab:raw-aggregation-coding}
\centering
\small
\setlength{\tabcolsep}{4pt}
\begin{tabular}{ll|ccccc|c}
\toprule
\textbf{Score} & \textbf{Agg.} & \textbf{LCB} & \textbf{BCB} & \textbf{SWE} & \textbf{CC} & \textbf{DeepSWE} & \textbf{Avg} \\
\midrule
\multirow{3}{*}{SP} & last & .250 & .230 & \textbf{.370} & .460 & -.063 & .249 \\
 & mean & \textbf{.280} & .250 & .360 & .550 & \textbf{.017} & \textbf{.291} \\
 & max & .250 & \textbf{.280} & .310 & \textbf{.560} & -.030 & .274 \\
\cmidrule(lr){1-8}
\multirow{3}{*}{Perplexity} & last & .030 & .160 & .200 & .180 & -.063 & .101 \\
 & mean & \textbf{.050} & .190 & \textbf{.220} & \textbf{.200} & \textbf{.007} & \textbf{.133} \\
 & max & .020 & \textbf{.200} & .210 & .090 & -.070 & .090 \\
\cmidrule(lr){1-8}
\multirow{3}{*}{MTE} & last & .040 & .140 & .210 & \textbf{.210} & \textbf{.030} & .126 \\
 & mean & \textbf{.060} & .190 & .220 & \textbf{.210} & .023 & \textbf{.141} \\
 & max & .040 & \textbf{.200} & \textbf{.230} & .120 & -.027 & .113 \\
\cmidrule(lr){1-8}
\multirow{3}{*}{Self-certainty} & last & .140 & .160 & .260 & \textbf{.390} & \textbf{.023} & \textbf{.195} \\
 & mean & \textbf{.170} & \textbf{.170} & \textbf{.270} & \textbf{.390} & -.030 & .194 \\
 & max & \textbf{.170} & .140 & .230 & .360 & .010 & .182 \\
\cmidrule(lr){1-8}
\multirow{3}{*}{Verb.} & last & \textbf{.120} & \textbf{.180} & \textbf{.320} & .340 & \textbf{.030} & .198 \\
 & mean & .110 & .160 & \textbf{.320} & \textbf{.410} & -.007 & \textbf{.199} \\
 & max & .100 & .160 & \textbf{.320} & .340 & -.043 & .175 \\
\bottomrule
\end{tabular}
\end{table}

\begin{table}[h]
\caption{\textbf{Aggregation of raw UQ scores: interactive tasks.} In-domain PRR@0.5$\uparrow$ of raw uncertainty scores aggregated over the trajectory by the last step, the mean, or the maximum over steps, averaged over the model and harness configurations of each benchmark except Qwen2.5-32B-Instruct. Avg is the mean over the benchmarks. The best aggregation for each score and column is \textbf{bold}.}
\label{tab:raw-aggregation-interactive}
\centering
\small
\setlength{\tabcolsep}{4pt}
\begin{tabular}{ll|cccccc|c}
\toprule
\textbf{Score} & \textbf{Agg.} & \textbf{$\tau^3$-retail} & \textbf{$\tau^3$-airline} & \textbf{$\tau^3$-telecom} & \textbf{WebArena} & \textbf{OSWorld} & \textbf{ALFWorld} & \textbf{Avg} \\
\midrule
\multirow{3}{*}{SP} & last & .126 & \textbf{.153} & .012 & \textbf{.435} & .220 & \textbf{.323} & \textbf{.211} \\
 & mean & \textbf{.129} & -.089 & .009 & .375 & \textbf{.400} & .320 & .191 \\
 & max & .019 & .048 & \textbf{.043} & .030 & .080 & -.085 & .022 \\
\cmidrule(lr){1-9}
\multirow{3}{*}{Perplexity} & last & \textbf{-.090} & -.220 & \textbf{.005} & \textbf{.470} & .040 & .250 & \textbf{.076} \\
 & mean & -.179 & \textbf{-.012} & -.357 & .400 & \textbf{.180} & .215 & .041 \\
 & max & -.181 & -.098 & -.288 & .310 & \textbf{.180} & \textbf{.353} & .046 \\
\cmidrule(lr){1-9}
\multirow{3}{*}{MTE} & last & \textbf{-.083} & -.218 & \textbf{-.017} & \textbf{.480} & .060 & .310 & \textbf{.089} \\
 & mean & -.142 & \textbf{-.021} & -.321 & .385 & .170 & .200 & .045 \\
 & max & -.186 & -.121 & -.297 & .305 & \textbf{.180} & \textbf{.323} & .034 \\
\cmidrule(lr){1-9}
\multirow{3}{*}{Self-certainty} & last & .114 & \textbf{.087} & .066 & \textbf{.545} & .280 & \textbf{.340} & .239 \\
 & mean & \textbf{.186} & -.025 & \textbf{.387} & .390 & \textbf{.320} & .275 & \textbf{.256} \\
 & max & .041 & -.066 & .088 & .320 & .150 & .002 & .089 \\
\cmidrule(lr){1-9}
\multirow{3}{*}{Verb.} & last & .146 & \textbf{.027} & .100 & \textbf{.285} & \textbf{.280} & .187 & \textbf{.171} \\
 & mean & \textbf{.192} & -.026 & \textbf{.211} & .220 & .200 & \textbf{.215} & .169 \\
 & max & .185 & -.090 & .153 & .200 & -.010 & .158 & .099 \\
\bottomrule
\end{tabular}
\end{table}

\ecoparagraph{Components of Bayes Fused.}
\Cref{tab:components-retrieval,tab:components-coding,tab:components-interactive} present the contribution of each evidence source of \Cref{eq:fused-cont}, using SP as the generation signal. The relative importance of the two sources depends on the domain. On retrieval, tool feedback is much stronger than generation uncertainty (PRR of 0.306 vs. 0.089), while on coding the ordering reverses (0.198 vs. 0.309). On interactive tasks, tool feedback is slightly stronger (0.301 vs. 0.254). Importantly, in all domains, the best combination of the two sources outperforms each of them alone.

We also compare four aggregations of the generation evidence. The results largely support the choices made in the main tables: history\,$+$\,last performs best on interactive tasks (0.384 vs. 0.352 for SEP), and continuous aggregation performs best on coding (0.337 vs. 0.321 for history\,$+$\,last). On retrieval, history\,$+$\,last (0.382) is close to the best variant, SEP (0.393), and substantially outperforms continuous aggregation (0.179). Continuous aggregation performs even worse than tool feedback alone, since summing one log-likelihood ratio per step lets long trajectories accumulate evidence from uninformative steps.

Note that continuous aggregation and history\,$+$\,last differ not only in separating the final step. Continuous aggregation adds one contribution per step, so its magnitude grows with trajectory length, and fits class-specific variances, which makes each contribution quadratic in the signal. In contrast, history\,$+$\,last adds exactly two contributions regardless of trajectory length and pools the variance within each temporal role, which makes each contribution linear. Thus, the comparison reflects all these design choices rather than the temporal split alone.

\begin{table}[t]
    \caption{\textbf{Components of Bayes Fused: retrieval tasks.} In-domain PRR@0.5$\uparrow$ with SP, averaged over the model and harness configurations of each benchmark. The two evidence sources are separated: $\ell_C$ carries the prior and the tool checks, $\Lambda_U$ carries generation uncertainty, and Bayes Fused is their sum. The last four rows differ only in how the generation evidence is aggregated over the trajectory: the final step only ($\Lambda_L$), the history mean plus the final step ($\Lambda_H+\Lambda_L$), a binarized per-step scores (SEP), or all steps ($\Lambda_U$). The best score in each column is \textbf{bold}. Avg is the mean over the benchmarks. HotpotQA, MuSiQue, and CRAG are averaged over the GPT-OSS-20B and Qwen3.6-35B-A3B configurations (without Qwen2.5-32B-Instruct).}
    \label{tab:components-retrieval}
    \centering
    \resizebox{\linewidth}{!}{%
    \begin{tabular}{ll|cccc|c}
    \toprule
    \textbf{Score} & & \textbf{HotpotQA} & \textbf{MuSiQue} & \textbf{CRAG} & \textbf{BrowseComp-Plus} & \textbf{Avg} \\
    \midrule
    Generation evidence only & $\widehat{\ell}_0+\Lambda_U$ & .015 & -.050 & \textbf{.148} & .245 & .089 \\
    Tool evidence only & $\ell_C$ & .388 & .413 & -.021 & .445 & .306 \\
    Bayes Fused, last only & $\ell_C+\Lambda_L$ & .392 & .357 & .089 & .495 & .333 \\
    Bayes Fused, history $+$ last & $\ell_C+\Lambda_H+\Lambda_L$ & .410 & \textbf{.417} & .117 & .585 & .382 \\
    Bayes Fused, SEP & $\ell_C+\Lambda_U^{\mathrm{SEP}}$ & \textbf{.427} & .399 & .134 & \textbf{.610} & \textbf{.393} \\
    Bayes Fused, continuous & $\ell_C+\Lambda_U$ & .217 & .146 & .110 & .245 & .179 \\
    \bottomrule
    \end{tabular}
    }
\end{table}

\begin{table}[t]
    \caption{\textbf{Components of Bayes Fused: coding tasks.} In-domain PRR@0.5$\uparrow$ with SP, averaged over the model and harness configurations of each benchmark. The two evidence sources are separated: $\ell_C$ carries the prior and the tool checks, $\Lambda_U$ carries generation uncertainty, and Bayes Fused is their sum. The last four rows differ only in how the generation evidence is aggregated over the trajectory: the final step only ($\Lambda_L$), the history mean plus the final step ($\Lambda_H+\Lambda_L$), a binarized per-step scores (SEP), or all steps ($\Lambda_U$). The best score in each column is \textbf{bold}. Avg is the mean over the benchmarks.}
    \label{tab:components-coding}
    \centering
    \resizebox{\linewidth}{!}{%
    \begin{tabular}{ll|ccccc|c}
    \toprule
    \textbf{Score} & & \textbf{LCB} & \textbf{BCB} & \textbf{SWE} & \textbf{CC} & \textbf{DeepSWE} & \textbf{Avg} \\
    \midrule
    Generation evidence only & $\widehat{\ell}_0+\Lambda_U$ & .280 & \textbf{.260} & .360 & \textbf{.570} & .077 & .309 \\
    Tool evidence only & $\ell_C$ & .190 & .130 & .170 & .380 & .120 & .198 \\
    Bayes Fused, last only & $\ell_C+\Lambda_L$ & .260 & .200 & .340 & .410 & .117 & .265 \\
    Bayes Fused, history $+$ last & $\ell_C+\Lambda_H+\Lambda_L$ & \textbf{.300} & \textbf{.260} & \textbf{.370} & .530 & .147 & .321 \\
    Bayes Fused, SEP & $\ell_C+\Lambda_U^{\mathrm{SEP}}$ & .290 & .250 & .330 & .500 & \textbf{.227} & .319 \\
    Bayes Fused, continuous & $\ell_C+\Lambda_U$ & \textbf{.300} & \textbf{.260} & .360 & .560 & .207 & \textbf{.337} \\
    \bottomrule
    \end{tabular}
    }
\end{table}

\begin{table}[t]
    \caption{\textbf{Components of Bayes Fused: interactive tasks.} In-domain PRR@0.5$\uparrow$ with SP, averaged over the model and harness configurations of each benchmark. The two evidence sources are separated: $\ell_C$ carries the prior and the tool checks, $\Lambda_U$ carries generation uncertainty, and Bayes Fused is their sum. The last four rows differ only in how the generation evidence is aggregated over the trajectory: the final step only ($\Lambda_L$), the history mean plus the final step ($\Lambda_H+\Lambda_L$), a binarized per-step scores (SEP), or all steps ($\Lambda_U$). The best score in each column is \textbf{bold}. Avg is the mean over the benchmarks. The $\tau^3$-bench columns are averaged over the GPT-OSS-20B and Qwen3.6-35B-A3B configurations (without Qwen2.5-32B-Instruct).}
    \label{tab:components-interactive}
    \centering
    \resizebox{\linewidth}{!}{%
    \begin{tabular}{ll|cccccc|c}
    \toprule
    \textbf{Score} & & \textbf{$\tau^3$-retail} & \textbf{$\tau^3$-airline} & \textbf{$\tau^3$-telecom} & \textbf{WebArena} & \textbf{OSWorld} & \textbf{ALFWorld} & \textbf{Avg} \\
    \midrule
    Generation evidence only & $\widehat{\ell}_0+\Lambda_U$ & .278 & .090 & .184 & .285 & \textbf{.390} & .295 & .254 \\
    Tool evidence only & $\ell_C$ & .105 & -.037 & .535 & .345 & .210 & \textbf{.648} & .301 \\
    Bayes Fused, last only & $\ell_C+\Lambda_L$ & .193 & -.069 & .534 & .500 & .200 & .585 & .324 \\
    Bayes Fused, history $+$ last & $\ell_C+\Lambda_H+\Lambda_L$ & .308 & .036 & .565 & \textbf{.505} & .360 & .528 & \textbf{.384} \\
    Bayes Fused, SEP & $\ell_C+\Lambda_U^{\mathrm{SEP}}$ & .229 & .074 & \textbf{.576} & .410 & .350 & .475 & .352 \\
    Bayes Fused, continuous & $\ell_C+\Lambda_U$ & \textbf{.322} & \textbf{.107} & .448 & .390 & .380 & .378 & .337 \\
    \bottomrule
    \end{tabular}
    }
\end{table}

\ecoparagraph{Components of Structured Bayesian Fusion.}
\Cref{tab:sbf-components} removes one component of \Cref{eq:structured-bayesian-fusion} at a time, leaving the rest of the estimator and the fitting procedure unchanged.

Removing any component reduces the average PRR@0.5. The largest drop is observed when removing the final-step model ($-.153$ on average), which is expected since the final step contains the agent's solution. Removing the history normalization also lowers the average score (\Cref{tab:sbf-normalization}): it prevents evidence from accumulating excessively across correlated steps. Among the remaining components, the generation history contributes the most ($-.018$ on average), followed by the critic history ($-.006$). The length term contributes least on average ($.488$ vs.\ $.486$), but it helps on CRAG ($.391$ vs.\ $.366$), where trajectories are long.

The last row separates the final-step evidence: it keeps the same final-step projection $r_N$ and critic indicator $z_N$, but replaces the joint model of \Cref{eq:tb-final} with two separate log-likelihood ratios, a Gaussian one for $r_N$ and a Bernoulli one for $z_N$, leaving all other components unchanged and refitting on the same splits. Modeling the same final-step signals separately performs on par with the joint model on every benchmark, so the gain comes from the final-step evidence itself rather than from joint modeling. The reason is structural: the joint model differs from the separate one only through the within-class correlation between the two signals, and the two coincide when this correlation is zero. On WebArena and OSWorld, the final step is the agent's answer rather than a tool call, so the indicator is constant and the two forms are equal; where the indicator varies, as on ALFWorld, they differ by at most one standard deviation. The joint form adds one parameter and is designed for settings where the two final-step signals are correlated, a concern similar to the one behind history normalization.

\begin{table}[t]
    \caption{\textbf{Components of Structured Bayesian Fusion.} In-domain PRR@0.5$\uparrow$,
    averaged over the model and harness configurations of each benchmark (mean $\pm$ SD over
    three split seeds). Each row removes one contribution of \Cref{eq:structured-bayesian-fusion}
    and leaves the rest of the estimator and the fitting procedure unchanged, so every row is
    fitted and scored on the same folds as the full model, whose row reproduces the SBF rows of
    \Cref{tab:prr-dataset-means-rag,tab:prr-dataset-means-other} for each setup. The last row
    replaces the joint final-step model of \Cref{eq:tb-final} by two separate log-likelihood
    ratios for the same signals. The best score in each column is \textbf{bold}. Avg is the
    mean over the 5 columns shown.}
    \label{tab:sbf-components}
    \centering
    \resizebox{\linewidth}{!}{%
    \begin{tabular}{ll|ccccc|c}
    \toprule
    \textbf{Variant} & & \textbf{WebArena} & \textbf{OSWorld} & \textbf{ALFWorld} & \textbf{CRAG} & \textbf{$\tau^3$-retail} & \textbf{Avg} \\
    \midrule
    Full model & $\ell_{\mathrm{SBF}}$ & .545 {\scriptsize $\pm$ .007} & .400 {\scriptsize $\pm$ .010} & .825 {\scriptsize $\pm$ .010} & .391 {\scriptsize $\pm$ .015} & \textbf{.278} {\scriptsize $\pm$ .019} & .488 {\scriptsize $\pm$ .006} \\
    \midrule
    Without trajectory length (\Cref{eq:tb-duration}) & $-\,\Lambda_{\mathrm{len}}$ & \textbf{.553} {\scriptsize $\pm$ .005} & \textbf{.402} {\scriptsize $\pm$ .005} & \textbf{.836} {\scriptsize $\pm$ .006} & .366 {\scriptsize $\pm$ .014} & .273 {\scriptsize $\pm$ .021} & .486 {\scriptsize $\pm$ .005} \\
    Without critic history (\Cref{eq:tb-critic}) & $-\,\Lambda_{H,C}$ & .542 {\scriptsize $\pm$ .003} & .388 {\scriptsize $\pm$ .008} & .835 {\scriptsize $\pm$ .011} & .388 {\scriptsize $\pm$ .015} & .259 {\scriptsize $\pm$ .018} & .482 {\scriptsize $\pm$ .005} \\
    Without generation history (\Cref{eq:tb-history}) & $-\,\Lambda_{H,U}$ & .542 {\scriptsize $\pm$ .003} & .400 {\scriptsize $\pm$ .018} & .820 {\scriptsize $\pm$ .007} & .368 {\scriptsize $\pm$ .014} & .221 {\scriptsize $\pm$ .012} & .470 {\scriptsize $\pm$ .005} \\
    Without the final step (\Cref{eq:tb-final}) & $-\,\Lambda_{\mathrm{last}}$ & .370 {\scriptsize $\pm$ .006} & .382 {\scriptsize $\pm$ .001} & .349 {\scriptsize $\pm$ .022} & .318 {\scriptsize $\pm$ .009} & .257 {\scriptsize $\pm$ .020} & .335 {\scriptsize $\pm$ .006} \\
    Separate final-step evidence & sep.\ $\Lambda_{\mathrm{last}}$ & .545 {\scriptsize $\pm$ .007} & .400 {\scriptsize $\pm$ .010} & .834 {\scriptsize $\pm$ .011} & \textbf{.398} {\scriptsize $\pm$ .014} & \textbf{.278} {\scriptsize $\pm$ .019} & \textbf{.491} {\scriptsize $\pm$ .006} \\
    \bottomrule
    \end{tabular}
    }
\end{table}

\ecoparagraph{History normalization.}
\Cref{tab:sbf-normalization} replaces the normalized history contributions of \Cref{eq:tb-normalization} by the raw accumulated scores $\Lambda_{H,C}^{\mathrm{raw}}$ and $\Lambda_{H,U}^{\mathrm{raw}}$. Normalization improves the average PRR from 0.412 to 0.488, with the largest gain on ALFWorld (0.647 to 0.825) and CRAG (0.292 to 0.391), while on OSWorld the raw scores rank slightly better. The decomposition of the score explains why the effect on ranking can have either sign. Writing $\ell$ for the SBF score and $\ell^{\mathrm{raw}}$ for the variant without normalization,
\[
\ell-\ell^{\mathrm{raw}}
=(\widehat{\beta}_{H,C}-1)\Lambda_{H,C}^{\mathrm{raw}}
+(\widehat{\beta}_{H,U}-1)\Lambda_{H,U}^{\mathrm{raw}}
-\widehat{\beta}_{H,C}\widehat{a}_{H,C}
-\widehat{\beta}_{H,U}\widehat{a}_{H,U}.
\]
The last two terms are the same for every trajectory with a non-empty history, so the centering parameters cannot change any ranking at all. Only the two slopes can, and only by changing how heavily each history channel weighs against $\Lambda_{\mathrm{len}}$ and $\Lambda_{\mathrm{last}}$, a reweighting that can help or hurt a ranking depending on whether the contributions agree about the order. Both slopes and offsets, by contrast, move every predicted probability.

\Cref{prop:effective-history} accounts for the direction of the change on the probability scale: independent accumulation inflates the history evidence by $1+(K-1)\rho$, so the raw score is too confident, and the inflation grows with the history length $K$. The raw score degrades most on ALFWorld and CRAG (\Cref{tab:sbf-normalization}).
In an additional analysis over $35$ model--harness--benchmark configurations, the absolute difference in PRR between SBF with and without normalization grows with history length: it correlates at $0.46$ with the mean number of history steps $K$ and is more than three times larger for the third of configurations with the longest histories than for the third with the shortest.

SBF normalizes the accumulated history evidence with a fitted slope and offset (\Cref{eq:tb-normalization}) rather than with the length-specific factor of \Cref{prop:effective-history}. For comparison, we also tested this length correction directly on ALFWorld and DeepSWE: we estimate $\rho$ on the training folds and divide each accumulated history term by $1+(K-1)\rho$ (\Cref{tab:length-correction}). The length correction improves both ranking and calibration, as \Cref{prop:effective-history} suggests. Finally, SBF's normalization reaches the same quality without estimating $\rho$, so it already provides a similar correction.

\begin{table}[t]
    \caption{\textbf{History normalization in Structured Bayesian Fusion.} In-domain PRR@0.5$\uparrow$ of SBF with and without the normalization of \Cref{eq:tb-normalization}; the latter uses the raw accumulated history scores $\Lambda_{H,C}^{\mathrm{raw}}$ and $\Lambda_{H,U}^{\mathrm{raw}}$. Same configurations and protocol as \Cref{tab:sbf-components}. The best result in each column is \textbf{bold}.}
    \label{tab:sbf-normalization}
    \centering
    \resizebox{\linewidth}{!}{%
    \begin{tabular}{l|ccccc|c}
    \toprule
    \textbf{Method} & \textbf{WebArena} & \textbf{OSWorld} & \textbf{ALFWorld} & \textbf{CRAG} & \textbf{$\tau^3$-retail} & \textbf{Avg} \\
    \midrule
    SBF & \textbf{.545} {\scriptsize $\pm$ .007} & .400 {\scriptsize $\pm$ .010} & \textbf{.825} {\scriptsize $\pm$ .010} & \textbf{.391} {\scriptsize $\pm$ .015} & \textbf{.278} {\scriptsize $\pm$ .019} & \textbf{.488} {\scriptsize $\pm$ .006} \\
    SBF without normalization & .443 {\scriptsize $\pm$ .010} & \textbf{.408} {\scriptsize $\pm$ .007} & .647 {\scriptsize $\pm$ .041} & .292 {\scriptsize $\pm$ .011} & .271 {\scriptsize $\pm$ .021} & .412 {\scriptsize $\pm$ .010} \\
    \bottomrule
    \end{tabular}
    }
\end{table}
\begin{table}[t]
    \caption{\textbf{Length correction from \Cref{prop:effective-history}.} In-domain PRR@0.5$\uparrow$ and ECE$\downarrow$ of SBF with three history terms: independent accumulation (\Cref{eq:tb-history}); the length correction, which estimates $\rho$ on the training folds and divides each accumulated history term by $1+(K-1)\rho$; and the affine normalization of SBF (\Cref{eq:tb-normalization}), as in \Cref{tab:prr-dataset-means-coding,tab:prr-dataset-means-other}. The best result in each column is \textbf{bold}.}
    \label{tab:length-correction}
    \centering
    \resizebox{\linewidth}{!}{%
    \begin{tabular}{l|ccc|ccc}
    \toprule
    & \multicolumn{3}{c|}{\textbf{PRR@0.5}$\uparrow$} & \multicolumn{3}{c}{\textbf{ECE}$\downarrow$} \\
    \textbf{History term} & \textbf{ALFWorld} & \textbf{DeepSWE} & \textbf{Avg} & \textbf{ALFWorld} & \textbf{DeepSWE} & \textbf{Avg} \\
    \midrule
    Independent accumulation (\Cref{eq:tb-history}) & .756 {\scriptsize $\pm$ .023} & .157 {\scriptsize $\pm$ .039} & .457 {\scriptsize $\pm$ .023} & .178 {\scriptsize $\pm$ .006} & .202 {\scriptsize $\pm$ .015} & .190 {\scriptsize $\pm$ .008} \\
    + length correction of \Cref{prop:effective-history} & .804 {\scriptsize $\pm$ .020} & \textbf{.177} {\scriptsize $\pm$ .029} & .490 {\scriptsize $\pm$ .018} & .109 {\scriptsize $\pm$ .008} & \textbf{.123} {\scriptsize $\pm$ .009} & .116 {\scriptsize $\pm$ .006} \\
    + affine normalization (SBF) & \textbf{.825} {\scriptsize $\pm$ .010} & .173 {\scriptsize $\pm$ .036} & \textbf{.499} {\scriptsize $\pm$ .019} & \textbf{.097} {\scriptsize $\pm$ .005} & .129 {\scriptsize $\pm$ .020} & \textbf{.113} {\scriptsize $\pm$ .010} \\
    \bottomrule
    \end{tabular}
    }
\end{table}

\ecoparagraph{Training and inference cost.}
\Cref{tab:runtime-cost} provides the training and inference cost of each estimator. Training is measured on a single CPU core over nine agent runs on ALFWorld, DeepSWE, and WebArena-Verified, and inference on WebArena-Verified trajectories with GPT-OSS~\citep{openai2025gptossb}, where one agent generation takes 614\,ms. Bayes Fused and Structured Bayesian Fusion take less than 0.1\,s to train, offline, and scoring a step adds less than 0.001\% to its LLM inference time, so their cost is negligible compared to the agent run. The cost of an estimator is therefore dominated by the signals it requires. Token-level UQ scores and tool feedback are produced during the agent run at no extra cost. Verbalized confidence requires one additional short LLM call per step, adding 317\,ms, or 52\% overhead relative to a generation. UProp~\citep{duan2025uprop} requires ten additional generations per step, adding 1.8\,s, or 300\% overhead relative to a generation, making it the most expensive method.

\begin{table}[t]
\centering
\caption{\textbf{Cost of each estimator.} \textit{Training} is the one-time cost of fitting the estimator offline on a single CPU core, median over nine agent runs on ALFWorld, DeepSWE, and WebArena-Verified. \textit{Inference} shows the extra cost of scoring one agent step: the number of additional LLM calls, the added time including these calls, and this time as a percentage of one agent generation. Inference time is the median across WebArena-Verified trajectories with GPT-OSS, with a median of 614\,ms per agent generation.}
\label{tab:runtime-cost}
\small
\setlength{\tabcolsep}{4pt}
\resizebox{\linewidth}{!}{%
\begin{tabular}{ll|r|rrr}
\toprule
\multirow{2}{*}{\textbf{Group}} & \multirow{2}{*}{\textbf{Method}} & \multicolumn{1}{c|}{\textbf{Training}} & \multicolumn{3}{c}{\textbf{Inference}} \\
 & & \textbf{Time (ms)} & \textbf{LLM calls} & \textbf{Time (ms)} & \textbf{Overhead (\%)} \\
\midrule
\multirow{2}{*}{\rotatebox[origin=c]{90}{\scriptsize Stat.}} & $N$ steps & \textendash & 0 & $<$0.001 & $<$0.001 \\
 & Tool success rate & \textendash & 0 & $<$0.001 & $<$0.001 \\
\cmidrule(lr){2-6}
\multirow{4}{*}{\rotatebox[origin=c]{90}{\scriptsize UQ}} & SP, Perplexity, MTE, Self-certainty & \textendash & 0 & $<$0.001 & $<$0.001 \\
 & Verb. & \textendash & 1 & 317 & 52 \\
 & SAUP & 2{,}595 & 0 & 0.009 & 0.001 \\
 & UProp & \textendash & 10 & 1843 & 300 \\
\cmidrule(lr){2-6}
\multirow{2}{*}{\rotatebox[origin=c]{90}{\scriptsize \shortstack{Super-\\vised}}} & HTC & 760 & 0 & 0.033 & 0.005 \\
 & Logistic regression & 399 & 0 & 0.025 & 0.004 \\
\cmidrule(lr){2-6}
\multirow{4}{*}{\rotatebox[origin=c]{90}{\scriptsize \textbf{Bayes (ours)}}} & Tool-only & 0.05 & 0 & $<$0.001 & $<$0.001 \\
 & Fused (SP, Perplexity, MTE, Self-cert.) & 1.8 & 0 & $<$0.001 & $<$0.001 \\
 & Fused (Verb.) & 0.9 & 1 & 317 & 52 \\
 & SBF & 23 & 0 & 0.005 & $<$0.001 \\
\bottomrule
\end{tabular}}
\end{table}

\clearpage
\section{Additional Results}
\label{app:results}

\ecoparagraph{Effect of reasoning effort.}
\Cref{tab:reasoning-effort} re-evaluates the strongest methods of \Cref{tab:prr-dataset-means-coding} at different reasoning efforts the models expose (low, medium, and high for GPT-OSS-120B~\citep{openai2025gptossb}; none, low, and medium for Qwen3.8-27B~\citep{qwen38}), averaged over LiveCodeBench-Hard and SWE-bench and over both agent harnesses. Structured Bayesian Fusion performs best at every effort level except for GPT-OSS-120B with medium thinking, and its margin over raw SP is largest in the no-thinking setup.

\begin{table}[h]
    \caption{\textbf{Reasoning effort.} In-domain PRR@0.5$\uparrow$, averaged over LiveCodeBench-Hard and SWE-Bench and two harnesses: smolagents CodeAgent and ReAct. SP denotes raw sum-logprob averaged over the trajectory. Fused (SP), Structured Bayesian Fusion (SBF), and logistic regression are the methods reported in \Cref{tab:prr-dataset-means-coding}. GPT-OSS-120B is evaluated at low, medium, and high effort, and Qwen3.8-27B at none, low, and medium. $\Delta$ is the PRR of the best fitted method minus that of SP.}
    \label{tab:reasoning-effort}
    \centering
    \resizebox{\linewidth}{!}{%
    \begin{tabular}{ll|c|ccc|c}
    \toprule
    \textbf{Model} & \textbf{Effort} & \textbf{SP} & \textbf{Fused (SP)} & \textbf{SBF} & \textbf{Logistic regression} & \textbf{$\Delta$} \\
    \midrule
    \multirow{3}{*}{GPT-OSS-120B} & low & .340 {\scriptsize $\pm$ .000} & .355 {\scriptsize $\pm$ .004} & .405 {\scriptsize $\pm$ .013} & .382 {\scriptsize $\pm$ .006} & +.065 \\
     & medium & .294 {\scriptsize $\pm$ .000} & .327 {\scriptsize $\pm$ .003} & .316 {\scriptsize $\pm$ .015} & .303 {\scriptsize $\pm$ .010} & +.033 \\
     & high & .172 {\scriptsize $\pm$ .000} & .142 {\scriptsize $\pm$ .006} & .229 {\scriptsize $\pm$ .013} & .193 {\scriptsize $\pm$ .014} & +.057 \\
    \midrule
    \multirow{3}{*}{Qwen3.8-27B} & none & .280 {\scriptsize $\pm$ .000} & .319 {\scriptsize $\pm$ .006} & .379 {\scriptsize $\pm$ .008} & .337 {\scriptsize $\pm$ .013} & +.099 \\
     & low & .274 {\scriptsize $\pm$ .000} & .254 {\scriptsize $\pm$ .009} & .297 {\scriptsize $\pm$ .013} & .245 {\scriptsize $\pm$ .010} & +.023 \\
     & medium & .261 {\scriptsize $\pm$ .000} & .255 {\scriptsize $\pm$ .004} & .306 {\scriptsize $\pm$ .010} & .264 {\scriptsize $\pm$ .012} & +.045 \\
    \bottomrule
    \end{tabular}
    }
\end{table}

\ecoparagraph{Calibration evaluation.}
We additionally report the expected calibration error (ECE; \citep{pmlr-v70-guo17a}) of the held-out predictions (\Cref{tab:ece}), averaged across seeds and configurations. Raw SP scores are min-max normalized to $[0, 1]$. SBF + Platt applies Platt scaling~\citep{platt2000probabilistic} fitted on out-of-fold SBF predictions within the training folds only. For DeepSWE, ECE is computed against the cohort-median binary labels.

A discriminative model trained with a log-loss objective, such as logistic regression, is expected to be better calibrated. Nevertheless, SBF is close to logistic regression on three of the five benchmarks (WebArena, ALFWorld, and BrowseComp-Plus), and both learned estimators are far better calibrated than raw SP. If absolute probabilities are needed, a monotone recalibration fitted on the training folds (e.g., Platt scaling) can be applied while preserving the ranking within each fold, so PRR is essentially unaffected. SBF + Platt achieves the lowest average ECE and outperforms logistic regression on three benchmarks, with ranking unchanged within each fold.

\begin{table}[h]
    \caption{\textbf{Calibration of predicted probabilities.} ECE$\downarrow$ with 10 equal-width bins of the held-out predictions, averaged across split seeds and configurations. Raw SP scores are min-max normalized to $[0, 1]$. SBF + Platt applies Platt scaling fitted on out-of-fold SBF predictions within the training folds only. The best result in each column is \textbf{bold}.}
    \label{tab:ece}
    \centering
    \resizebox{\linewidth}{!}{%
    \begin{tabular}{l|ccccc|c}
    \toprule
    \textbf{Method} & \textbf{WebArena} & \textbf{OSWorld} & \textbf{ALFWorld} & \textbf{BrowseComp} & \textbf{DeepSWE} & \textbf{Avg} \\
    \midrule
    SP & .467 {\scriptsize $\pm$ .000} & .059 {\scriptsize $\pm$ .000} & .262 {\scriptsize $\pm$ .000} & .507 {\scriptsize $\pm$ .000} & .219 {\scriptsize $\pm$ .000} & .303 {\scriptsize $\pm$ .000} \\
    Logistic regression & .054 {\scriptsize $\pm$ .006} & \textbf{.058} {\scriptsize $\pm$ .009} & .094 {\scriptsize $\pm$ .016} & .065 {\scriptsize $\pm$ .002} & \textbf{.075} {\scriptsize $\pm$ .017} & .069 {\scriptsize $\pm$ .005} \\
    Bayes Fused (SP) & .099 {\scriptsize $\pm$ .002} & .064 {\scriptsize $\pm$ .013} & .080 {\scriptsize $\pm$ .009} & .077 {\scriptsize $\pm$ .005} & .237 {\scriptsize $\pm$ .012} & .111 {\scriptsize $\pm$ .004} \\
    SBF & .062 {\scriptsize $\pm$ .005} & .112 {\scriptsize $\pm$ .006} & .097 {\scriptsize $\pm$ .005} & .060 {\scriptsize $\pm$ .002} & .129 {\scriptsize $\pm$ .020} & .092 {\scriptsize $\pm$ .004} \\
    SBF + Platt & \textbf{.042} {\scriptsize $\pm$ .005} & .072 {\scriptsize $\pm$ .020} & \textbf{.064} {\scriptsize $\pm$ .004} & \textbf{.022} {\scriptsize $\pm$ .003} & .079 {\scriptsize $\pm$ .016} & \textbf{.056} {\scriptsize $\pm$ .005} \\
    \bottomrule
    \end{tabular}
    }
\end{table}

\ecoparagraph{Failure prediction at intermediate steps.} Each estimator sees only the trajectory up to step $k$, and we test two setups: (1) whether it predicts the final outcome of the episode (\Cref{tab:prefix-final}), and (2) whether it predicts the state of the task at step $k$ (\Cref{tab:prefix-state}). To obtain state labels on BrowseComp-Plus, we stop the agent after 10 steps and force it to give a final answer. On DeepSWE, we restore the repository after a random sample of the commands that modified the code (up to 8 per episode) and re-run the hidden tests.

SBF has the best average in both setups, with the largest margin for the intermediate state (.625 vs. .559 for the second-best estimator), and the highest PRR on DeepSWE in both. Prediction is harder very early in a trajectory: on WebArena, all methods stay below .15 after 2 to 4 steps.

\begin{table}[h]
    \caption{\textbf{Predicting the final outcome from the first $k$ steps.} In-domain PRR@0.5$\uparrow$ of estimators that see only the trajectory up to step $k$, scored against the final outcome of the episode. On DeepSWE, $k$ is a mid-trajectory step after a command that modified the code. The best result in each column is \textbf{bold}.}
    \label{tab:prefix-final}
    \centering
    \resizebox{0.8\linewidth}{!}{%
    \begin{tabular}{l|ccc|c}
    \toprule
    \textbf{Method} & \textbf{WebArena, 8 steps} & \textbf{BrowseComp-Plus, 10 steps} & \textbf{DeepSWE, mid-trajectory step} & \textbf{Avg} \\
    \midrule
    SP & .171 {\scriptsize $\pm$ .000} & \textbf{.471} {\scriptsize $\pm$ .000} & $-$.135 {\scriptsize $\pm$ .000} & .169 {\scriptsize $\pm$ .000} \\
    Logistic regression & \textbf{.307} {\scriptsize $\pm$ .016} & .431 {\scriptsize $\pm$ .009} & .035 {\scriptsize $\pm$ .027} & .258 {\scriptsize $\pm$ .011} \\
    Bayes Fused (SP) & .196 {\scriptsize $\pm$ .022} & .446 {\scriptsize $\pm$ .001} & .025 {\scriptsize $\pm$ .091} & .222 {\scriptsize $\pm$ .031} \\
    SBF & .294 {\scriptsize $\pm$ .014} & .453 {\scriptsize $\pm$ .006} & \textbf{.095} {\scriptsize $\pm$ .061} & \textbf{.281} {\scriptsize $\pm$ .021} \\
    \bottomrule
    \end{tabular}
    }
\end{table}

\begin{table}[h]
    \caption{\textbf{Predicting the state of the task at step $k$.} In-domain PRR@0.5$\uparrow$ of estimators that see only the trajectory up to step $k$, scored against the state of the task at that step. On BrowseComp-Plus, the agent is stopped after 10 steps and forced to give a final answer. On DeepSWE, the repository is restored after a random sample of the commands that modified the code (up to 8 per episode) and the hidden tests are re-run. The best result in each column is \textbf{bold}.}
    \label{tab:prefix-state}
    \centering
    \resizebox{0.65\linewidth}{!}{%
    \begin{tabular}{l|cc|c}
    \toprule
    \textbf{Method} & \textbf{BrowseComp-Plus, 10 steps} & \textbf{DeepSWE, mid-trajectory step} & \textbf{Avg} \\
    \midrule
    SP & .902 {\scriptsize $\pm$ .000} & $-$.040 {\scriptsize $\pm$ .000} & .431 {\scriptsize $\pm$ .000} \\
    Logistic regression & .934 {\scriptsize $\pm$ .003} & .095 {\scriptsize $\pm$ .091} & .515 {\scriptsize $\pm$ .046} \\
    Bayes Fused (SP) & .908 {\scriptsize $\pm$ .001} & .210 {\scriptsize $\pm$ .034} & .559 {\scriptsize $\pm$ .017} \\
    SBF & \textbf{.939} {\scriptsize $\pm$ .002} & \textbf{.310} {\scriptsize $\pm$ .016} & \textbf{.625} {\scriptsize $\pm$ .008} \\
    \bottomrule
    \end{tabular}
    }
\end{table}

\ecoparagraph{LLM-as-judge verifier.} Our main experiments include the verbalized uncertainty score, which measures how well the LLM can assess the correctness of its own responses. However, for a more independent comparison, we additionally evaluate an LLM-as-judge verifier on ALFWorld and DeepSWE (\Cref{tab:llm-judge}). The judge performs worse than SBF on both benchmarks, despite requiring an additional LLM call, whereas SBF relies only on signals already produced by the agent. Its performance also varies substantially across agent configurations: on ALFWorld, it achieves .78 and .81 with the two GPT-OSS configurations, but only .17 and $-$.11 with the Qwen configurations.

\begin{table}[h]
    \caption{\textbf{LLM-as-judge verifier.} In-domain PRR@0.5$\uparrow$ at the final step. The judge (GPT-OSS-20B) receives the task, the agent's final response, and the last 10 steps of its trajectory, and predicts the probability that the agent successfully completed the task. Same protocol as the other methods. The best result in each column is \textbf{bold}.}
    \label{tab:llm-judge}
    \centering
    \resizebox{0.5\linewidth}{!}{%
    \begin{tabular}{l|cc}
    \toprule
    \textbf{Method} & \textbf{ALFWorld} & \textbf{DeepSWE} \\
    \midrule
    LLM judge & .410 {\scriptsize $\pm$ .000} & $-$.020 {\scriptsize $\pm$ .000} \\
    Bayes Fused (SP) & .528 {\scriptsize $\pm$ .015} & \textbf{.207} {\scriptsize $\pm$ .021} \\
    SBF & \textbf{.825} {\scriptsize $\pm$ .010} & .173 {\scriptsize $\pm$ .036} \\
    \bottomrule
    \end{tabular}
    }
\end{table}

\ecoparagraph{Terminated trajectories.} For completeness, we re-ran the evaluation on all episodes, counting trajectories terminated due to the step budget or harness failures as failures, while keeping the same protocol and fold stratification (\Cref{tab:terminated}). The conclusions remain unchanged: SBF is still the best-performing method on average, and Bayesian fusion consistently improves over the raw UQ signals. All scores increase because terminated trajectories are relatively easy to flag. As expected, trajectory length alone ($N$ steps) becomes a strong predictor of failure. This further motivates our focus on completed trajectories in the main evaluation, where failure detection is more challenging.

\begin{table}[h]
    \caption{\textbf{Including terminated trajectories as failures.} In-domain PRR@0.5$\uparrow$ on all episodes, counting trajectories terminated due to the step budget or harness failures as failures, with the same protocol and fold stratification as the main evaluation. The best result in each column is \textbf{bold}.}
    \label{tab:terminated}
    \centering
    \resizebox{\linewidth}{!}{%
    \begin{tabular}{l|ccccc|c}
    \toprule
    \textbf{Method} & \textbf{HotpotQA} & \textbf{MuSiQue} & \textbf{CRAG} & \textbf{WebArena} & \textbf{OSWorld} & \textbf{Avg} \\
    \midrule
    N steps & .371 {\scriptsize $\pm$ .010} & .511 {\scriptsize $\pm$ .010} & \textbf{.454} {\scriptsize $\pm$ .000} & .619 {\scriptsize $\pm$ .000} & .690 {\scriptsize $\pm$ .000} & .529 {\scriptsize $\pm$ .003} \\
    SP & .124 {\scriptsize $\pm$ .000} & .088 {\scriptsize $\pm$ .000} & .076 {\scriptsize $\pm$ .000} & .534 {\scriptsize $\pm$ .000} & .402 {\scriptsize $\pm$ .000} & .245 {\scriptsize $\pm$ .000} \\
    Logistic regression & .395 {\scriptsize $\pm$ .030} & .471 {\scriptsize $\pm$ .010} & .394 {\scriptsize $\pm$ .000} & \textbf{.812} {\scriptsize $\pm$ .012} & \textbf{.736} {\scriptsize $\pm$ .026} & .562 {\scriptsize $\pm$ .009} \\
    Bayes Fused (SP) & \textbf{.455} {\scriptsize $\pm$ .010} & .443 {\scriptsize $\pm$ .000} & .210 {\scriptsize $\pm$ .010} & .660 {\scriptsize $\pm$ .004} & .607 {\scriptsize $\pm$ .005} & .475 {\scriptsize $\pm$ .003} \\
    SBF & .446 {\scriptsize $\pm$ .010} & \textbf{.528} {\scriptsize $\pm$ .010} & .432 {\scriptsize $\pm$ .020} & \textbf{.812} {\scriptsize $\pm$ .003} & .729 {\scriptsize $\pm$ .004} & \textbf{.589} {\scriptsize $\pm$ .005} \\
    \bottomrule
    \end{tabular}
    }
\end{table}

\ecoparagraph{Paired task-level bootstrap.} To address task-sampling uncertainty directly, we compute paired bootstrap intervals over tasks for ALFWorld, WebArena, OSWorld, and DeepSWE (\Cref{tab:bootstrap}). Tasks are resampled jointly across all configurations of each benchmark, and both methods are compared on the same out-of-fold predictions. Five of the six intervals exclude zero, and for the sixth (Bayes Fused (SP) vs. logistic regression on DeepSWE), Bayes Fused (SP) outperforms logistic regression with bootstrap probability .97.

\begin{table}[h]
    \caption{\textbf{Paired task-level bootstrap.} Difference in PRR@0.5 ($\Delta$PRR) between two methods and its 95\% bootstrap confidence interval. Tasks are resampled jointly across all configurations of each benchmark, and both methods are compared on the same out-of-fold predictions. ALFWorld and DeepSWE: 20 split seeds, $B=5000$; WebArena and OSWorld: the 3 split seeds of the main tables, $B=2000$.}
    \label{tab:bootstrap}
    \centering
    \resizebox{0.7\linewidth}{!}{%
    \begin{tabular}{ll|cc}
    \toprule
    \textbf{Benchmark} & \textbf{Comparison} & \textbf{$\Delta$PRR} & \textbf{95\% CI} \\
    \midrule
    ALFWorld & SBF vs.\ logistic regression & $+$.078 & [$+$.012, $+$.153] \\
    ALFWorld & Bayes Fused (SP) vs.\ SP & $+$.188 & [$+$.043, $+$.353] \\
    WebArena & SBF vs.\ SP & $+$.116 & [$+$.052, $+$.177] \\
    OSWorld & SBF vs.\ SP & $+$.175 & [$+$.035, $+$.312] \\
    DeepSWE & Bayes Fused (SP) vs.\ logistic regression & $+$.104 & [$-$.003, $+$.230] \\
    DeepSWE & Bayes Fused (SP) vs.\ SP & $+$.190 & [$+$.022, $+$.357] \\
    \bottomrule
    \end{tabular}
    }
\end{table}

\section{Limitations}
\label{app:limitations}

While our methods substantially improve failure detection for AI agents, no uncertainty quantification method is perfect. Some failures may remain undetected, and some successful trajectories may still receive low confidence. Therefore, these methods should be viewed as assistance for verification rather than as an endpoint oracle. In addition, although we evaluate across a wide range of environments, models, and harnesses, our evaluation is limited to single-agent systems and does not cover multi-agent systems. Finally, critics are defined for each benchmark from the information its harness provides, and their informativeness varies across benchmarks (e.g., CRAG), which limits the gain from tool feedback in such environments.

\end{document}